%% file: main.tex
\documentclass{article}

\usepackage{arxiv}

\usepackage[utf8]{inputenc} 
\usepackage[T1]{fontenc}    
\usepackage{hyperref}       
\usepackage{url}            
\usepackage{booktabs}       
\usepackage{amsfonts}       
\usepackage{nicefrac}       
\usepackage{microtype}      
\usepackage{lipsum}
\usepackage{graphicx}

\usepackage{amsmath,amsfonts}

\usepackage{algorithm}
\usepackage{algpseudocode}
\usepackage{array}
\usepackage[caption=false,font=normalsize,labelfont=sf,textfont=sf]{subfig}
\usepackage{textcomp}
\usepackage{stfloats}
\usepackage{hyperref}
\usepackage{url}
\usepackage{verbatim}
\usepackage{graphicx}
\usepackage{multirow}
\usepackage[table,xcdraw]{xcolor}
\usepackage{colortbl}
\usepackage{tcolorbox}
\usepackage{lipsum}
\usepackage{subfig}
\usepackage{pifont}%
\usepackage{pgf,tikz}
\usepackage{pgfplots,pgfplotstable}
\pgfplotsset{compat=1.17}
\usepackage[utf8]{inputenc}
\usepackage[export]{adjustbox}
\newcommand{\cmark}{\ding{51}}%
\newcommand{\xmark}{\ding{55}}%
\tcbuselibrary{listings,breakable}
\usepgfplotslibrary{colorbrewer}
\usetikzlibrary{pgfplots.statistics, pgfplots.colorbrewer} 
\usetikzlibrary{patterns}

\usepackage{balance}
\newtcolorbox{textbox}{
    colback=white,
    colframe=black,
    boxrule=.5pt,
    breakable,
}

\title{UserCentrix: An Agentic Memory-augmented AI Framework for Smart Spaces}

\author{
 Alaa Saleh \\
  Center for Applied Computing\\
  University of Oulu\\
  Oulu, 90014, Finland  \\
  \texttt{alaa.saleh@oulu.fi} \\
   \And
 Sasu Tarkoma \\
  Department of Computer Science\\
  University of Helsinki\\
  Helsinki, 00100, Finland \\
  \texttt{sasu.tarkoma@helsinki.fi} \\
  \And
 Praveen Kumar Donta \\
  Department of Computer and Systems Sciences\\
  Stockholm University\\
  Stockholm, 106 91, Sweden \\
  \texttt{praveen@dsv.su.se} \\
  \And
  Anders Lindgren \\
  RISE Research Institutes of Sweden \\
  Stockholm, 166 40, Sweden\\
  Department of Computer Science \\
  Luleå University of Technology\\
  Luleå, 971 87, Sweden \\
  \texttt{anders.lindgren@ri.se} \\
 \And
 Naser Hossein Motlagh \\
  Department of Computer Science\\
  University of Helsinki\\
  Helsinki, 00100, Finland \\
  \texttt{naser.motlagh@helsinki.fi} \\
  \And  
 Schahram Dustdar \\
  Distributed Systems Group\\
  TU Wien and ICREA Barcelona\\
  Vienna, 1040, Austria \\
  \texttt{dustdar@dsg.tuwien.ac.at} \\
  \And  
 Susanna Pirttikangas \\
  Center for Applied Computing\\
  University of Oulu\\
  Oulu, 90014, Finland  \\
  \texttt{susanna.pirttikangas@oulu.fi} \\
   \And
 Lauri Lovén \\
  Center for Applied Computing\\
  University of Oulu\\
  Oulu, 90014, Finland  \\
  \texttt{lauri.loven@oulu.fi} \\
}

\begin{document}
\input{results.dat}
\maketitle
\begin{abstract}
Agentic Artificial Intelligence (AI) constitutes a transformative paradigm in the evolution of intelligent agents and decision-support systems, redefining smart environments by enhancing operational efficiency, optimizing resource allocation, and strengthening systemic resilience. This paper presents \texttt{\textbf{UserCentrix}}, a hybrid agentic orchestration framework for smart spaces that optimizes resource management and enhances user experience through urgency-aware and intent-driven decision-making mechanisms. The framework integrates interactive modules equipped with agentic behavior and autonomous decision-making capabilities to dynamically balance latency, accuracy, and computational cost. User intent functions as a governing control signal that prioritizes decisions, regulates task execution and resource allocation, and guides the adaptation of decision-making strategies to balance trade-offs between speed and accuracy. Experimental results demonstrate that the framework autonomously enables efficient intent processing and real-time monitoring, while balancing reasoning quality and computational efficiency, particularly under resource-constrained edge conditions.

\end{abstract}

\maketitle

\input{sections/Introduction}
\input{sections/RelatedWork}
\input{sections/Methodology}
\input{sections/Implementation}
\input{sections/Results}

\input{sections/Discussion}

\input{sections/Conclusion}

\input{sections/Acknowledgments}

\def\refname{REFERENCES}
\bibliographystyle{ieeetr}
\bibliography{references}


\end{document}

%% file: sections/Introduction.tex
\section{Introduction}
Personalization and real-time adaptability have emerged as critical factors in user experience quality, driven by the rapid growth of user-centric ecosystems~\cite{10.1145/3641289,motlagh2023digital}. These factors involve the accurate interpretation of user intent and the capacity to adjust system behavior in accordance with the urgency of user needs. In this context, language model (LM)-powered artificial intelligence (AI) agents demonstrate significant potential to enhance user experiences by performing intent reasoning, formulating structured action plans, and executing them to achieve user-specific goals~\cite{10.1145/3640794.3665572,10.1145/3643505,10.1145/3712701,10.1145/3686803}.

These agents are capable of inferring user intent, capturing and leveraging evolving user preferences, and retrieving relevant information to deliver designed and tailored user experiences~\cite{10.1145/3686215.3688372,10.1145/3698145,10.1145/3701716.3717734}. By continuously adapting to dynamic and context-dependent needs, LLM-based agents contribute to the development of smart spaces that evolve toward increasingly personalized and intelligent configurations~\cite{10.1145/3731446,10.1145/3715114}.

Over time, these agents refine their performance through the accumulation and integration of experiential knowledge, enabling more accurate, context-aware, and goal-oriented interactions that align with the principles of next-generation user-centric systems~\cite{10.1145/3716132}. In this regard, memory-augmented agents support both individual reasoning processes and collaborative inter-agent coordination~\cite{10.1145/3748302,saleh2025memindex}. By retaining and structuring relevant experiences, agents learn from prior interactions and adapt efficiently to emerging user requirements while maintaining real-time responsiveness. Operating within dynamic environments, these intelligent agents orchestrate adaptive decision-making processes across user devices and edge servers to deliver personalized assistance~\cite{saleh2025llm}. This orchestration introduces increasing complexity as the number of agents grows and their interactions extend across distributed and resource-constrained infrastructures~\cite{saleh2023pub,saleh2024follow}.

Within this context, it is essential to investigate how the number of agents, their reasoning capabilities, and their individual resource requirements affect overall system performance and output quality. Scaling the agent population may enhance problem-solving capacity and robustness; however, it can also introduce additional communication overhead, synchronization costs, and latency. A systematic understanding of these trade-offs is therefore crucial for designing multi-agent AI systems that achieve an effective balance between scalability, efficiency, and output quality in dynamic operational contexts. Addressing this growing complexity requires deploying auto-scaling mechanisms capable of allocating computational and communication resources in accordance with user intent urgency levels. Such mechanisms should be complemented by scalable memory management and continual learning strategies that enhance responsiveness and adaptability across the multi-agent network~\cite{10.1145/3716629,10.1145/3706418,10.1145/3748302}.

These considerations highlight the necessity of a design framework that integrates core agentic AI principles, intentionality, adaptability, and autonomy~\cite{biswas2025building}. The framework should balance reactivity and proactivity within an agentic architecture. It enables agents to learn from experience, continuously refine their strategies, and dynamically regulate resources in response to real-time fluctuations and varying levels of intent urgency. Concurrently, it equips agents with proactive capabilities, allowing them to anticipate needs, plan accordingly, and act in advance of emerging demands. Through these capabilities, the framework delivers adaptive assistance while preserving robustness and fault tolerance in dynamic operational environments.

Building upon this foundation, we propose the \texttt{\textbf{UserCentrix}} framework, a hybrid agentic orchestration architecture that integrates multiple interactive modules, enabling rapid adaptation to dynamic conditions while supporting planning, reasoning, and decision-making aligned with user intent to deliver user-centric services within smart spaces. The primary contributions of this work are summarized as follows:
\begin{itemize}
\item We develop a personalized LM agent designed as a knowledge-driven AI system with scalable memory and self-evaluation mechanisms to enhance decision reliability and response efficiency.
\item We propose an agentic architecture incorporating proactive auto-scaling, in-context learning, and pareto-based optimization, which dynamically adapts decision-making strategies and allocates inference time budgets according to the urgency of user intent.
\item We design cooperative reasoning networks composed of multiple agents that leverage rendezvous points (RPs) to negotiate collaboration terms, mitigate intent conflicts, and ensure alignment with diverse user requirements.
\item We implement a management and analysis module to regulate ongoing commands dispatched through a message queue equipped with time-to-launch (TTL) settings, ensuring that control commands are delivered to both the control system and the environment agent at the specified time.
\item We incorporate environmental awareness mechanisms that enable the environment agent to dynamically monitor changes and generate adaptive commands, thereby maintaining alignment with user intents and ensuring fault tolerance.
\item Our experiments examine the performance of each agentic module in generating intent-driven, resource-efficient autonomous decisions, assessed through Human-in-the-Loop~\cite{WU2022364} as a reference and LLM-as-a-Judge assessment~\cite{gu2024survey}.
\end{itemize}

The remainder of this paper is structured as follows: Section~\ref{sec:RelatedWork} reviews recent studies, highlighting their objectives and limitations. Section~\ref{sec:Methodology} outlines a general framework for smart spaces. Section~\ref{sec:Implementation} details the implemented scenario along with its requirements. Section~\ref{sec:Results} analyzes the experimental outcomes using various LLMs and Small Language Models (SLMs). Section~\ref{sec:Discussion} discusses key insights for deploying framework, limitations, and directions for future research. Finally, Section~\ref{sec:Conclusion} summarizes the findings.


%% file: sections/RelatedWork.tex
\section{Related Works} \label{sec:RelatedWork}
In this section, we review recent research published between 2024 and 2025, with a focus on the integration of LM agents within the computing continuum, recent structures of multi-agent systems, and emerging techniques for enhancing the reasoning capabilities and response quality of LMs, as described below:

\subsection{LLMs for edge-cloud continuum}
The integration LLM agents within the computing continuum represents a promising research direction~\cite{meuser2024revisiting}, paving the way for more effective applications. Several studies have explored deploying LLMs in edge-cloud computing environments and highlighted the potential of LLM agents across the edge-cloud continuum by addressing computational, latency, and resource management challenges through innovative edge-cloud collaboration and optimization strategies. 


Shen et al.~\cite{shen2024large} proposed a cloud-edge-client hierarchical framework that enables edge AI systems to automatically organize, adapt, and optimize themselves to meet users’ diverse requirements. By leveraging LLMs, the framework efficiently coordinates edge AI models to interpret user intentions and cater to personalized demands. A collaborative edge computing framework for LLM inference was proposed in~\cite{zhang2024edgeshard}. This framework employs dynamic programming to partition models into shards and deploy them on distributed devices spanning edge devices and cloud servers. This hybrid approach facilitates collaboration between edge and cloud resources.

Hao et al.~\cite{hao2024hybrid} proposed a hybrid inference framework featuring dynamic token-level edge-cloud collaboration. This framework balances both edge and cloud resources utilization to enhance inference performance. Yu et al.~\cite{yu2024edge} developed Edge-LLM, a decentralized framework focused on optimizing LLM adaptation on edge devices through integrating layer-wise compression, adaptive layer tuning, and a hardware scheduling strategy for computational efficiency. 
Ding et al.~\cite{ding2024hybrid} propose a method for optimizing service placement strategies by considering both model requests and the associated computational resource requirements. Their approach employs a routing mechanism that dynamically assigns queries to either a small or large model, based on the predicted difficulty of the query. DLoRA~\cite{gao2024dlora} presents a distributed PEFT framework, in which the LLM is executed on cloud servers, while the PEFT modules are trained entirely on user devices. A cloud-edge collaborative inference framework for edge intelligence to efficiently deploy LLM agents is proposed in~\cite{xu2024cached}. This work focuses on optimizing service placement and inference task offloading strategies, leveraging cached LLMs on both cloud and edge servers.

\subsection{Structures design of multi-agent systems}
Several recent approaches provide valuable insights into designing system structures that effectively harness collective capabilities in multi-agent systems. Some approaches focus on hierarchical architectures. For instance, a hierarchical structure with dynamic organization based on task requirements is introduced in~\cite{guo2024embodied}. MAP architecture~\cite{10.1145/3706599.3719853} employs a multi-agent architecture in which specialized LLM-powered agents collaborate across different stages to provide adaptive personalization in multi-user settings. Mixture-of-Agents (MoA) architecture~\cite{wang2406mixture} uses multiple LLMs organized across layers, allowing iterative refinement of outputs through collective agent input. These recent approaches underscore how multi-agent systems that employ hierarchical coordination and dynamic role adaptation can enhance response quality by leveraging collective capabilities. 




\subsection{Reasoning LLM agent}
Recent advancements in LLM research have focused on enhancing reasoning and response quality, aiming to leverage the thinking capabilities of LLM agents more effectively. These include scaling inference-time computing by adapting based on prompt difficulty~\cite{snell2024scaling}, illustrating the need for efficient use of computational resources during inference. 
Other strategies focus on real-time adaptation of reasoning techniques to meet specific task requirements. For example, a dual-system approach~\cite{christakopoulou2024agents} enables dynamic adaptation through the two modes of thinking based on task requirement, while Qi et al.~\cite{qi2024mutual} explore generating multiple reasoning paths to enhance LLM reasoning capabilities. 


Additional methods focus on enhancing reasoning during the text generation process itself, such as refining internal thoughts~\cite{wu2024thinking,zelikman2403quiet}, adjusting reasoning depth~\cite{10.1145/3706598.3713606}, and determining recall probability to enable more contextually coherent and personalized responses~\cite{hou2024my}.\\

\noindent
Table~\ref{tab:Related} presents a summary of these works, outlining their goals, algorithms, main tasks, and limitations. While prior studies have focused individually on hierarchical multi-agent coordination, LLM reasoning enhancement, or edge–cloud optimization, \texttt{\textbf{UserCentrix}} integrates these elements into a hybrid agentic architecture for smart spaces, guided by user intent. Through adaptive orchestration, cooperative negotiation, and continuous memory-augmented reasoning, it balances accuracy, cost, and resource efficiency in decision-making under dynamic, resource-constrained conditions.

\begin{table*}[]
\centering
\caption{Summary of Related Works.}
\label{tab:Related}
\resizebox{!}{.23\paperheight}{%
\begin{tabular}{|
>{\columncolor[HTML]{FFFFFF}}c |
>{\columncolor[HTML]{FFFFFF}}l |
>{\columncolor[HTML]{FFFFFF}}l |
>{\columncolor[HTML]{FFFFFF}}l |
>{\columncolor[HTML]{FFFFFF}}l |}
\hline
\textbf{Name} &
  \multicolumn{1}{c|}{\cellcolor[HTML]{FFFFFF}\textbf{Task}} &
  \multicolumn{1}{c|}{\cellcolor[HTML]{FFFFFF}\textbf{Algorithm}} &
  \multicolumn{1}{c|}{\cellcolor[HTML]{FFFFFF}\textbf{Goal}} &
  \multicolumn{1}{c|}{\cellcolor[HTML]{FFFFFF}\textbf{Limitation}} \\ \hline
\textbf{\begin{tabular}[c]{@{}c@{}}Cloud-Edge-Client \\Framework~\cite{shen2024large}\end{tabular}} & \begin{tabular}[c]{@{}l@{}}Develop a hierarchical framework for autonomous \\edge AI systems. \end{tabular} & \begin{tabular}[c]{@{}l@{}} LLMs to organize, adapt, and optimize\\ edge AI systems automatically.\end{tabular} & \begin{tabular}[c]{@{}l@{}} Meet diverse user requirements with \\minimal latency.\end{tabular} & \begin{tabular}[c]{@{}l@{}} -Limited focus on energy efficiency.\\
-Resource allocation challenges in \\highly dynamic environments.\end{tabular} \\ \hline 
    \textbf{EdgeShard~\cite{zhang2024edgeshard}} & \begin{tabular}[c]{@{}l@{}} Design a collaborative edge computing framework\\ for efficient LLM inference.\end{tabular} & \begin{tabular}[c]{@{}l@{}} Dynamic programming to partition LLMs \\between edge and cloud resources. \end{tabular} & \begin{tabular}[c]{@{}l@{}} Minimize inference latency and maximize\\ throughput.\end{tabular} & \begin{tabular}[c]{@{}l@{}} Resource allocation challenge.\end{tabular} \\ \hline 
    \textbf{\begin{tabular}[c]{@{}c@{}}Hybrid Inference\\Framework~\cite{hao2024hybrid}\end{tabular}} & \begin{tabular}[c]{@{}l@{}}Propose a hybrid inference framework for LLM\\ inference. \end{tabular} & \begin{tabular}[c]{@{}l@{}} Dynamic token-level interactions during\\ decoding time, combining edge and cloud \\resources for inference.  \end{tabular} & Enhance inference performance & \begin{tabular}[c]{@{}l@{}} Limited latency improvements.\end{tabular} \\ \hline 
    \textbf{Edge-LLM~\cite{yu2024edge}}  & \begin{tabular}[c]{@{}l@{}} Develop a framework for optimizing LLM adaptation\\ on edge devices.  \end{tabular} &  \begin{tabular}[c]{@{}l@{}}Layer-wise compression (pruning and \\quantization), adaptive layer tuning (voting \\mechanism), and hardware scheduling\end{tabular} & \begin{tabular}[c]{@{}l@{}} -Mitigate high computational and memory\\ demands of LLMs.\\-Optimize performance on resource-\\constrained edge devices.  \end{tabular} & \begin{tabular}[c]{@{}l@{}} Lack focus on power consumption.\end{tabular} \\ \hline 
    \textbf{Hybrid LLM~\cite{ding2024hybrid}} & \begin{tabular}[c]{@{}l@{}} Optimizing service placement strategies for LLMs. \end{tabular} &  \begin{tabular}[c]{@{}l@{}}Considering model requests and \\computational resource requirements, and \\dynamically assign queries to either a \\small or large model\end{tabular} & \begin{tabular}[c]{@{}l@{}} Improve memory and storage efficiency.\end{tabular} & \begin{tabular}[c]{@{}l@{}} Insufficient to address the real-world\\ need for a diverse array of LLMs.\end{tabular} \\ \hline 
    \textbf{\begin{tabular}[c]{@{}c@{}}Cached model-as-\\a-resource~\cite{xu2024cached}\end{tabular}} & \begin{tabular}[c]{@{}l@{}} Propose service placement and inference task \\offloading strategies in a cloud-edge collaborative\\ inference framework. \end{tabular} &  \begin{tabular}[c]{@{}l@{}}Auction mechanism.\end{tabular} & \begin{tabular}[c]{@{}l@{}} Minimize the total inference cost of\\ edge servers and the cloud.\end{tabular} & \begin{tabular}[c]{@{}l@{}} Scalability challenge to serve a \\higher volume of user requests \\simultaneously.\end{tabular} \\ \hline 
    \textbf{Dlora\cite{gao2024dlora}}& \begin{tabular}[c]{@{}l@{}} Propose framework for collaborative training of \\LLMs across cloud and edge devices. \end{tabular} &  \begin{tabular}[c]{@{}l@{}}Distributed PEFT approach where the LLM\\ parameters are stored in cloud servers,\\ while PEFT modules are fine-tuned on edge\\ devices.\end{tabular} & \begin{tabular}[c]{@{}l@{}} -Reduce the computational workload on \\edge devices.\\
-Minimize communication overhead.\end{tabular} & \begin{tabular}[c]{@{}l@{}} -Challenges in resource-constrained \\networks. \\
-Limitations of computational resources \\on edge devices.\end{tabular} \\ \hline 
\textbf{Criticize-Reflect~\cite{guo2024embodied}} &
  \begin{tabular}[c]{@{}l@{}}Explore how hierarchical structures and leadership \\ roles impact multi-agent coordination.\end{tabular} &
  \begin{tabular}[c]{@{}l@{}}Dynamically organizes LLMs based on task \\ requirements.\end{tabular} &
  \begin{tabular}[c]{@{}l@{}}Achieve continuous improvement in \\ communication efficiency and cooperation.\end{tabular} &
  \begin{tabular}[c]{@{}l@{}}Lack scalability with more agents in \\large environments.\end{tabular} \\ \hline
\textbf{MAP~\cite{10.1145/3706599.3719853}} &
   \begin{tabular}[c]{@{}l@{}}Address multi-user personalization, \\focusing on conflicting user preferences.\end{tabular}&
  \begin{tabular}[c]{@{}l@{}}Delegate sub-tasks to specialized agents \\that collaborate across three stages\end{tabular} &
  \begin{tabular}[c]{@{}l@{}}Design a transparent and adaptable \\personalization AI framework for multi-user \\environments\end{tabular} & Need for monitoring mechanisms
  \\ \hline
\textbf{Mixture-of-Agents~\cite{wang2406mixture}} &
  \begin{tabular}[c]{@{}l@{}}Utilize multiple LLMs organized across layers for \\ iterative output refinement.\end{tabular} &
  \begin{tabular}[c]{@{}l@{}}Agents share and refine information across \\layers through cycles.\end{tabular} &
  \begin{tabular}[c]{@{}l@{}}Enhance response quality by leveraging \\the unique strengths of diverse LLMs, \\ demonstrating "collaborativeness".\end{tabular} &
  \begin{tabular}[c]{@{}l@{}}Dependence on multiple MoA layers, \\ increasing computational overhead.\end{tabular} \\ \hline
\textbf{\begin{tabular}[c]{@{}c@{}}Compute-Optimal \\ Scaling~\cite{snell2024scaling}\end{tabular}} &
  \begin{tabular}[c]{@{}l@{}}Adapt compute allocation during testing based on \\ prompt difficulty.\end{tabular} &
  \begin{tabular}[c]{@{}l@{}}Smarter use of computational resources \\ during the test phase.\end{tabular} &
  \begin{tabular}[c]{@{}l@{}}Improve efficiency and performance during \\ inference.\end{tabular} &
  \begin{tabular}[c]{@{}l@{}}Limiting LLMs' reasoning robustness \\ and generalizability due to difficulties \\ for verifiers in identifying errors.\end{tabular} \\ \hline
\textbf{\begin{tabular}[c]{@{}c@{}}Talker-Reasoner \\ Agent Model~\cite{christakopoulou2024agents}\end{tabular}} &
  \begin{tabular}[c]{@{}l@{}}Utilize a dual-agent framework inspired by \\ Kahneman's "Thinking Fast and Slow".\end{tabular} &
  \begin{tabular}[c]{@{}l@{}}Combines real-time interaction (Talker) \\ with multi-step problem-solving (Reasoner).\end{tabular} &
  \begin{tabular}[c]{@{}l@{}}Enable dynamic adaptation and improve \\ efficiency.\end{tabular} &
  \begin{tabular}[c]{@{}l@{}}-No strategy for minimizing Reasoner\\ use when Talker suffices.  \\ -Lack of automatic Talker-Reasoner \\ switching based on query complexity.\end{tabular} \\ \hline
\textbf{\begin{tabular}[c]{@{}c@{}}Thought Preference\\  Optimization~\cite{wu2024thinking}\end{tabular}} &
  \begin{tabular}[c]{@{}l@{}}Equip LLMs with the ability to think before \\ responding.\end{tabular} &
  \begin{tabular}[c]{@{}l@{}}Train LLMs to generate and optimize \\ internal thoughts iteratively.\end{tabular} &
  \begin{tabular}[c]{@{}l@{}}Produce thoughtful and accurate outputs \\ for complex instructions.\end{tabular} &
  \begin{tabular}[c]{@{}l@{}}Lack of exploration the thinking with \\ larger-scale models and a more diverse \\ set of thought prompts.\end{tabular} \\ \hline
\textbf{Quiet-STaR~\cite{zelikman2403quiet}} &
  \begin{tabular}[c]{@{}l@{}}Simulate internal thought processes during text\\ generation.\end{tabular} &
  \begin{tabular}[c]{@{}l@{}}Generate useful internal rationales for each \\ token to guide predictions using \\ REINFORCE learning.\end{tabular} &
  \begin{tabular}[c]{@{}l@{}}Predict future text more accurately.\\ Improve capabilities for reasoning tasks.\end{tabular} &
  \begin{tabular}[c]{@{}l@{}}Computationally intensive as it is \\ applied at every token.\end{tabular} \\ \hline
\textbf{rStar~\cite{qi2024mutual}} &
  \begin{tabular}[c]{@{}l@{}}Propose collaborative problem-solving approach.\end{tabular} &
  \begin{tabular}[c]{@{}l@{}}SLM with Monte Carlo Tree Search to\\ generate reasoning pathways, while another \\SLM evaluates them.\end{tabular} &
  \begin{tabular}[c]{@{}l@{}}Enhance reasoning capabilities of SLMs.\end{tabular} &
  \begin{tabular}[c]{@{}l@{}}Dependency on the ability of SLM\\ to verify reasoning quality.\end{tabular} \\ \hline
\textbf{\begin{tabular}[c]{@{}c@{}}Human-like\\ Memory~\cite{hou2024my}\end{tabular}} &
  \begin{tabular}[c]{@{}l@{}}Integrate human-like memory processes into\\ LLM-based dialogue agents.\end{tabular} &
  \begin{tabular}[c]{@{}l@{}}Calculate memory recall probability using an \\exponential decay function modulated by \\relevance using cosine similarity, elapsed time\\, and recall frequency.\end{tabular} &
  \begin{tabular}[c]{@{}l@{}}Enable agents to autonomously produce \\more personalized dialogues.\end{tabular} &
  \begin{tabular}[c]{@{}l@{}}Adaptability to user’s behavior \\significant changes.\end{tabular} \\ \hline
\textbf{\begin{tabular}[c]{@{}c@{}}LLM Coaching\\ Agent~\cite{10.1145/3706598.3713606}\end{tabular}} &
  \begin{tabular}[c]{@{}l@{}}Investigate how to manage the reasoning depth \\of LLM-based coaching agents.\end{tabular} &
  \begin{tabular}[c]{@{}l@{}}A discrepancy mechanism which integrate \\the GROW model and behavioral change\\ techniques.\end{tabular} &
  \begin{tabular}[c]{@{}l@{}}Align long-term goals with \\short-term user expectations.\end{tabular} &
  \begin{tabular}[c]{@{}l@{}}Generalizability constrained where \\ objectives’ nature differs.\end{tabular} \\ \hline
\textbf{\begin{tabular}[c]{@{}c@{}}UserCentrix\\(Proposed \\Framework)\end{tabular}} &
  \begin{tabular}[c]{@{}l@{}}Introduce an agentic, memory-augmented\\ AI framework for autonomous, personalized,\\ intent-aware decision-making in smart spaces.\end{tabular} &
  \begin{tabular}[c]{@{}l@{}}Hybrid agentic architecture \\with value of intent prioritization,\\ in-context learning, and pareto \\optimization.\end{tabular} &
  \begin{tabular}[c]{@{}l@{}}Balance responsiveness, accuracy, and \\computational cost in smart spaces\end{tabular} &
  \begin{tabular}[c]{@{}l@{}}Decision quality is tightly coupled to \\the accuracy and reasoning capacity\\ of LMs.\end{tabular} \\ \hline
  
\end{tabular}%
}
\end{table*}

%% file: sections/Methodology.tex
\section{\texttt{\textbf{UserCentrix}} Framework}\label{sec:Methodology}

This paper introduces \texttt{\textbf{UserCentrix}}, an agentic AI orchestration framework for smart spaces. This framework dynamically adapts to the urgency of user intent and autonomously scales computational resources according to task demands. It adopts a dual-layer architecture comprising a user side and a building side, each designed to enable intelligent and efficient interaction within smart environments as shown in Fig.~\ref{fig:methodology}. On the user side, personalized knowledge-driven AI agents~\cite{russell2020artificial}, supported by LM and scalable memory, function as intelligent assistants. These agents maintain individualized knowledge bases and continuously learn, evaluate, and adapt to evolving user intents, thereby enhancing user experience. On the building side, multiple AI agents manage the smart environment by optimizing resource allocation through three core modules equipped with adaptive agentic behavior and autonomous decision-making capabilities. Subsequent sections detail the workflow, module functionalities, and agent roles. Algorithms and use cases, along with illustrative figures presenting inputs, outputs, and intermediate processes, are included to clarify the framework’s operation.

\begin{figure*}[t]
\centering
\includegraphics[width=1\textwidth]{images/centrix.png}
\caption{\texttt{\textbf{UserCentrix}} Framework.}\label{fig:methodology}
\end{figure*}

\subsection{User Side}
The effectiveness of personalized LM-powered agents in the \texttt{\textbf{UserCentrix}} framework depends on their capacity to interpret user intent and adapt to evolving requests. This adaptability extends beyond immediate intent understanding to the reuse of prior knowledge through memorizing capability, enabling efficient and intent-aware responses. The integration of self-evaluation further improves response reliability. These knowledge-based LM agents maintain an internal memory representation that supports Case-Based Reasoning (CBR)~\cite{watson1994case}. Each user intent is analyzed and stored as a textual description within a structured knowledge base. This repository contains records of prior interactions, including plan types, timestamps, preferences, and execution details. By leveraging this historical information, the agent refines decision-making in new scenarios and improves intent accuracy as shown in Figure~\ref{fig:methodology}.

The personalized agent initializes with an empty, scalable repository that expands as new intents are submitted. Historical data are automatically incorporated into subsequent prompts, ensuring coherent responses. Supported intents include smart space configuration, meeting room reservations, meal ordering, and other personalized services, thereby enhancing operational efficiency and user experience. Upon receiving a new intent, the agent evaluates whether explicit preferences are provided. If absent, it computes semantic similarity between the embedding of the current plan type and those of previous plans, while also comparing timestamps. When high similarity is detected within a one-hour interval, the most recent matching case is retrieved and its preferences are applied automatically. The knowledge base is then updated accordingly, ensuring adaptive intent generation and intent-aware user experience. The self-evaluation mechanism further validates the alignment between generated outputs and the original user intent. By comparing responses with intended objectives, the agent enhances accuracy and trustworthiness. If prior cases are retrieved, the agent justifies the retrieval to ensure transparency.


\subsection{Smart Building Side}
The framework is composed of three modules as shown in Fig.\ref{fig:methodology}. The \textbf{Decision-making Module} comprises high-level, utility-based AI agents that select actions to maximize overall effectiveness while balancing responsiveness and precision according to intent urgency. Selected actions are decomposed into sub-tasks according to their dependencies and executed by the \textbf{Sub-tasks Execution Module}, which consists of low-level agents. These agents generate executable commands using real-time data from the building knowledge layer to adjust environmental settings. The commands are handled by the \textbf{Management and Analysis Module}, which stores them in a message queue and dispatches them based on a predefined schedule. Its environment agent continuously monitors applied settings during the assigned time window and issues corrective commands if unexpected deviations occur. Detailed descriptions of each module follow.

\subsubsection{Decision-making Module:} This module begins with a \textbf{Classifier AI Agent}, which assigns time slices to user intents based on urgency. The classifier extracts the current time and compares it with the planned execution time retrieved from the repository. Intents are categorized into two levels:
\begin{itemize}
    \item High-urgency ($\mathcal{U} \geq \vartheta_1$) when the time difference is less than two hours.
    \item Low-urgency ($\mathcal{U} \leq \vartheta_1$) when the difference is two hours or more.
\end{itemize}
These time slices guide workflow design, reasoning depth, and communication among low-level agents. 


For critical intents, the module prioritizes speed through a \textbf{High-urgency AI Agent} ($\mathcal{A}_{High}$). This agent generates streamlined reasoning paths to accelerate execution while maintaining acceptable accuracy. It reduces sub-tasks, prioritizes high-impact actions, minimizes interdependencies, and limits LM calls. Independent calls are grouped for parallel execution to further enhance responsiveness. This strategy ensures timely handling of critical requests. 


For non-critical intents, the \textbf{Decision-making Module} allocates additional time to improve decision quality through the \textbf{Low-urgency AI Agent} ($\mathcal{A}_{Low}$). Unlike the high-urgency agent, this agent emphasizes accuracy and thorough analysis, enabling more resource-intensive computations. The low-urgency agent evaluates the decision process by generating multiple reasoning paths that differ in depth, complexity, and evaluation criteria. Each path decomposes the intent into detailed sub-tasks, including necessary LM calls, representing alternative solution strategies. These reasoning paths may prioritize specific criteria, such as real-time environmental analysis (e.g., room availability, meal options, temperature, and lighting conditions), alignment with user preferences, or optimization through natural and smart adjustments (e.g., increasing natural light by opening curtains). This diversity enables broad assessment of decision strategies and resource allocation options. 


The \textbf{Low-urgency AI Agent} is memory-augmented and follows an iterative learning process. After generating candidate solutions, each solution–intent pair is stored in external memory for selective reuse. The solutions are also evaluated by an \textbf{Evaluator LM Agent} ($\mathcal{A}_{Evalu}$), which selects the most optimal solution and provides logical justification. If no solution satisfies efficiency and success criteria, the evaluator returns actionable comments as a feedback to guide refinement. 

For new intents, the agent first computes semantic similarity between the current intent and stored intents. When a highly similar case is identified, its best solution, reasons, and feedback are retrieved and incorporated into the prompt as a \textit{hint} to guide the new solution generation. This feedback loop enables continuous improvement through in-context learning, allowing the agents to evolve their prompt and refine its solution-generation strategy and learn factors associated with superior outcomes. Final candidate solutions are assessed by a \textbf{Pareto Analyzer}, which applies multiple fitness functions to optimize decision-making. These functions evaluate: (i) semantic similarity, reflecting intent fulfillment; (ii) precision, measuring alignment between generated output and original intent; and (iii) LM call usage cost, quantified by the number of LM calls relative to a maximum calls of solutions ($N_{calls}$, $N_{max}$). This multi-objective evaluation ensures that gains in precision justify additional computational resources and extended inference time.

\begin{equation}\label{eq1}
\text{LM Call Usage Cost} = 1 - \exp\left(-\frac{N_{calls}}{N_{max}}\right)
\end{equation}

Eq.(\ref{eq1}) creates a decay effect where the cost scales non-linearly as the LM call count increases. For small call counts, the cost increases slowly, but it rises more steeply as the call count approaches maximum. This encourages minimizing resource consumption relative to available limits. To identify the optimal solution, we applied \textit{Pareto dominance}, aiming to minimize resource costs while maximizing semantic similarity ($\mathcal{S}$) and precision score.

\begin{equation}\label{eq2}
    Pareto = \min(LM Call Usage Cost), \max(\mathcal{S}), \max(Precision)
\end{equation}

This approach ensures an optimal balance between accuracy and resource efficiency. By evaluating trade-offs between accuracy and resource usage, the agent allocates resources according to expected utility. Solutions are ranked using \textit{Pareto} efficiency, and the dominant solution is chosen for execution. The selected solution is then forwarded to low-level agents. The number of sub-tasks determines whether a single agent can handle the intent independently or whether multiple agents must collaborate to address more complex, reasoning-intensive requests.

\input{Algorithms/algo1}

To clarify the workflow of the \textit{\texttt{\textbf{UserCentrix}} Decision-making Module}, we present Algorithm~\ref{alg:task_solution}. For each intent $T$ provided by the user, the classifier agent determines its urgency level by computing the difference between the intent's deadline $\mathcal{D}_i$ and the current time $t$ (Lines 1 \& 2). If the resulting classification score is below zero, indicating that the intent is no longer relevant, it is removed from the intent list (Lines 3 \& 4). Otherwise, intents are further categorized into low or high urgency based on a predefined threshold $\vartheta_1$ (Lines 5-6 \& 27-28). Low-urgency intents are assigned to a low-urgency agent $\mathcal{A}_{\text{Low}}$ (Line 7), which retrieves relevant past memory and computes semantic embeddings of both current and previous solutions using MiniLM (Lines 8–11). A similarity score is then calculated between the new and past embeddings. If the similarity is below a specified threshold $\vartheta_2$, the system proceeds to generate $n$ potential new solutions (Lines 12 \& 13); otherwise, it retrieves suitable existing solution with the reason and factors provided by the evaluator agent (Line 15) and use them as hints to generate potential solutions (Line 16). Each solution is subsequently evaluated in terms of semantic similarity (using LlamaIndex), precision, and usage cost (Lines 18–21). A Pareto optimization method is employed to select the most optimal solution $\mathcal{E}_i$ (Line 22). The evaluator agent is responsible for assessing the generated solutions and selecting the most appropriate one with its logical reasoning (Line 23), which will be injected along with the corresponding intent into the memory module $\mathcal{M}$ (Lines 24 \& 25). In contrast, high-urgency intents $\mathcal{A}_{\text{High}}$ are directed to the high-urgency agent, which quickly generates suitable solutions to meet tight deadlines (Lines 29 \& 30). The module concludes by returning the final set of updated sub-tasks $\mathcal{T}_i$ (Line 33).

\subsubsection{Sub-tasks Execution Module}
After a solution is selected, the \textbf{Sub-tasks Execution Module} organizes \textbf{Low-level AI Agents} into groups to execute sub-tasks and generate commands for adjusting smart building settings. These groups operate in parallel and use RPs as a shared memory to share intermediate outputs, accelerating decision-making while respecting intent timing constraints. When no execution conflicts exist, agents generate commands according to a hierarchical execution order. For parallel sub-tasks, agents can access relevant prior responses within the same execution stage. If conflicts arise (e.g., simultaneous room allocation), a meta-cooperative reasoning network is activated. In this setting, agents from different groups negotiate using RPs, exchanging intermediate reasoning results to prevent conflicts. This cooperative mechanism improves both response speed and decision accuracy.

\input{Algorithms/algo2}

Algorithm~\ref{alg:sub_tasks_execution} presents the \textbf{Sub-tasks Execution Module}, which is designed to execute a set of sub-task solutions $\mathcal{T} = \{1:k\}$ using a corresponding dataset $D$. The execution process begins with the creation of $k$ low-level agents, where each agent is assigned a specific sub-task $\mathcal{T}_i $ (Line 1). For each sub-task $\mathcal{T}_i $, the respective agent retrieves a compatible dataset $D$ that satisfies the intent’s requirements (Line 2 \& 3) to perform the execution of the sub-task (Line 4). The resulting commands are collected and aggregated into a final command set $\mathcal{C}$ (Line 5), which is returned upon completion of the execution phase (Line 7).
 
\subsubsection{Management and Analysis Module}
The \textbf{Management and Analysis Module} maintains a message queue that stores generated commands, including TTL parameter. It aggregates commands from low-level agents and dispatches them to the control system at the scheduled execution time. The \textbf{Environment Agent}, powered by a LM, monitors ongoing execution by comparing user-defined requirements with real-time resource status. If deviations are detected, it generates corrective or alert commands and sends them to the control system. This approach ensures user satisfaction and QoE remain high by proactively addressing potential issues during intent execution. 


\subsection{Use Case}

\subsubsection{User Intent Processing Scenario:}
\begin{figure}[!t]
\centering
\includegraphics[width=0.85\textwidth]{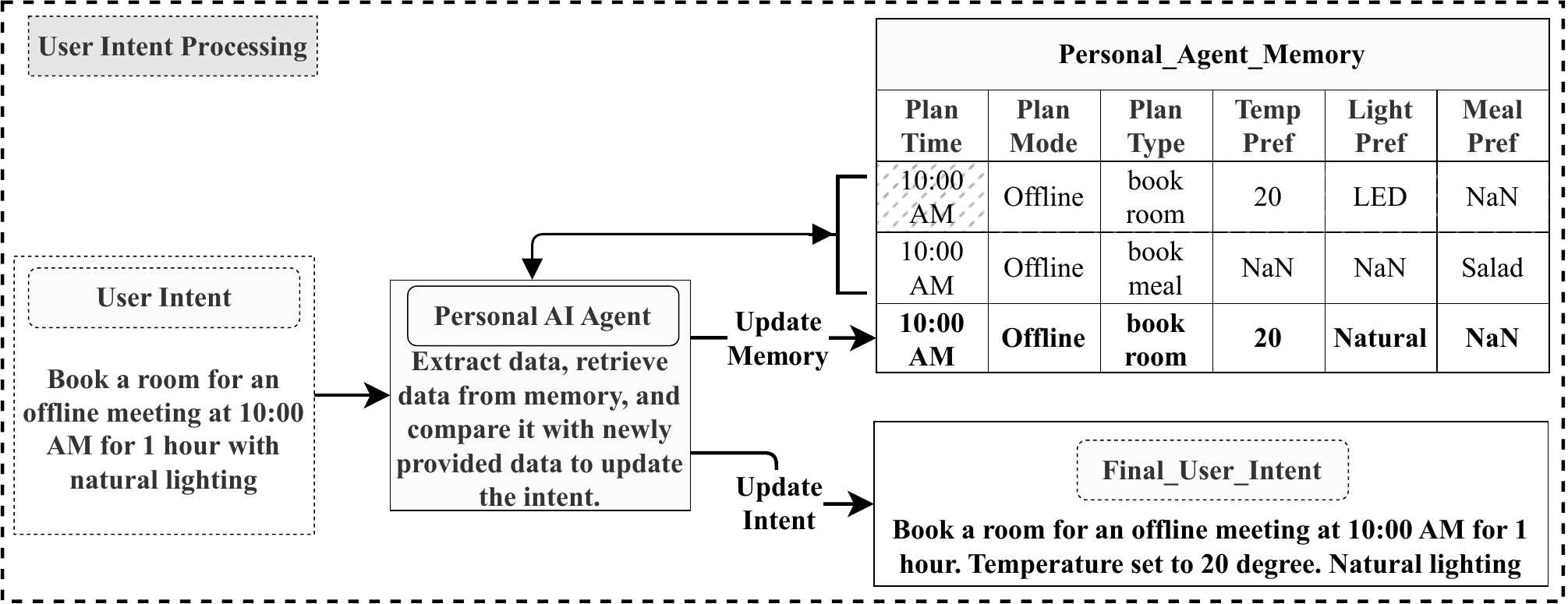}
\caption{User Intent Processing Use Case within \texttt{\textbf{UserCentrix}} Framework.}\label{fig:user}
\end{figure}

As shown in Fig.~\ref{fig:user} the personalized agent is equipped with an external personal memory that serves as a repository. It initially starts empty and gradually fills up as the user submits intents. This repository retains information about past intents. In this scenario, the user has previously submitted two intents, each with its own preferences stored in the repository. When the user submits a new intent without specifying preferences, the agent evaluates the semantic similarity between the new plan type embeddings and those stored in personal memory. It also compares the timestamp with previous timestamps. If the time difference is less than one hour and the similarity score exceeds 0.5, the agent retrieves preferences from the most recent matching entry in personal memory. The agent then updates the new intent to incorporate these preferences, allowing it to adapt its responses based on user history. This process ensures a context-aware experience, enabling more personalized and relevant intent updates.

\subsubsection{High-urgency Scenario:}
\begin{figure}[!]
\centering
\includegraphics[width=1\textwidth]{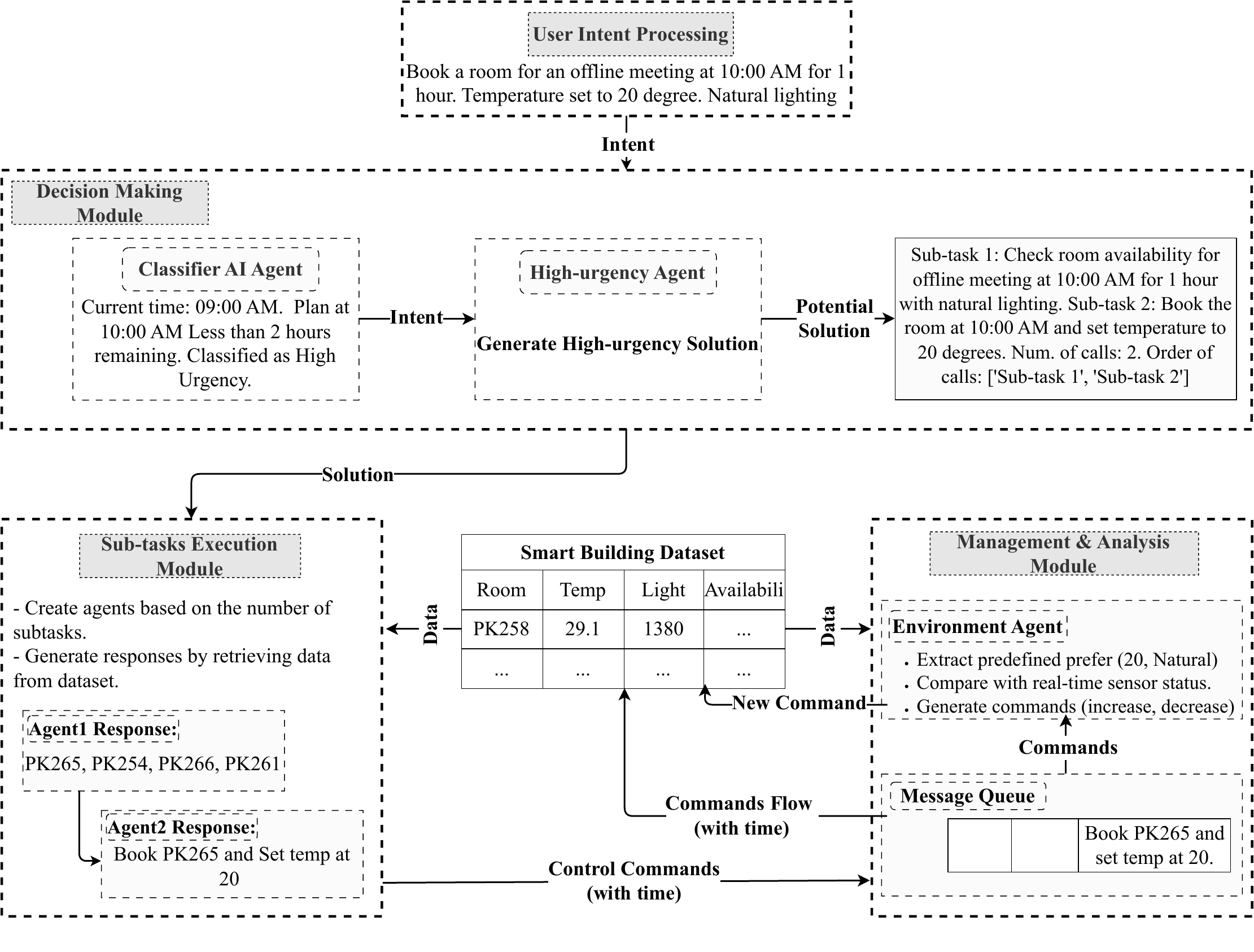}
\caption{High-urgency Workflow within \texttt{\textbf{UserCentrix}} Framework.}\label{fig:high}
\end{figure}

Fig.~\ref{fig:high} illustrates the workflow of the framework’s building modules, beginning with the classifier agent, which determines the urgency level of a submitted user intent. In this scenario, the agent classified the intent as high urgency because the time difference between the current time and the intent time was less than two hours. It then forwarded the intent to the high-urgency agent, which is responsible for generating a time-sensitive solution with minimal sub-tasks. This agent identifies the necessity for two separate LM calls, each corresponding to a distinct sub-task. These sub-tasks are organized into a hierarchical structure with two levels, enabling efficient intent decomposition and execution. 

Next, the \textbf{Sub-task Execution Module} is activated, generating two agents, one for each sub-task. Each agent executes its assigned sub-task by issuing commands to book a specific room and adjust environmental settings based on user needs. The agents utilize the Smart Campus dataset to ensure accurate configurations. These generated commands are placed into a message queue within the \textbf{Management and Analysis Module}, which then forwards them to actuators for execution and to the environment agent for monitoring. The environment agents continuously monitors for any changes between user preferences and the Smart Campus dataset during the booking period. If changes are detected, it generates new commands to adjust the environment accordingly, ensuring real-time adaptation to user needs.\\

\subsubsection{Low-urgency Scenario:} 
\begin{figure}[!t]
\centering
\includegraphics[width=0.95\textwidth]{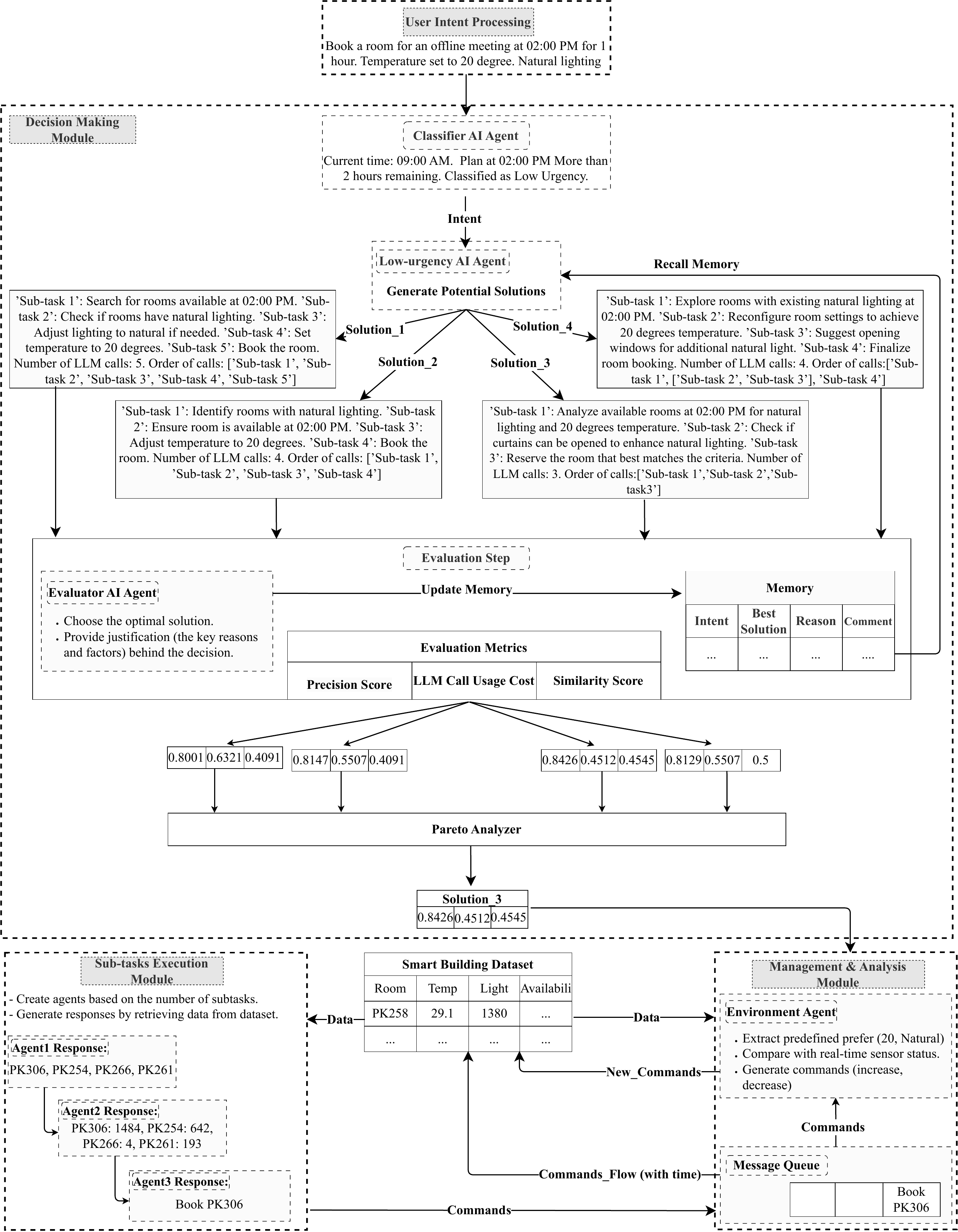}
\caption{Low-urgency Workflow within \texttt{\textbf{UserCentrix}} Framework.}\label{fig:low}
\end{figure}

Fig.~\ref{fig:low} illustrates the workflow of the framework’s building modules, beginning with the classifier agent, which determines the urgency level of a submitted user intent. In this scenario, the agent classified the intent as low urgency because the time difference between the current time and the intent time was more than two hours. It then forwarded the intent to the low-urgency agent, which is responsible for generating all possible reasoning solutions for each intent in a smart building, such as booking rooms, scheduling meals, or adjusting environmental settings with leveraging from the best solutions stored in memory with the reason and factors which impact of the selecting the solution as best as well as insights to refine future solutions. Before generating solutions, the low-urgency agent retrieves memory to check for previously stored intents. If any are found, it calculates the similarity between the current plan embeddings and the embeddings of past plans. If a highly similar intent is identified, the corresponding solutions, along with the reasons and comments provided by the evaluator agent, are injected into the prompt as hints for generating new solutions.

After generating several solutions based on different criteria, the evaluation process begins. This process involves two key components: the pareto analyzer and the evaluator agent. Pareto analyzer applies fitness functions to each solution and calculates corresponding values. The goal is to identify the solution that achieves maximum semantic similarity, maximum precision score, and minimum cost. Meanwhile, evaluator agent selects the best solution and stores it in memory along with the reason and factors influencing the decision. Once the pareto analyzer completes its evaluation, it forward the solution to the \textbf{Sub-task Execution Module}, which generates three agents based on the number of sub-tasks. Each agent executes its assigned sub-task, issuing commands to book a specific room and adjust environmental settings based on user needs using the Smart Campus dataset.
These generated commands are placed into a message queue within the \textbf{Management and Analysis Module}, which then forwards them to actuators for execution and to the environment agent for monitoring. The environment agent continuously monitors any changes between user preferences and the Smart Campus dataset during the booking period. If changes are detected, it generates new commands to adjust the environment accordingly, ensuring real-time adaptation to user needs.\\

%% file: Algorithms/algo1.tex
\begin{algorithm}[t]
\caption{\texttt{\textbf{UserCentrix}} Decision-Making Module}
\label{alg:task_solution}
\begin{algorithmic}[1]
\Require User\_Intents, $T=\{1:m\} \forall T_i = \mathcal{D}_i\exists \mathcal{D}_i\in\mathcal{D}$ and $i\in\{1:m\}$
\Ensure $\mathcal{T} = \left\{ \mathcal{T}_1, \mathcal{T}_2, \dots, \mathcal{T}_n \right\} 
\quad \text{where each} \quad \mathcal{T}_i \quad \text{is a sub-task for} \quad i \in \{1, 2, \dots, n\}.$

\For{each $T_i\in T$}
\State Classify = $\mathcal{D}_i - t$ \Comment{$t$ is current time.}
\If{(Classify $\leq 0$)}
\State $T=T-\{T_i\}$ \Comment{$T_i$ is removed from intent list.}
\ElsIf{(Classify $\geq \vartheta_1$)} \Comment{$\vartheta_1$ is Threshold.}
\State  $\mathcal{U}_i = 0$; \Comment{Low urgency level}
\State $\mathcal{A}_{\text{Low}} \longleftarrow T_i$ \Comment{Send intent to low-urgency agent} 
\State $M_{Low}\longleftarrow$RecallMemory($\mathcal{A}_{\text{Low}}$)
\State $ \mathbf{E}_{\text{new}} \longleftarrow \text{Embed\_all-MiniLM-L6-v2}(\mathcal{T}_i) $\Comment{Calculate embedding of new solution}
\State $ \mathbf{E}_{\text{past}} \longleftarrow \text{Embed\_all-MiniLM-L6-v2}(\mathcal{M}_{\text{Low}}) $ \Comment{Calculate embeddings of past solutions}
\State $ \text{Similarity}(\mathbf{E}_{\text{new}}, \mathbf{E}_{\text{past}}) \longleftarrow \frac{\mathbf{E}_{\text{new}} \cdot \mathbf{E}_{\text{past}}}{\|\mathbf{E}_{\text{new}}\| \|\mathbf{E}_{\text{past}}\|} $ \Comment{Calculate similarity between new and past solution embeddings}
\If{$\text{Similarity}(\mathbf{E}_{\text{new}}, \mathbf{E}_{\text{past}}) \leq \vartheta_2$} \Comment{$\vartheta_2 = 0.7$ in our work}
    \State Generate $n$ potential solutions for $\mathcal{T}_i$ \Comment{Generate solutions from scratch}
\Else
\State Retrieve solution $\mathcal{E}_i$ with the reason $\mathcal{R}_i$ and factors $\mathcal{F}_i$ to $\mathcal{T}_i$ by $\mathcal{A}_{\text{Low}}$ \Comment{Inject corresponding solution with the reason and factors provided by evaluator agent of this high similarity intent into the agent's prompt}
\State Generate $n$ potential solutions for $\mathcal{T}_i$ using $\mathcal{E}_i \& \mathcal{R}_i \& \mathcal{F}_i$ \Comment{Generate solutions by leveraging previous responses from the evaluator agent}
\EndIf
\For{each $\mathcal{T}_i$}
\State Calculate semantic similarity ($\mathcal{S}(\mathcal{T}_i)$): using \texttt{llama-index}
\State Precision ($\mathcal{T}_i$) $=\frac{TP}{TP + FP}$
\State $LLM Call Usage Cost (\mathcal{T}_i) = 1 - \exp\left(-\frac{N_{calls}}{N_{max}}\right)$
\State Pareto($\mathcal{T}_i$) = $\min(LLM Call Usage Cost (\mathcal{T}_i)), \max(\mathcal{S}(\mathcal{T}_i)), \max(Precision (\mathcal{T}_i))$
\State $\mathcal{E}_i , \mathcal{R}_i , \mathcal{F}_i \longleftarrow \mathcal{A}_{Evalu}(\mathcal{T}_i)$ \Comment{Evaluator agent selects the most optimal solution with giving reason and factors}
\State $\mathcal{M} \longleftarrow \mathcal{M} \cup \mathcal{E}_i, \mathcal{R}_i, \mathcal{F}_i$ \Comment{Inject evaluator agent' response into memory}
\State $\mathcal{M} \longleftarrow \mathcal{M} \cup \mathcal{T}_i $ \Comment{Inject intent into memory}
\EndFor
 \Else 
\State  $\mathcal{U}_i = 1$; \Comment{High urgency level}
\State $\mathcal{A}_{\text{High}} \longleftarrow T_i$ \Comment{Send intent to high-urgency agent} 
\State $\mathcal{A}_{\text{High}}$ generates a quick solution $\mathcal{T}_{i}$ for intent $T_i$
\EndIf

\EndFor
\State \Return $\mathcal{T}$
\end{algorithmic}
\end{algorithm}

\subsubsection*{Time complexity of Algorithm~\ref{alg:task_solution}}
Estimating the time complexity for AI-agents can be challenging, but we approach it using a step-count method based on the algorithm's structure. The time complexity of this algorithm is primarily driven by the number of user intents ($m$) and the operations performed for each intent. For each intent, embedding computation is a key operation, with a complexity of $O(d)$, where $d$ represents the embedding dimension. Determining whether an intent is high or low urgency requires constant time $O(1)$ and does not significantly impact the overall complexity. High urgency intents have constant complexity since they involve only quick solution generation without additional operations. However, low urgency intents generate $n$ sub-task solutions, and each sub-task solution performs semantic similarity calculation (using llama-index) with complexity $O(s)$, where $s$ depends on the embedding dimension of sub-task solutions. Additional operations for precision calculation, LLM Call Usage Cost computation, and Pareto optimization each require constant time $O(1)$. Thus, the time complexity for $n$ sub-task solutions can be expressed as $O(n \times s)$. Memory recall and injection operations depend on the size of memory i.e., $k$, which express $O(k)$ to the complexity. Overall, the time complexity of Algorithm~\ref{alg:task_solution} can be expressed as $O(m \times (k + d + n \times s))$.  

%% file: Algorithms/algo2.tex
\begin{algorithm}[t]
\caption{Sub-tasks Execution Module}\label{alg:sub_tasks_execution}
\begin{algorithmic}[1]
\Require $k$, Sub-tasks Solutions $\mathcal{T} = \{1:k\}$ and Dataset $D = \{1:\Delta\}$
\Ensure Command $\mathcal{C}$

\State Generate $k$  low-level agents i.e., $A = \{ A_i \mid A_i \text{ executes } \mathcal{T}_i, \forall i \in \{1, 2, \dots, k\} \}$, $\forall A_i$ handling a $\mathcal{T}_i $

\For{$\forall \mathcal{T}_i \in A_i$}
    \State Retrieve the dataset $D_i = \{\delta \in D \mid \mathbf{1}_{\text{compatible}}(\delta, \mathcal{T}_i) = 1\}$ \Comment{$\mathbf{1}_{\text{compatible}}$ give appropriate dataset}
    \State $\mathcal{C}_i \longleftarrow \text{Execute}(\mathcal{T}_i, D_i)$.
    \State $\mathcal{C} \longleftarrow \mathcal{C} \cup \{\mathcal{C}_i\}$,  $\forall i \in \{1, 2, \dots, k\}$.
\EndFor
\State Return commands $\mathcal{C}$
\end{algorithmic}
\end{algorithm}

\subsubsection*{Time complexity for Algorithm~\ref{alg:sub_tasks_execution}}
In Algorithm~\ref{alg:sub_tasks_execution}, we perform agent generation, dataset selection, intent execution, and solution aggregation. Our algorithm uses $k$ agents based on sub-task solutions, which has a linear complexity of $O(k)$ for initialization. For each agent, we need to select the most appropriate dataset from a collection of datasets (our case we assume, $\Delta$ datasets). This selection process takes $O(\Delta)$ time for each agent, as we need to evaluate the compatibility of each dataset. Once a dataset is selected, the execution step ($\text{Execute}(\mathcal{T}i, D_i)$) processes the chosen dataset. Let's denote the size of the largest dataset as $\eta$. In the worst case, the execution time for each agent would be $O(\eta)$. These operations are performed for each of the $k$ agents. The final step of aggregating solutions into $\mathcal{C}$ takes constant time $O(1)$ per agent. Therefore, the overall time complexity of Algorithm 2 can be expressed as $O(k \times (\Delta + \eta))$.

%% file: sections/Implementation.tex
\section{Implementation}\label{sec:Implementation}
In practice, we perform all experiments using two computing environments: a desktop equipped with an Intel(R) Core(TM) i5-1135G7 CPU with 16GB RAM is assumed as edge, and simultaneously on Google Colab using Intel(R) Xeon(R) CPU with 52GB RAM treated as cloud. We use the \textit{University of Oulu} as our experimental setting, leveraging data from the \textit{University of Oulu Smart Campus} Dataset~\cite{10.23729/b9adb0a2-7381-45db-b32f-7e78ae1bc9e3}. Specifically, we focus on meeting rooms equipped with \textit{Elsys ERS CO2} sensors, which provide comprehensive indoor environmental measurements, including motion, temperature and light intensity. 

These sensors are calibrated before deployment, and their data quality is extensively validated to ensure reliability and accuracy~\cite{motlagh2021monitoring}. The selected rooms include: \textit{TS501, PK258, PK265, PK306, PK254, PK266, PK261, PK253, PK267, PK262, PK309, PK268, PK263, PK308, and PK264}. In addition to the selected meeting rooms, since our goal is to enable low-level agents to identify rooms that match user preferences defined by room temperature, light status, and availability, we generated a synthetic dataset. This synthetic dataset includes room availability based on typical working hours, along with temperature and light intensity, and will be used by low-level agents for room selection and by the environment agent for ongoing environmental changes tracking.

All implemented agents in the experiments are built using \textit{LangChain}~\cite{Langchain}. We developed the memory as a custom repository type using \textit{LangChain}~\cite{Langchain}. For embeddings and similarity, we use \textit{all-MiniLM-L6-v2} model\footnote{\url{https://huggingface.co/sentence-transformers/all-MiniLM-L6-v2}} which is designed for semantic textual similarity (STS) tasks. It creates embeddings (vector representations) for sentences to capture their semantic meaning. These embeddings allow the model to compute similarity scores between texts.

To enable the evaluator agent to assess the potential solutions generated by the low-urgency agent and select the most optimal one, we employ the \textit{o1 model}\footnote{\url{https://openai.com/o1/}}, due to its advanced reasoning capabilities, ensuring that the chosen solution aligns best with the original intent objectives. We utilized \textit{Pareto dominance} with the \textit{paretoset 1.2.4} library \footnote{\url{https://pypi.org/project/paretoset/}}. For fitness functions, semantic similarity was evaluated using the \textit{LlamaIndex} framework\footnote{\url{https://docs.llamaindex.ai/en/stable/}}, while precision was assessed using the \textit{ragas} framework\footnote{\url{https://docs.ragas.io/en/latest/concepts/metrics/}}.

We selected models that allow for a comprehensive evaluation of reasoning capabilities across a diverse range of both large and small language models. LMs can play a crucial role in facilitating user-centric interactions and dynamically scaling computational resources. In our experiment, we incorporate the following language models:
\begin{itemize}

\item  \textit{Gemini 1.5 Flash} (8B), a lightweight model, developed by \textit{Google DeepMind}\footnote{\url{https://deepmind.google/technologies/gemini/}}. This model is selected due to its advanced capabilities in long-context reasoning, and optimized for low-latency performance and enhanced efficiency in agentic interactions. 

\item  \textit{GPT-4o}\footnote{\url{https://platform.openai.com/docs/models}}, a large model developed by \textit{OpenAI}. It is incorporated due to its advanced reasoning capabilities, particularly in real-time analysis, making it well-suited for real-world applications. 

\item  \textit{Claude 3.5 Sonnet}(8.03B)\footnote{\url{https://www.anthropic.com/news/claude-3-5-sonnet}}, developed by \textit{Anthropic} and known for its strong agentic capabilities. 

\item \textit{Command-r7b} (8.03B)\footnote{\url{https://cohere.com/blog/command-r7b}}, the smallest model in \textit{Cohere}’s R series with powerful agentic capabilities, is optimized for diverse use cases, including deployment on edge devices.  

\item \textit{Mistral} (7.25B)\footnote{\url{https://mistral.ai/}}, an open-source model developed by \textit{Mistral AI} with advanced reasoning capabilities and rapid inference speed.

\item \textit{IBM}’s Granite models\footnote{\url{https://www.ibm.com/granite/}}, with \textit{granite3.1-MoE} (3B) which employs a mixture-of-experts (MoE) architecture, making it particularly suitable for low-latency applications.

\end{itemize}

As there are no prior studies in the literature that address the same problem while considering the specific requirements and objectives of \texttt{\textbf{UserCentrix}} framework. Since the proposed framework integrates multiple agentic modules, each characterized by distinct techniques, objectives, and requirements, we evaluate each module independently while considering its specific requirements. Therefore, we analyze elapsed time, CPU usage, and memory utilization, together with module-specific performance metrics. In addition, we evaluate the accuracy of responses generated by different agents across modules against a baseline model, using Human-in-the-Loop assessment~\cite{WU2022364} as the reference standard to measure improvements and ensure user satisfaction. For more details:
\begin{itemize}

\item \textbf{User Intent Processing Module:} We evaluate the personal agent's responses in analyzing the user’s intent when executed on both a cloud server and an edge device. The evaluation involves elapsed time, CPU usage, and memory utilization across various models. Additionally, we assess accuracy under two conditions, when memory is empty and when it is full, by determining whether the agent retrieves relevant information from memory or operates without memory access. This assessment is based on the criteria outlined in the \textit{LangChain} primary evaluation template\footnote{\url{https://github.com/langchain-ai/langchain/blob/master/libs/langchain/langchain/evaluation/criteria/prompt.py}}. \\
\textbf{Our objective:} is to evaluate the performance of a personal agent using an appropriate model that accurately interprets user intent and retrieves relevant past information with low latency and minimal resource consumption, while maintaining high accuracy across both edge and cloud environments.

\item \textbf{Decision-Making Module:}
\begin{enumerate}
\item \textbf{Classifier Agent Performance:} We evaluate the classifier agent's responses to determine the urgency level of different intents as either \textless{}High\textgreater or \textless{}Low\textgreater when executed on both a cloud server and an edge device, measuring elapsed time, CPU usage, and memory utilization across various models. Additionally, we measure factual correctness in precision mode by comparing agents’ responses across multiple models against a human reference, using the o1 model as the evaluator.\\
\textbf{Our objective:} is to evaluate the performance of the classifier agent using an appropriate model that accurately interprets intent requirements and precisely categorizes them into high- and low-urgency levels, while maintaining minimal computational overhead and low latency.

\item \textbf{Performance of Low-urgency and High-urgency Agents:} We evaluate the performance of high-urgency and low-urgency agents in solution generation when executed on a cloud server and an edge device by measuring elapsed time, CPU usage, and memory utilization across various models.\\
\textbf{Our objective:} is to evaluate the performance of the agent using an appropriate model that accurately understands task requirements and achieves an optimal trade-off between execution speed and resource utilization in solution generation.

\item \textbf{Low-urgency Agent Performance:} We verify whether the pareto-optimal solution aligns with the preferred requirements selected by the o1 model. Additionally, we assess improvements in the agent’s responses through prompt evolution via an in-context learning loop, using evaluator agent. \\
\textbf{Our objective:} is to evaluate the performance of the low-urgency agent using an appropriate model that generates the optimal solution.
\end{enumerate}

\item \textbf{Sub-tasks Execution Module:} We evaluate the performance of low-level agents in executing sub-tasks when deployed on a cloud server and an edge device, measuring elapsed time, CPU usage, and memory utilization across various models.\\
\textbf{Our objective:} is to evaluate the performance of the low-level agents using an appropriate model that executes sub-tasks, retrieves the relevant dataset, performs negotiation, and generates appropriate commands without conflict, while minimizing time and resource consumption.

\item \textbf{Management and Analysis Module:} We evaluate the environment agent's responses in detecting changes and generating the commands when executed on a cloud server and an edge device, measuring elapsed time, CPU usage, and memory utilization across various models. Additionally, we measure factual correctness in recall mode to assess how well the environment agent’s response aligns with the human reference using o1 model as evaluator. \\
\textbf{Our objective:} is to evaluate the performance of an environment agent using an appropriate model that accurately monitors and detects changes in the environmental context that deviate from user intent, and generates updated commands to be transmitted to actuators with low latency and minimal resource consumption, while maintaining high QoE and accuracy across both edge and cloud environments.

\end{itemize}
This evaluation framework ensures a comprehensive assessment of model performance in terms of efficiency, accuracy, and decision-making quality.

We utilize the factual correctness metric from RAGAS\footnote{\url{https://docs.ragas.io/en/stable/}}, leveraging GPT-4o. This metric evaluates the factual accuracy and alignment of generated responses with a reference, ranging from 0 to 1, where higher values indicate superior performance.
The precision is calculated using the Eq.(\ref{eq3}):

\begin{equation}\label{eq3}
\text{Precision} = \frac{TP}{TP + FP}
\end{equation}

Meanwhile, the recall is calculated using the Eq.(\ref{eq4}):

\begin{equation}\label{eq4}
\text{Recall} = \frac{TP}{TP + FN}
\end{equation}

Where True Positive (TP) is number of claims in the response that are present in the reference, False Positive (FP) is number of claims in the response that are not present in the reference, and False Negative (FN) is number of claims in the reference that are not present in the response.

We select precision and recall as evaluation metrics because they align with the nature of references and responses in our framework. Where the references are generated by human.
\begin{itemize}
\item Classifier Agent: Precision is used to determine the proportion of responses that differ from the reference, distinguishing between \textit{High} and \textit{Low} variations in the output.

\item  Environment Agent: Recall is chosen to evaluate the accuracy of commands (e.g., increase or decrease), the degree of change, and other key features. Since complete coverage of the response relative to the reference is essential, recall ensures a thorough assessment of correctness and consistency.
\end{itemize}

%% file: sections/Results.tex
\section{Results Analysis}\label{sec:Results}
In this section, we present a comprehensive evaluation of the framework’s performance across various modules. The analysis aims to assess the effectiveness, efficiency, and robustness of the proposed approach through multiple evaluation criteria.

\subsection{Personal Agent Performance Evaluation}
\input{tables/table-user}
The performance evaluation of the personal agent across different models highlights significant variations in execution efficiency under edge and cloud deployment, resource utilization, and accuracy in retrieving relevant information when memory is occupied, as summarized in Table~\ref{tab:user}. Fig.~\ref{Fig21} shows that GPT-4o outperforms other models, showing the shortest elapsed time and the lowest CPU utilization across both cloud and edge devices, demonstrating its superior efficiency in processing user intents. 

Under the empty memory state, cloud deployment demonstrates greater resource efficiency, while edge devices incur higher CPU and memory utilization. GPT-4o achieves the lowest latency on both cloud (6.708 s) and edge (5.738 s), with minimal CPU usage (3.60\% cloud; 6.20\% edge). Similarly, Gemini-1.5 Flash achieves latency (10.034 s cloud; 6.801 s edge) and very low cloud CPU utilization (1.30\%). In contrast, Claude 3.5 Sonnet and Mistral exhibit longer execution times (97.228–86.828 s cloud; 369.936–381.648 s edge) and higher edge CPU consumption (56.90\% and 59.90\%, respectively), which poses challenges for deployment on resource-constrained edge devices. Memory utilization further under cloud deployment: GPT-4o (9.10\%) and Gemini-1.5 Flash (6.50\%) remain lightweight on the cloud, whereas other large models exceed 32\% cloud memory and reach over 63\% on edge devices.

When memory is occupied, the agent should retrieve relevant stored information instead of reprocessing the intent from scratch, a critical factor in enhancing user experience. Gemini-1.5 flash effectively retrieves stored information, ensuring optimal performance and minimizing redundant computation. Moreover, Claude 3.5 Sonnet, Mistral, and command-r7b fail to retrieve memory-stored information, leading to inefficiencies, increased computational overhead, and degraded performance as shown in Table~\ref{tab:user}. GPT-4o maintains the shortest latency (10.41 s cloud; 7.11 s edge) and the lowest CPU utilization (2.10\% cloud; 4.10\% edge), indicating stable performance under increased memory pressure. Gemini-1.5 Flash continues to demonstrate moderate latency and low cloud memory usage (6.60\%). Conversely, Claude 3.5 Sonnet and Mistral show high computational demands, with edge CPU utilization exceeding 52\% and memory usage remaining high ($\approx$63–66\%).

\subsection{Classifier Agent Performance Evaluation}
\input{tables/table-class}
The evaluation of the classifier agent's performance across different models demonstrates substantial variations when categorizing intents into high and low urgency levels. The results, summarized in Table~\ref{tab:class}, highlight key differences in elapsed time, CPU usage, memory consumption, and factual correctness when compared against the reference under edge and cloud deployment. Gemini-1.5 Flash records the shortest elapsed time in both environments (5.39 s on cloud vs. 2.44 s on edge), followed by GPT-4o (6.33 s on cloud vs. 8.67 s on edge), with both models achieving high accuracy (1.0). 

In contrast, command-r7b and Mistral experience slowdowns on edge devices, increasing from 119.79 s to 127.89 s and from 88.08 s to 143.50 s, achieving accuracies of 0.75 and 1.0, respectively. Claude 3.5 Sonnet shows nearly identical latency across environments (119.29 s on cloud; 118.65 s on edge) with an accuracy of 0.75. Granite3.1-MoE achieves also high latency (61.71 s cloud; 16.45 s edge) but records the lowest accuracy (0.5), limiting its reliability for precise classification.

Resource utilization increases substantially on edge devices. For instance, claude 3.5 Sonnet’s CPU usage rises from 33.30\% (cloud) to 58.50\% (edge), while command-r7b increases from 6.80\% to 62.40\%. Memory consumption shows a similar pattern: Mistral increases from 13.50\% (cloud) to 66.30\% (edge), and command-r7b from 16.00\% to 71.90\%. Even relatively efficient models such as GPT-4o demonstrate higher memory usage on edge devices (12.60\% on cloud vs. 42.20\% on edge).

GPT-4o and Gemini-1.5 Flash consistently demonstrate the shortest elapsed time and the lowest memory consumption across both cloud and edge devices, highlighting their effectiveness in performing classification with minimal computational overhead. In contrast, Claude 3.5 Sonnet, Mistral, and command-r7b demonstrate significantly higher elapsed times and increased memory utilization, particularly on edge devices. Furthermore, granite3.1-MoE's low accuracy underscores its limitations in classification where precision is critical.

\subsection{Evaluation of Low-urgency and High-urgency Agent Performance}
\input{tables/table-level}
The evaluation of low-urgency and high-urgency agents across different models highlights significant variations in elapsed time, CPU utilization, and memory consumption when generating solutions. As shown in Table~\ref{tab:level}, high-urgency agents generally execute intents faster than their low-urgency agents, which is expected given their prioritization in processing. However, this efficiency often comes at the cost of increased CPU and memory utilization, particularly on resource-constrained edge devices. Fig.~\ref{Fig41} and Fig.~\ref{Fig42} show that GPT-4o and Gemini-1.5 flash models achieve the best trade-off between execution speed and resource utilization, making them ideal for real-time solution generation, especially in high-urgency scenarios. 
For instance, Gemini-1.5 Flash records 4.33 s with 1.50\% CPU utilization on the cloud, compared to 4.83 s with 8.40\% CPU utilization on the edge. For low-urgency tasks, it achieves 10.75 s with 1.40\% CPU on the cloud versus 14.10 s with 7.60\% CPU on the edge. Similarly, GPT-4o achieves 6.17 s with 1.40\% CPU utilization on the cloud and 10.16 s with 9.20\% on the edge under high urgency. For low urgency, it records 25.93 s with 1.60\% CPU on the cloud compared to 22.22 s with 13.30\% on the edge.

In contrast, Claude 3.5 Sonnet and Mistral show higher latency and resource demands, particularly on edge devices, limiting their feasibility for real-time applications. For high-urgency tasks, Claude 3.5 Sonnet requires 74.27 s with 24\% CPU (cloud) and 514.06 s with 61.5\% CPU (edge). Meanwhile, for low-urgency tasks, it records 149.22 s with 8.3\% CPU (cloud) and 3058.60 s with 24.4\% CPU (edge). These elevated computational and memory requirements significantly constrain their practicality for real-time, latency-sensitive deployments, especially on resource-limited edge hardware. Proper model selection reduces latency under high-urgency settings, whereas under low-urgency configurations it enables more extensive solution generation while balancing execution speed and resource utilization, making the appropriate model selection critical for effective resource management in real-time solutions generation.

\subsection{Evaluation of Low-level Agent Performance and Pareto Analyzer}
\input{tables/table-pareto}
The evaluation of low-urgency agent performance highlights critical differences in reasoning-based solution generation, LM call count, execution hierarchy depth, and resource efficiency across different models. Table~\ref{tab:pareto} highlights the execution evaluation of low-level agents in terms of elapsed time, CPU utilization, and memory consumption.

Under the low-urgency setting, the agent generates multiple reasoning solutions characterized by higher LM call counts and deeper hierarchy levels as illustrated in Table~\ref{tab:pareto}. For example, Gemini-1.5 Flash generates three solutions with configurations 3–4 LM calls and hierarchy depths of three. In contrast, under the high-urgency setting, the agent prioritizes latency reduction and immediate decision-making by decreasing LM call counts and hierarchy depth. For instance, Gemini-1.5 Flash reduces to one solution with two LM calls and a depth of two. Meanwhile, GPT-4o generates five low-urgency solutions with 2–3 LM calls and hierarchy depths of 2–3, whereas under high urgency it produces a single solution with two LM calls and a depth of two, reinforcing the importance of minimizing computational overhead for real-time responsiveness. 

The trade-off analysis between semantic similarity, precision, and LM call usage cost reveals clear quantitative differences across models. Claude 3.5 Sonnet exhibits low precision under low urgency (0.2727–0.3636), despite similarity scores ranging from 0.8421 to 0.8702 and LM call usage costs of 0.5507–0.6321, making it less suitable for intents requiring structured reasoning or high accuracy as shown Fig.~\ref{Fig51c}. In contrast, GPT-4o achieves the highest precision scores, reaching 0.5263 with similarity scores up to 0.8288 and moderate LM call usage costs (0.4866–0.6321) as shown Fig.~\ref{Fig51f}. These results highlight its strong capability to generate accurate and reliable outputs.

Mistral, command-r7b, and Gemini-1.5 Flash demonstrate competitive contextual relevance, with similarity scores reaching 0.8752 (Mistral), 0.8913 (command-r7b), and 0.8704 (Gemini-1.5 Flash) as shown Fig.~\ref{Fig51d}, Fig.~\ref{Fig51b}, and Fig.~\ref{Fig51a}. Their precision values range between 0.3 and 0.4, with LM call usage costs typically at 0.6321, reflecting balanced reasoning quality and computational expense. The gray-highlighted rows in the table represent Pareto-optimal solutions. The bolded values indicate the final selections made by the evaluator agent after assessing each solution. Notably, these Pareto-optimal solutions consistently align with the evaluator selections, particularly under low-urgency settings, confirming the framework’s ability to identify resource-efficient configurations while preserving solution quality.

After executing the sub-tasks with low-level agents, we evaluate elapsed time, CPU utilization, and memory usage across different models to assess their performance under varying urgency levels. Fig.~\ref{Fig52a}, Fig.~\ref{Fig52g} show that Gemini-1.5 Flash, under low urgency, the selected configuration achieves 15.79 s on the cloud with 4.80\% CPU and 14.00\% memory, and 3.85 s on the edge with 7.00\% CPU and 36.70\% memory. Under high urgency, latency further decreases to 3.71 s, 5.60\% CPU, 14.00\% memory on cloud and 3.43 s, 15.50\% CPU, 36.30\% memory on edge. These consistently low CPU values and short execution times indicate strong suitability for low-resource and latency-sensitive environments. Fig.~\ref{Fig52f}, and Fig.~\ref{Fig52l} show that Gpt-4o is even faster in execution time with minimal CPU consumption. Under low urgency, the optimal configuration records 13.89 s on the cloud with only 2.00\% CPU and 3.60\% memory, and 9.43 s on the edge with 2.30\% CPU and 53.20\% memory. Under high urgency, performance improves to 1.97 s (cloud, 3.40\% CPU, 3.60\% memory) and 1.38 s (edge, 5.60\% CPU, 49.00\% memory). These results highlight GPT-4o’s exceptional efficiency and scalability, particularly in edge scenarios where latency is critical.

In contrast, Claude 3.5 Sonnet and Mistral are among the most resource-intensive models as shown in Fig.~\ref{Fig52c}, Fig.~\ref{Fig52i}, Fig.~\ref{Fig52d}, and Fig.~\ref{Fig52j}. Claude 3.5 Sonnet, under low urgency, requires 2034.62 s on the cloud (51.10\% CPU) and 1621.68 s on the edge (57.50\% CPU). Even under high urgency, it records 643.44 s (cloud, 50.00\% CPU) and 666.34 s (edge, 57.50\% CPU). Similarly, Mistral under low urgency consumes 491.51 s (cloud, 40.30\% CPU) and 627.37 s (edge, 57.90\% CPU), while high urgency still incurs 514.69 s and 49.8\% (cloud) and 801.52 s and 58.7\% (edge). These elevated latencies and CPU levels reflect inefficiencies in reasoning and computationally expensive and less scalable for real-time applications. command-r7b and granite3.1-MoE provide intermediate performance with lower elapsed times than Mistral or Claude 3.5 Sonnet as shown in Fig.~\ref{Fig52b}, Fig.~\ref{Fig52h}, Fig.~\ref{Fig52e}, and Fig.~\ref{Fig52k}. Command-r7b records 498.16 s (cloud, 30.00\% CPU) and 625.26 s (edge, 64.00\% CPU) under low urgency, and 676.50 s with 49.8\% (cloud) and 616.29 s and 60.5\% (edge) under high urgency. Granite3.1-MoE performs comparatively better, with 202.65 s (cloud, 37.20\% CPU) and 285.34 s (edge, 56.00\% CPU) under low urgency, and 199.45 s and 46.5\% (cloud) and 297.27 s and 56.7\%(edge) under high urgency, representing a middle-ground trade-off between latency and resource consumption. Overall, Gpt-4o and Gemini-1.5 Flash emerge as the most efficient in both speed and CPU consumption across urgency levels and deployment environments. Combined with their strong accuracy and semantic performance, these models provide the most effective balance between computational efficiency and reasoning quality, making them well-suited for structured reasoning and real-time decision-making tasks.

\subsection{Low-urgency Agent Learning Performance Evaluation}
\input{tables/table-learning}

Table~\ref{tab:learn} provides a comparative analysis of the low-urgency agent's performance using the Gpt-4o model, evaluated before and after learning across multiple metrics in a low-urgency setting on a cloud server. Key evaluated metrics include LM call count, hierarchy depth, similarity score, LM call usage cost, precision score, as well as elapsed time, CPU utilization, and memory utilization.

Before learning, Fig.~\ref{Fig6-1} shows multiple candidate solutions, with two Pareto-optimal configurations highlighted. Among them, the selected solution achieves a similarity score of 0.7937, an LM call usage cost of 0.6321, and the highest precision score of 0.5263. Another competitive configuration reaches similarity 0.8288 with lower cost (0.4866) but reduced precision (0.4211). The evaluator agent powered by the o1 model selected one of these solutions, confirming the system’s ability to identify high-quality, balanced outputs. When the selected solution was used as input for the next round (after learning), Fig.~\ref{Fig6-2} shows improved alignment between Pareto-optimal selection and evaluator choice. The selected configuration achieves a higher similarity score of 0.8421 with LM call usage cost 0.6321 and precision 0.3636. Compared to other post-learning candidates, the chosen solution reflects consistent trade-off prioritization. Overall, the post-learning results demonstrate stronger alignment between pareto-optimal analysis and evaluator selection, improved similarity performance by $\approx$6.1\%, and stable low CPU and memory utilization, indicating enhanced consistency and confidence in the agent’s decision-making process after learning.

\subsection{Environment Agent Performance Evaluation}
\input{tables/table-monitor}
The evaluation of the environment agent across different models reveals substantial differences in execution efficiency, resource utilization, and factual correctness when tracking real-time datasets and generating new commands when changes in temperature or lighting values are detected. The results presented in Table~\ref{tab:monitor} compare execution performance in terms of elapsed time, CPU utilization, memory consumption, and accuracy when benchmarked against the human commands using o1 model as an evaluator.

Fig.~\ref{Fig7} shows that GPT-4o and Gemini-1.5 flash emerge as the most efficient models, offering the best trade-off between execution speed, low resource consumption, and high accuracy, making them ideal for real-time environment monitoring and command generation. On the cloud server, Gemini-1.5 Flash achieves the lowest elapsed time (2.70 s) with moderate CPU usage (5.60\%) and minimal memory consumption (3.70\%), reaching perfect accuracy (1.0). GPT-4o follows with 8.36 s elapsed time, the lowest CPU utilization (2.50\%), low memory usage (3.80\%), and accuracy of 1.0. On the edge device, Gemini-1.5 Flash records 3.51 s with 14.40\% CPU and 35.70\% memory, while GPT-4o achieves 10.15 s, 10.80\% CPU, and 36.50\% memory—both maintaining perfect accuracy (1.0). These results confirm their suitability for real-time monitoring and command generation, ensuring framework responsiveness under constrained computational resources and without the need for human intervention.

In contrast, Claude 3.5 Sonnet, Mistral, and command-r7b exhibit significantly higher latency and CPU consumption. Claude 3.5 Sonnet requires 298.50 s (cloud) and 284.11 s (edge), with CPU usage of 45.20\% and 57.30\%, respectively, and achieves 0 accuracy, indicating unreliable command generation. Mistral records 251.78 s (cloud) and 242.45 s (edge) with CPU usage exceeding 44\%–57\%, while achieving only 0.5 accuracy. Command-r7b performs slightly better in accuracy (1.0) but still incurs substantial delays (220.23 s cloud; 192.20 s edge) and high CPU usage (35.00\% cloud; 57.60\% edge). This leads to inefficiencies in processing real-time data and potential performance bottlenecks, particularly in resource-constrained environments. Moreover, granite3.1-MoE demonstrates moderate latency (38.57 s cloud; 35.59 s edge) but low accuracy (0) and high edge memory consumption (77.30\%), showing poor command reliability, posing risks in critical applications where incorrect actuator commands could degrade system performance and reduce overall QoE.

\subsection{Key Findings:}
The experimental results demonstrate that the personal agent efficiently processes user intents, while the environment agent enables real-time monitoring and command generation across both cloud and edge deployments without human intervention. We observe that high-urgency agents prioritize rapid processing and consistently execute intents faster than low-urgency configurations. However, appropriate model selection is critical to achieving a balance between responsiveness and resource demand in low-urgency scenarios, which require generating more extensive solutions, particularly on resource-constrained edge devices. The pareto-based selection mechanism validates the framework’s capability to identify configurations that balance semantic quality and computational cost. Furthermore, the selected configurations after in-context learning demonstrate improved alignment, consistency, and confidence in decision-making, confirming the robustness of the learning and evaluation process. The findings show that choosing the appropriate model is critical, as the correct models provide the most effective balance between computational efficiency and reasoning quality, making them well-suited for structured reasoning and real-time decision-making in dynamic deployment environments.

%% file: tables/table-user.tex
\begin{table*}[]
\centering
\caption{Personal Agent Performance Evaluation.}
\label{tab:user}
\resizebox{!}{.045\paperheight}{%
\begin{tabular}{|
>{\columncolor[HTML]{FFFFFF}}l |
>{\columncolor[HTML]{FFFFFF}}c 
>{\columncolor[HTML]{FFFFFF}}c 
>{\columncolor[HTML]{FFFFFF}}c 
>{\columncolor[HTML]{FFFFFF}}c 
>{\columncolor[HTML]{FFFFFF}}c 
>{\columncolor[HTML]{FFFFFF}}c 
>{\columncolor[HTML]{FFFFFF}}c 
>{\columncolor[HTML]{FFFFFF}}c 
>{\columncolor[HTML]{FFFFFF}}c 
>{\columncolor[HTML]{FFFFFF}}c 
>{\columncolor[HTML]{FFFFFF}}c 
>{\columncolor[HTML]{FFFFFF}}c 
>{\columncolor[HTML]{FFFFFF}}c 
>{\columncolor[HTML]{FFFFFF}}c 
>{\columncolor[HTML]{FFFFFF}}c 
>{\columncolor[HTML]{FFFFFF}}c |}
\hline
\multicolumn{1}{|c|}{\cellcolor[HTML]{FFFFFF}} &
  \multicolumn{16}{c|}{\cellcolor[HTML]{FFFFFF}\textbf{Memory State}} \\ \cline{2-17} 
\multicolumn{1}{|c|}{\cellcolor[HTML]{FFFFFF}} &
  \multicolumn{7}{c|}{\cellcolor[HTML]{FFFFFF}\textbf{Empty}} &
  \multicolumn{1}{c|}{\cellcolor[HTML]{FFFFFF}} &
  \multicolumn{8}{c|}{\cellcolor[HTML]{FFFFFF}\textbf{Occupied}} \\ \cline{2-8} \cline{10-17} 
\multicolumn{1}{|c|}{\cellcolor[HTML]{FFFFFF}} &
  \multicolumn{3}{c|}{\cellcolor[HTML]{FFFFFF}\textbf{Cloud Server}} &
  \multicolumn{1}{c|}{\cellcolor[HTML]{FFFFFF}} &
  \multicolumn{3}{c|}{\cellcolor[HTML]{FFFFFF}\textbf{Edge Device}} &
  \multicolumn{1}{c|}{\cellcolor[HTML]{FFFFFF}} &
  \multicolumn{3}{c|}{\cellcolor[HTML]{FFFFFF}\textbf{Cloud Server}} &
  \multicolumn{1}{c|}{\cellcolor[HTML]{FFFFFF}} &
  \multicolumn{3}{c|}{\cellcolor[HTML]{FFFFFF}\textbf{Edge Device}} &
  \cellcolor[HTML]{FFFFFF} \\ \cline{2-4} \cline{6-8} \cline{10-12} \cline{14-16}
\multicolumn{1}{|c|}{\multirow{-4}{*}{\cellcolor[HTML]{FFFFFF}\textbf{Model Name}}} &
  \multicolumn{1}{c|}{\cellcolor[HTML]{FFFFFF}\textbf{\begin{tabular}[c]{@{}c@{}}Elapsed\\ Time\end{tabular}}} &
  \multicolumn{1}{c|}{\cellcolor[HTML]{FFFFFF}\textbf{\begin{tabular}[c]{@{}c@{}}CPU\\ Utilization\end{tabular}}} &
  \multicolumn{1}{c|}{\cellcolor[HTML]{FFFFFF}\textbf{\begin{tabular}[c]{@{}c@{}}Memory\\ Utilization\end{tabular}}} &
  \multicolumn{1}{c|}{\cellcolor[HTML]{FFFFFF}} &
  \multicolumn{1}{c|}{\cellcolor[HTML]{FFFFFF}\textbf{\begin{tabular}[c]{@{}c@{}}Elapsed\\ Time\end{tabular}}} &
  \multicolumn{1}{c|}{\cellcolor[HTML]{FFFFFF}\textbf{\begin{tabular}[c]{@{}c@{}}CPU\\ Utilization\end{tabular}}} &
  \multicolumn{1}{c|}{\cellcolor[HTML]{FFFFFF}\textbf{\begin{tabular}[c]{@{}c@{}}Memory\\ Utilization\end{tabular}}} &
  \multicolumn{1}{c|}{\cellcolor[HTML]{FFFFFF}} &
  \multicolumn{1}{c|}{\cellcolor[HTML]{FFFFFF}\textbf{\begin{tabular}[c]{@{}c@{}}Elapsed\\ Time\end{tabular}}} &
  \multicolumn{1}{c|}{\cellcolor[HTML]{FFFFFF}\textbf{\begin{tabular}[c]{@{}c@{}}CPU\\ Utilization\end{tabular}}} &
  \multicolumn{1}{c|}{\cellcolor[HTML]{FFFFFF}\textbf{\begin{tabular}[c]{@{}c@{}}Memory\\ Utilization\end{tabular}}} &
  \multicolumn{1}{c|}{\cellcolor[HTML]{FFFFFF}} &
  \multicolumn{1}{c|}{\cellcolor[HTML]{FFFFFF}\textbf{\begin{tabular}[c]{@{}c@{}}Elapsed\\ Time\end{tabular}}} &
  \multicolumn{1}{c|}{\cellcolor[HTML]{FFFFFF}\textbf{\begin{tabular}[c]{@{}c@{}}CPU\\ Utilization\end{tabular}}} &
  \multicolumn{1}{c|}{\cellcolor[HTML]{FFFFFF}\textbf{\begin{tabular}[c]{@{}c@{}}Memory\\ Utilization\end{tabular}}} &
  \multirow{-2}{*}{\cellcolor[HTML]{FFFFFF}\textbf{Accuracy}} \\ \cline{1-4} \cline{6-8} \cline{10-12} \cline{14-17} 
\textbf{Gemini-1.5 flash} &
  \multicolumn{1}{c|}{\cellcolor[HTML]{FFFFFF}10.034} &
  \multicolumn{1}{c|}{\cellcolor[HTML]{FFFFFF}1.30\%} &
  \multicolumn{1}{c|}{\cellcolor[HTML]{FFFFFF}6.50\%} &
  \multicolumn{1}{c|}{\cellcolor[HTML]{FFFFFF}} &
  \multicolumn{1}{c|}{\cellcolor[HTML]{FFFFFF}6.801} &
  \multicolumn{1}{c|}{\cellcolor[HTML]{FFFFFF}5.40\%} &
  \multicolumn{1}{c|}{\cellcolor[HTML]{FFFFFF}42.20\%} &
  \multicolumn{1}{c|}{\cellcolor[HTML]{FFFFFF}} &
  \multicolumn{1}{c|}{\cellcolor[HTML]{FFFFFF}14.018} &
  \multicolumn{1}{c|}{\cellcolor[HTML]{FFFFFF}1.30\%} &
  \multicolumn{1}{c|}{\cellcolor[HTML]{FFFFFF}6.60\%} &
  \multicolumn{1}{c|}{\cellcolor[HTML]{FFFFFF}} &
  \multicolumn{1}{c|}{\cellcolor[HTML]{FFFFFF}5.1216} &
  \multicolumn{1}{c|}{\cellcolor[HTML]{FFFFFF}21.80\%} &
  \multicolumn{1}{c|}{\cellcolor[HTML]{FFFFFF}43.90\%} &
 \cmark \\ \cline{1-4} \cline{6-8} \cline{10-12} \cline{14-17} 
\textbf{command-r7b} &
  \multicolumn{1}{c|}{\cellcolor[HTML]{FFFFFF}78.8173} &
  \multicolumn{1}{c|}{\cellcolor[HTML]{FFFFFF}14.60\%} &
  \multicolumn{1}{c|}{\cellcolor[HTML]{FFFFFF}20.80\%} &
  \multicolumn{1}{c|}{\cellcolor[HTML]{FFFFFF}} &
  \multicolumn{1}{c|}{\cellcolor[HTML]{FFFFFF}353.6496} &
  \multicolumn{1}{c|}{\cellcolor[HTML]{FFFFFF}58.10\%} &
  \multicolumn{1}{c|}{\cellcolor[HTML]{FFFFFF}67.70\%} &
  \multicolumn{1}{c|}{\cellcolor[HTML]{FFFFFF}} &
  \multicolumn{1}{c|}{\cellcolor[HTML]{FFFFFF}61.3768} &
  \multicolumn{1}{c|}{\cellcolor[HTML]{FFFFFF}26.90\%} &
  \multicolumn{1}{c|}{\cellcolor[HTML]{FFFFFF}20.70\%} &
  \multicolumn{1}{c|}{\cellcolor[HTML]{FFFFFF}} &
  \multicolumn{1}{c|}{\cellcolor[HTML]{FFFFFF}331.3826} &
  \multicolumn{1}{c|}{\cellcolor[HTML]{FFFFFF}50.80\%} &
  \multicolumn{1}{c|}{\cellcolor[HTML]{FFFFFF}67.30\%} &
  \xmark \\ \cline{1-4} \cline{6-8} \cline{10-12} \cline{14-17} 
\textbf{Claude 3.5 Sonnet} &
  \multicolumn{1}{c|}{\cellcolor[HTML]{FFFFFF}97.228} &
  \multicolumn{1}{c|}{\cellcolor[HTML]{FFFFFF}18.90\%} &
  \multicolumn{1}{c|}{\cellcolor[HTML]{FFFFFF}32.60\%} &
  \multicolumn{1}{c|}{\cellcolor[HTML]{FFFFFF}} &
  \multicolumn{1}{c|}{\cellcolor[HTML]{FFFFFF}369.9368} &
  \multicolumn{1}{c|}{\cellcolor[HTML]{FFFFFF}56.90\%} &
  \multicolumn{1}{c|}{\cellcolor[HTML]{FFFFFF}65.90\%} &
  \multicolumn{1}{c|}{\cellcolor[HTML]{FFFFFF}} &
  \multicolumn{1}{c|}{\cellcolor[HTML]{FFFFFF}86.7674} &
  \multicolumn{1}{c|}{\cellcolor[HTML]{FFFFFF}46.10\%} &
  \multicolumn{1}{c|}{\cellcolor[HTML]{FFFFFF}19.80\%} &
  \multicolumn{1}{c|}{\cellcolor[HTML]{FFFFFF}} &
  \multicolumn{1}{c|}{\cellcolor[HTML]{FFFFFF}276.9919} &
  \multicolumn{1}{c|}{\cellcolor[HTML]{FFFFFF}56.20\%} &
  \multicolumn{1}{c|}{\cellcolor[HTML]{FFFFFF}65.50\%} &
  \xmark \\ \cline{1-4} \cline{6-8} \cline{10-12} \cline{14-17} 
\textbf{Mistral} &
  \multicolumn{1}{c|}{\cellcolor[HTML]{FFFFFF}86.8284} &
  \multicolumn{1}{c|}{\cellcolor[HTML]{FFFFFF}15.80\%} &
  \multicolumn{1}{c|}{\cellcolor[HTML]{FFFFFF}18.00\%} &
  \multicolumn{1}{c|}{\cellcolor[HTML]{FFFFFF}} &
  \multicolumn{1}{c|}{\cellcolor[HTML]{FFFFFF}381.648} &
  \multicolumn{1}{c|}{\cellcolor[HTML]{FFFFFF}59.90\%} &
  \multicolumn{1}{c|}{\cellcolor[HTML]{FFFFFF}63.20\%} &
  \multicolumn{1}{c|}{\cellcolor[HTML]{FFFFFF}} &
  \multicolumn{1}{c|}{\cellcolor[HTML]{FFFFFF}65.9482} &
  \multicolumn{1}{c|}{\cellcolor[HTML]{FFFFFF}27.40\%} &
  \multicolumn{1}{c|}{\cellcolor[HTML]{FFFFFF}17.90\%} &
  \multicolumn{1}{c|}{\cellcolor[HTML]{FFFFFF}} &
  \multicolumn{1}{c|}{\cellcolor[HTML]{FFFFFF}313.3207} &
  \multicolumn{1}{c|}{\cellcolor[HTML]{FFFFFF}52.30\%} &
  \multicolumn{1}{c|}{\cellcolor[HTML]{FFFFFF}63.30\%} &
  \xmark \\ \cline{1-4} \cline{6-8} \cline{10-12} \cline{14-17} 
\textbf{granite3.1-MoE} &
  \multicolumn{1}{c|}{\cellcolor[HTML]{FFFFFF}20.6631} &
  \multicolumn{1}{c|}{\cellcolor[HTML]{FFFFFF}12.50\%} &
  \multicolumn{1}{c|}{\cellcolor[HTML]{FFFFFF}8.90\%} &
  \multicolumn{1}{c|}{\cellcolor[HTML]{FFFFFF}} &
  \multicolumn{1}{c|}{\cellcolor[HTML]{FFFFFF}43.1752} &
  \multicolumn{1}{c|}{\cellcolor[HTML]{FFFFFF}41.40\%} &
  \multicolumn{1}{c|}{\cellcolor[HTML]{FFFFFF}43.50\%} &
  \multicolumn{1}{c|}{\cellcolor[HTML]{FFFFFF}} &
  \multicolumn{1}{c|}{\cellcolor[HTML]{FFFFFF}20.6424} &
  \multicolumn{1}{c|}{\cellcolor[HTML]{FFFFFF}20.00\%} &
  \multicolumn{1}{c|}{\cellcolor[HTML]{FFFFFF}9.00\%} &
  \multicolumn{1}{c|}{\cellcolor[HTML]{FFFFFF}} &
  \multicolumn{1}{c|}{\cellcolor[HTML]{FFFFFF}61.2677} &
  \multicolumn{1}{c|}{\cellcolor[HTML]{FFFFFF}45.80\%} &
  \multicolumn{1}{c|}{\cellcolor[HTML]{FFFFFF}43.40\%} &
  \cmark \\ \cline{1-4} \cline{6-8} \cline{10-12} \cline{14-17} 
\textbf{Gpt-4o} &
  \multicolumn{1}{c|}{\cellcolor[HTML]{FFFFFF}6.708} &
  \multicolumn{1}{c|}{\cellcolor[HTML]{FFFFFF}3.60\%} &
  \multicolumn{1}{c|}{\cellcolor[HTML]{FFFFFF}9.10\%} &
  \multicolumn{1}{c|}{\multirow{-8}{*}{\cellcolor[HTML]{FFFFFF}\textbf{}}} &
  \multicolumn{1}{c|}{\cellcolor[HTML]{FFFFFF}5.738} &
  \multicolumn{1}{c|}{\cellcolor[HTML]{FFFFFF}6.20\%} &
  \multicolumn{1}{c|}{\cellcolor[HTML]{FFFFFF}41.80\%} &
  \multicolumn{1}{c|}{\multirow{-9}{*}{\cellcolor[HTML]{FFFFFF}\textbf{}}} &
  \multicolumn{1}{c|}{\cellcolor[HTML]{FFFFFF}10.4106} &
  \multicolumn{1}{c|}{\cellcolor[HTML]{FFFFFF}2.10\%} &
  \multicolumn{1}{c|}{\cellcolor[HTML]{FFFFFF}9.10\%} &
  \multicolumn{1}{c|}{\multirow{-8}{*}{\cellcolor[HTML]{FFFFFF}\textbf{}}} &
  \multicolumn{1}{c|}{\cellcolor[HTML]{FFFFFF}7.11} &
  \multicolumn{1}{c|}{\cellcolor[HTML]{FFFFFF}4.10\%} &
  \multicolumn{1}{c|}{\cellcolor[HTML]{FFFFFF}41.70\%} &
  \cmark \\ \cline{1-4} \cline{6-8} \cline{10-12} \cline{14-17} 
\end{tabular}
}
\end{table*}

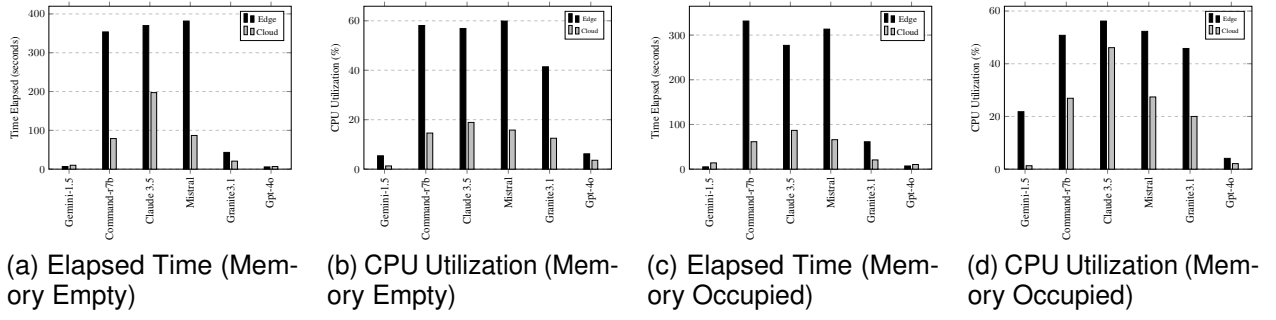
\begin{figure*}[h]
	\centering
	\subfloat[Elapsed Time (Memory Empty)]{\resizebox{38mm}{32mm}{\begin{tikzpicture}
    \begin{axis}[
    ybar,
    width=0.56\textwidth,
    height=.42\textwidth,
            bar width=.18cm,
            symbolic x coords={Gemini-1.5, Command-r7b, Claude 3.5, Mistral, Granite3.1, Gpt-4o}, xtick=data, 
            legend columns=1,
            legend pos=north east,legend style={font=\fontsize{8}{8}\selectfont},
	ylabel=Time Elapsed (seconds), ylabel style={font=\fontsize{10}{10}\selectfont},ymajorgrids=true,   grid style=dashed,ymin=0,
	 xticklabel style={rotate=90, anchor=east}, 
]
\addplot[fill=black] table[x=interval,y=ElapsedEE]{\agentperagentpe};
\addplot[fill=lightgray] table[x=interval,y=ElapsedCE]{\agentperagentpe};
\legend{Edge,Cloud}
\end{axis}  
\end{tikzpicture} }}\label{Fig21a}
\ \ \ \ 
\subfloat[CPU Utilization (Memory Empty)]{\resizebox{38mm}{32mm}{\begin{tikzpicture}
    \begin{axis}[
    ybar,
    width=0.56\textwidth,
    height=.42\textwidth,
            bar width=.18cm,
            symbolic x coords={Gemini-1.5, Command-r7b, Claude 3.5, Mistral, Granite3.1, Gpt-4o}, xtick=data, 
            legend columns=1,ymin=0,
            legend pos=north east,legend style={font=\fontsize{6}{6}\selectfont},
	ylabel=CPU Utilization (\%), ylabel style={font=\fontsize{10}{10}\selectfont},ymajorgrids=true,   grid style=dashed,
	 xticklabel style={rotate=90, anchor=east}, 
]
\addplot[fill=black] table[x=interval,y=CPUEE]{\agentperagentpe};
\addplot[fill=lightgray] table[x=interval,y=CPUCE]{\agentperagentpe};
\legend{Edge,Cloud}
\end{axis}  
\end{tikzpicture} }}\label{Fig21b}
\ \ \ \ 
\subfloat[Elapsed Time (Memory Occupied)]{\resizebox{38mm}{32mm}{\begin{tikzpicture}
    \begin{axis}[
    ybar,
    width=0.56\textwidth,
    height=.42\textwidth,
            bar width=.18cm,
            symbolic x coords={Gemini-1.5, Command-r7b, Claude 3.5, Mistral, Granite3.1, Gpt-4o}, xtick=data, 
            legend columns=1,
            legend pos=north east,legend style={font=\fontsize{8}{8}\selectfont},
	ylabel=Time Elapsed (seconds), ylabel style={font=\fontsize{10}{10}\selectfont},ymajorgrids=true,   grid style=dashed,ymin=0,
	 xticklabel style={rotate=90, anchor=east}, 
]
\addplot[fill=black] table[x=interval,y=ElapsedEO]{\agentperagentpe};
\addplot[fill=lightgray] table[x=interval,y=ElapsedCO]{\agentperagentpe};
\legend{Edge,Cloud}
\end{axis}  
\end{tikzpicture} }}\label{Fig21c}
\ \ \ \ 
\subfloat[CPU Utilization (Memory Occupied)]{\resizebox{38mm}{32mm}{\begin{tikzpicture}
    \begin{axis}[
    ybar,
    width=0.56\textwidth,
    height=.42\textwidth,
            bar width=.18cm,
            symbolic x coords={Gemini-1.5, Command-r7b, Claude 3.5, Mistral, Granite3.1, Gpt-4o}, xtick=data, 
            legend columns=1,ymin=0,
            legend pos=north east,legend style={font=\fontsize{6}{6}\selectfont},
	ylabel=CPU Utilization (\%), ylabel style={font=\fontsize{10}{10}\selectfont},ymajorgrids=true,   grid style=dashed,
	 xticklabel style={rotate=90, anchor=east}, 
]
\addplot[fill=black] table[x=interval,y=CPUEO]{\agentperagentpe};
\addplot[fill=lightgray] table[x=interval,y=CPUCO]{\agentperagentpe};
\legend{Edge,Cloud}
\end{axis}  
\end{tikzpicture} }}\label{Fig21d}\\
\caption{Personal Agent Performance Evaluation.}\label{Fig21}
\end{figure*}

%% file: tables/table-class.tex
\begin{table*}[]
\centering
\caption{Classifier Agent Performance Evaluation.}
\label{tab:class}
\resizebox{!}{.06\paperheight}{%
\begin{tabular}{|
>{\columncolor[HTML]{FFFFFF}}l |
>{\columncolor[HTML]{FFFFFF}}c 
>{\columncolor[HTML]{FFFFFF}}c 
>{\columncolor[HTML]{FFFFFF}}c |
>{\columncolor[HTML]{FFFFFF}}c |
>{\columncolor[HTML]{FFFFFF}}c 
>{\columncolor[HTML]{FFFFFF}}c 
>{\columncolor[HTML]{FFFFFF}}c 
>{\columncolor[HTML]{FFFFFF}}c |
>{\columncolor[HTML]{FFFFFF}}c |}
\cline{1-4} \cline{6-10}
\multicolumn{1}{|c|}{\cellcolor[HTML]{FFFFFF}} &
  \multicolumn{3}{c|}{\cellcolor[HTML]{FFFFFF}\textbf{Cloud Server}} &
  \cellcolor[HTML]{FFFFFF} &
  \multicolumn{4}{c|}{\cellcolor[HTML]{FFFFFF}\textbf{Edge Device}} &
  \cellcolor[HTML]{FFFFFF} \\ \cline{2-4} \cline{6-9}
\multicolumn{1}{|c|}{\multirow{-2}{*}{\cellcolor[HTML]{FFFFFF}\textbf{Model Name}}} &
  \multicolumn{1}{c|}{\cellcolor[HTML]{FFFFFF}\textbf{\begin{tabular}[c]{@{}c@{}}Elapsed\\ Time\end{tabular}}} &
  \multicolumn{1}{c|}{\cellcolor[HTML]{FFFFFF}\textbf{\begin{tabular}[c]{@{}c@{}}CPU\\ Utilization\end{tabular}}} &
  \textbf{\begin{tabular}[c]{@{}c@{}}Memory\\ Utilization\end{tabular}} &
  \cellcolor[HTML]{FFFFFF} &
  \multicolumn{1}{c|}{\cellcolor[HTML]{FFFFFF}\textbf{\begin{tabular}[c]{@{}c@{}}Elapsed\\ Time\end{tabular}}} &
  \multicolumn{1}{c|}{\cellcolor[HTML]{FFFFFF}\textbf{\begin{tabular}[c]{@{}c@{}}CPU\\ Utilization\end{tabular}}} &
  \multicolumn{1}{c|}{\cellcolor[HTML]{FFFFFF}\textbf{\begin{tabular}[c]{@{}c@{}}Memory\\ Utilization\end{tabular}}} &
  \cellcolor[HTML]{FFFFFF} &
  \multirow{-2}{*}{\cellcolor[HTML]{FFFFFF}\textbf{Accuracy}} \\ \cline{1-4} \cline{6-8} \cline{10-10} 
\textbf{Gemini-1.5 flash} &
  \multicolumn{1}{c|}{\cellcolor[HTML]{FFFFFF}5.3925} &
  \multicolumn{1}{c|}{\cellcolor[HTML]{FFFFFF}23.60\%} &
  14.40\% &
  \cellcolor[HTML]{FFFFFF} &
  \multicolumn{1}{c|}{\cellcolor[HTML]{FFFFFF}2.4466} &
  \multicolumn{1}{c|}{\cellcolor[HTML]{FFFFFF}7.90\%} &
  \multicolumn{1}{c|}{\cellcolor[HTML]{FFFFFF}31.30\%} &
  \cellcolor[HTML]{FFFFFF} &
  1 \\ \cline{1-4} \cline{6-8} \cline{10-10} 
\textbf{command-r7b} &
  \multicolumn{1}{c|}{\cellcolor[HTML]{FFFFFF}119.7979} &
  \multicolumn{1}{c|}{\cellcolor[HTML]{FFFFFF}6.80\%} &
  16.00\% &
  \cellcolor[HTML]{FFFFFF} &
  \multicolumn{1}{c|}{\cellcolor[HTML]{FFFFFF}127.8999} &
  \multicolumn{1}{c|}{\cellcolor[HTML]{FFFFFF}62.40\%} &
  \multicolumn{1}{c|}{\cellcolor[HTML]{FFFFFF}71.90\%} &
  \cellcolor[HTML]{FFFFFF} &
  0.75 \\ \cline{1-4} \cline{6-8} \cline{10-10} 
\textbf{Claude 3.5 Sonnet} &
  \multicolumn{1}{c|}{\cellcolor[HTML]{FFFFFF}119.2964} &
  \multicolumn{1}{c|}{\cellcolor[HTML]{FFFFFF}33.30\%} &
  15.20\% &
  \cellcolor[HTML]{FFFFFF} &
  \multicolumn{1}{c|}{\cellcolor[HTML]{FFFFFF}118.6518} &
  \multicolumn{1}{c|}{\cellcolor[HTML]{FFFFFF}58.50\%} &
  \multicolumn{1}{c|}{\cellcolor[HTML]{FFFFFF}70.50\%} &
  \cellcolor[HTML]{FFFFFF} &
  0.75 \\ \cline{1-4} \cline{6-8} \cline{10-10} 
\textbf{Mistral} &
  \multicolumn{1}{c|}{\cellcolor[HTML]{FFFFFF}88.08312} &
  \multicolumn{1}{c|}{\cellcolor[HTML]{FFFFFF}5.00\%} &
  13.50\% &
  \cellcolor[HTML]{FFFFFF} &
  \multicolumn{1}{c|}{\cellcolor[HTML]{FFFFFF}143.5072} &
  \multicolumn{1}{c|}{\cellcolor[HTML]{FFFFFF}61.20\%} &
  \multicolumn{1}{c|}{\cellcolor[HTML]{FFFFFF}66.30\%} &
  \cellcolor[HTML]{FFFFFF} &
  1 \\ \cline{1-4} \cline{6-8} \cline{10-10} 
\textbf{granite3.1-MoE} &
  \multicolumn{1}{c|}{\cellcolor[HTML]{FFFFFF}61.7082} &
  \multicolumn{1}{c|}{\cellcolor[HTML]{FFFFFF}23.30\%} &
  32.40\% &
  \cellcolor[HTML]{FFFFFF} &
  \multicolumn{1}{c|}{\cellcolor[HTML]{FFFFFF}16.4528} &
  \multicolumn{1}{c|}{\cellcolor[HTML]{FFFFFF}59.30\%} &
  \multicolumn{1}{c|}{\cellcolor[HTML]{FFFFFF}55.80\%} &
  \cellcolor[HTML]{FFFFFF} &
  0.5 \\ \cline{1-4} \cline{6-8} \cline{10-10} 
\textbf{Gpt-4o} &
  \multicolumn{1}{c|}{\cellcolor[HTML]{FFFFFF}6.33347} &
  \multicolumn{1}{c|}{\cellcolor[HTML]{FFFFFF}9.80\%} &
  12.60\% &
  \multirow{-11}{*}{\cellcolor[HTML]{FFFFFF}\textbf{}} &
  \multicolumn{1}{c|}{\cellcolor[HTML]{FFFFFF}8.671} &
  \multicolumn{1}{c|}{\cellcolor[HTML]{FFFFFF}8.70\%} &
  \multicolumn{1}{c|}{\cellcolor[HTML]{FFFFFF}42.20\%} &
  \multirow{-10}{*}{\cellcolor[HTML]{FFFFFF}} &
  1 \\ \cline{1-4} \cline{6-8} \cline{10-10} 
\end{tabular}

}
\end{table*}

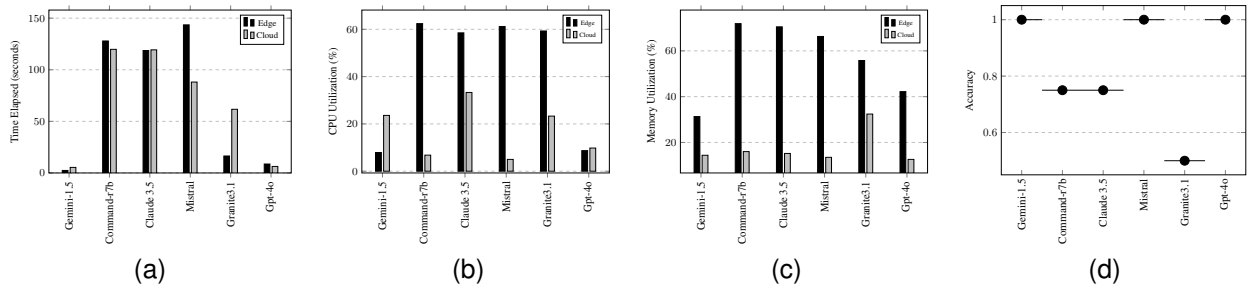
\begin{figure*}
	\centering
	\subfloat[]{\resizebox{38mm}{32mm}{\begin{tikzpicture}
    \begin{axis}[
    ybar,
    width=0.56\textwidth,
    height=.42\textwidth,
            bar width=.18cm,
            symbolic x coords={Gemini-1.5, Command-r7b, Claude 3.5, Mistral, Granite3.1, Gpt-4o}, xtick=data, 
            legend columns=1,
            legend pos=north east,legend style={font=\fontsize{8}{8}\selectfont},
	ylabel=Time Elapsed (seconds), ylabel style={font=\fontsize{10}{10}\selectfont},ymajorgrids=true,   grid style=dashed,ymin=0,
	 xticklabel style={rotate=90, anchor=east}, 
]
\addplot[fill=black] table[x=interval,y=ElapsedE]{\agentclassifier};
\addplot[fill=lightgray] table[x=interval,y=ElapsedC]{\agentclassifier};
\legend{Edge,Cloud}
\end{axis}  
\end{tikzpicture} }}\label{Fig4a}
\ \ \ \ 
\subfloat[]{\resizebox{38mm}{32mm}{\begin{tikzpicture}
    \begin{axis}[
    ybar,
    width=0.56\textwidth,
    height=.42\textwidth,
            bar width=.18cm,
            symbolic x coords={Gemini-1.5, Command-r7b, Claude 3.5, Mistral, Granite3.1, Gpt-4o}, xtick=data, 
            legend columns=1,
            legend pos=north east,legend style={font=\fontsize{6}{6}\selectfont},
	ylabel=CPU Utilization (\%), ylabel style={font=\fontsize{10}{10}\selectfont},ymajorgrids=true,   grid style=dashed,
	 xticklabel style={rotate=90, anchor=east}, 
]
\addplot[fill=black] table[x=interval,y=CPUE]{\agentclassifier};
\addplot[fill=lightgray] table[x=interval,y=CPUC]{\agentclassifier};
\legend{Edge,Cloud}
\end{axis}  
\end{tikzpicture} }}\label{Fig4b}
\ \ \ \ 
\subfloat[]{\resizebox{38mm}{32mm}{\begin{tikzpicture}
    \begin{axis}[
    ybar,
    width=0.56\textwidth,
    height=.42\textwidth,
            bar width=.18cm,
            symbolic x coords={Gemini-1.5, Command-r7b, Claude 3.5, Mistral, Granite3.1, Gpt-4o}, xtick=data, 
            legend columns=1,
            legend pos= north east,legend style={font=\fontsize{6}{6}\selectfont},
	ylabel= Memory Utilization (\%), ylabel style={font=\fontsize{10}{10}\selectfont},ymajorgrids=true,   grid style=dashed,
	 xticklabel style={rotate=90, anchor=east}, 
]
\addplot[fill=black] table[x=interval,y=MemoryE]{\agentclassifier};
\addplot[fill=lightgray] table[x=interval,y=MemoryC]{\agentclassifier};
\legend{Edge,Cloud}
\end{axis}  
\end{tikzpicture} }}\label{Fig4c}
\ \ \ \ 
\subfloat[]{\resizebox{38mm}{32mm}{\begin{tikzpicture}
    \begin{axis}[
    width=0.56\textwidth,
    height=.42\textwidth,
            bar width=.18cm,
            symbolic x coords={Gemini-1.5, Command-r7b, Claude 3.5, Mistral, Granite3.1, Gpt-4o}, xtick=data, 
            legend columns=1,
            legend pos=north east,legend style={font=\fontsize{8}{8}\selectfont},
	ylabel= Accuracy, ylabel style={font=\fontsize{10}{10}\selectfont},ymajorgrids=true,   grid style=dashed,
	 xticklabel style={rotate=90, anchor=east}, 
]
\addplot+[jump mark mid, color=black, mark size=4pt, mark options={mark color=darkgray}]  table[x=interval,y=Accuracy]{\agentclassifier};
\end{axis}  
\end{tikzpicture} }}\label{Fig4d}
\caption{Classifier Agent Performance Evaluation.}\label{Fig4}
\end{figure*}

%% file: tables/table-level.tex
\begin{table*}[]
\centering
\caption{Evaluation of Low-urgency and High-urgency Agent Performance.}
\label{tab:level}
\resizebox{!}{.06\paperheight}{%

\begin{tabular}{|l|cccccc|c|cccccc|}
\cline{1-7} \cline{9-14}
\rowcolor[HTML]{FFFFFF} 
\multicolumn{1}{|c|}{\cellcolor[HTML]{FFFFFF}} &
  \multicolumn{6}{c|}{\cellcolor[HTML]{FFFFFF}\textbf{Cloud Server}} &
  \cellcolor[HTML]{FFFFFF} &
  \multicolumn{6}{c|}{\cellcolor[HTML]{FFFFFF}\textbf{Edge Device}} \\ \cline{2-7} \cline{9-14} 
\rowcolor[HTML]{FFFFFF} 
\multicolumn{1}{|c|}{\cellcolor[HTML]{FFFFFF}} &
  \multicolumn{2}{c|}{\cellcolor[HTML]{FFFFFF}\textbf{\begin{tabular}[c]{@{}c@{}}Elapsed\\ Time\end{tabular}}} &
  \multicolumn{2}{c|}{\cellcolor[HTML]{FFFFFF}\textbf{\begin{tabular}[c]{@{}c@{}}CPU\\ Utilization\end{tabular}}} &
  \multicolumn{2}{c|}{\cellcolor[HTML]{FFFFFF}\textbf{\begin{tabular}[c]{@{}c@{}}Memory\\ Utilization\end{tabular}}} &
  \cellcolor[HTML]{FFFFFF} &
  \multicolumn{2}{c|}{\cellcolor[HTML]{FFFFFF}\textbf{\begin{tabular}[c]{@{}c@{}}Elapsed\\ Time\end{tabular}}} &
  \multicolumn{2}{c|}{\cellcolor[HTML]{FFFFFF}\textbf{\begin{tabular}[c]{@{}c@{}}CPU\\ Utilization\end{tabular}}} &
  \multicolumn{2}{c|}{\cellcolor[HTML]{FFFFFF}\textbf{\begin{tabular}[c]{@{}c@{}}Memory\\ Utilization\end{tabular}}} \\ \cline{2-7} \cline{9-14} 
\rowcolor[HTML]{FFFFFF} 
\multicolumn{1}{|c|}{\cellcolor[HTML]{FFFFFF}} &
  \multicolumn{6}{c|}{\cellcolor[HTML]{FFFFFF}\textbf{Urgency Level}} &
  \cellcolor[HTML]{FFFFFF} &
  \multicolumn{6}{c|}{\cellcolor[HTML]{FFFFFF}\textbf{Urgency Level}} \\ \cline{2-7} \cline{9-14} 
\rowcolor[HTML]{FFFFFF} 
\multicolumn{1}{|c|}{\multirow{-4}{*}{\cellcolor[HTML]{FFFFFF}\textbf{Model Name}}} &
  \multicolumn{1}{c|}{\cellcolor[HTML]{FFFFFF}\textbf{Low}} &
  \multicolumn{1}{c|}{\cellcolor[HTML]{FFFFFF}\textbf{High}} &
  \multicolumn{1}{c|}{\cellcolor[HTML]{FFFFFF}\textbf{Low}} &
  \multicolumn{1}{c|}{\cellcolor[HTML]{FFFFFF}\textbf{High}} &
  \multicolumn{1}{c|}{\cellcolor[HTML]{FFFFFF}\textbf{Low}} &
  \textbf{High} &
  \cellcolor[HTML]{FFFFFF} &
  \multicolumn{1}{c|}{\cellcolor[HTML]{FFFFFF}\textbf{Low}} &
  \multicolumn{1}{c|}{\cellcolor[HTML]{FFFFFF}\textbf{High}} &
  \multicolumn{1}{c|}{\cellcolor[HTML]{FFFFFF}\textbf{Low}} &
  \multicolumn{1}{c|}{\cellcolor[HTML]{FFFFFF}\textbf{High}} &
  \multicolumn{1}{c|}{\cellcolor[HTML]{FFFFFF}\textbf{Low}} &
  \textbf{High} \\ \cline{1-7} \cline{9-14} 
\rowcolor[HTML]{FFFFFF} 
\textbf{Gemini-1.5 flash} &
  \multicolumn{1}{c|}{\cellcolor[HTML]{FFFFFF}10.756} &
  \multicolumn{1}{c|}{\cellcolor[HTML]{FFFFFF}4.339} &
  \multicolumn{1}{c|}{\cellcolor[HTML]{FFFFFF}1.40\%} &
  \multicolumn{1}{c|}{\cellcolor[HTML]{FFFFFF}1.50\%} &
  \multicolumn{1}{c|}{\cellcolor[HTML]{FFFFFF}6.60\%} &
  6.70\% &
  \cellcolor[HTML]{FFFFFF} &
  \multicolumn{1}{c|}{\cellcolor[HTML]{FFFFFF}14.1069} &
  \multicolumn{1}{c|}{\cellcolor[HTML]{FFFFFF}4.8304} &
  \multicolumn{1}{c|}{\cellcolor[HTML]{FFFFFF}7.60\%} &
  \multicolumn{1}{c|}{\cellcolor[HTML]{FFFFFF}8.40\%} &
  \multicolumn{1}{c|}{\cellcolor[HTML]{FFFFFF}31.00\%} &
  33.40\% \\ \cline{1-7} \cline{9-14} 
\rowcolor[HTML]{FFFFFF} 
\textbf{command-r7b} &
  \multicolumn{1}{c|}{\cellcolor[HTML]{FFFFFF}94.521} &
  \multicolumn{1}{c|}{\cellcolor[HTML]{FFFFFF}53.97} &
  \multicolumn{1}{c|}{\cellcolor[HTML]{FFFFFF}40.30\%} &
  \multicolumn{1}{c|}{\cellcolor[HTML]{FFFFFF}31.80\%} &
  \multicolumn{1}{c|}{\cellcolor[HTML]{FFFFFF}19.90\%} &
  19.90\% &
  \cellcolor[HTML]{FFFFFF} &
  \multicolumn{1}{c|}{\cellcolor[HTML]{FFFFFF}3601.6761} &
  \multicolumn{1}{c|}{\cellcolor[HTML]{FFFFFF}295.8871} &
  \multicolumn{1}{c|}{\cellcolor[HTML]{FFFFFF}15.20\%} &
  \multicolumn{1}{c|}{\cellcolor[HTML]{FFFFFF}60.80\%} &
  \multicolumn{1}{c|}{\cellcolor[HTML]{FFFFFF}36.30\%} &
  71.70\% \\ \cline{1-7} \cline{9-14} 
\rowcolor[HTML]{FFFFFF} 
\textbf{Claude 3.5 Sonnet} &
  \multicolumn{1}{c|}{\cellcolor[HTML]{FFFFFF}149.222} &
  \multicolumn{1}{c|}{\cellcolor[HTML]{FFFFFF}74.278} &
  \multicolumn{1}{c|}{\cellcolor[HTML]{FFFFFF}8.30\%} &
  \multicolumn{1}{c|}{\cellcolor[HTML]{FFFFFF}24\%} &
  \multicolumn{1}{c|}{\cellcolor[HTML]{FFFFFF}18.30\%} &
  18.30\% &
  \cellcolor[HTML]{FFFFFF} &
  \multicolumn{1}{c|}{\cellcolor[HTML]{FFFFFF}3058.6028} &
  \multicolumn{1}{c|}{\cellcolor[HTML]{FFFFFF}514.0639} &
  \multicolumn{1}{c|}{\cellcolor[HTML]{FFFFFF}24.40\%} &
  \multicolumn{1}{c|}{\cellcolor[HTML]{FFFFFF}61.50\%} &
  \multicolumn{1}{c|}{\cellcolor[HTML]{FFFFFF}45.20\%} &
  70.10\% \\ \cline{1-7} \cline{9-14} 
\rowcolor[HTML]{FFFFFF} 
\textbf{Mistral} &
  \multicolumn{1}{c|}{\cellcolor[HTML]{FFFFFF}151.503} &
  \multicolumn{1}{c|}{\cellcolor[HTML]{FFFFFF}47.847} &
  \multicolumn{1}{c|}{\cellcolor[HTML]{FFFFFF}41.60\%} &
  \multicolumn{1}{c|}{\cellcolor[HTML]{FFFFFF}17.20\%} &
  \multicolumn{1}{c|}{\cellcolor[HTML]{FFFFFF}17.40\%} &
  18.10\% &
  \cellcolor[HTML]{FFFFFF} &
  \multicolumn{1}{c|}{\cellcolor[HTML]{FFFFFF}787.05659} &
  \multicolumn{1}{c|}{\cellcolor[HTML]{FFFFFF}316.2107} &
  \multicolumn{1}{c|}{\cellcolor[HTML]{FFFFFF}63.00\%} &
  \multicolumn{1}{c|}{\cellcolor[HTML]{FFFFFF}58.10\%} &
  \multicolumn{1}{c|}{\cellcolor[HTML]{FFFFFF}53.80\%} &
  53.80\% \\ \cline{1-7} \cline{9-14} 
\rowcolor[HTML]{FFFFFF} 
\textbf{granite3.1-MoE} &
  \multicolumn{1}{c|}{\cellcolor[HTML]{FFFFFF}49.2456} &
  \multicolumn{1}{c|}{\cellcolor[HTML]{FFFFFF}17.9563} &
  \multicolumn{1}{c|}{\cellcolor[HTML]{FFFFFF}36.70\%} &
  \multicolumn{1}{c|}{\cellcolor[HTML]{FFFFFF}17.60\%} &
  \multicolumn{1}{c|}{\cellcolor[HTML]{FFFFFF}9.10\%} &
  9.10\% &
  \cellcolor[HTML]{FFFFFF} &
  \multicolumn{1}{c|}{\cellcolor[HTML]{FFFFFF}80.8196} &
  \multicolumn{1}{c|}{\cellcolor[HTML]{FFFFFF}47.7772} &
  \multicolumn{1}{c|}{\cellcolor[HTML]{FFFFFF}61.90\%} &
  \multicolumn{1}{c|}{\cellcolor[HTML]{FFFFFF}60.80\%} &
  \multicolumn{1}{c|}{\cellcolor[HTML]{FFFFFF}65.50\%} &
  42.50\% \\ \cline{1-7} \cline{9-14} 
\rowcolor[HTML]{FFFFFF} 
\textbf{Gpt-4o} &
  \multicolumn{1}{c|}{\cellcolor[HTML]{FFFFFF}25.938} &
  \multicolumn{1}{c|}{\cellcolor[HTML]{FFFFFF}6.171} &
  \multicolumn{1}{c|}{\cellcolor[HTML]{FFFFFF}1.60\%} &
  \multicolumn{1}{c|}{\cellcolor[HTML]{FFFFFF}1.40\%} &
  \multicolumn{1}{c|}{\cellcolor[HTML]{FFFFFF}17.80\%} &
  6.80\% &
  \multirow{-13}{*}{\cellcolor[HTML]{FFFFFF}\textbf{}} &
  \multicolumn{1}{c|}{\cellcolor[HTML]{FFFFFF}22.227} &
  \multicolumn{1}{c|}{\cellcolor[HTML]{FFFFFF}10.1647} &
  \multicolumn{1}{c|}{\cellcolor[HTML]{FFFFFF}13.30\%} &
  \multicolumn{1}{c|}{\cellcolor[HTML]{FFFFFF}9.20\%} &
  \multicolumn{1}{c|}{\cellcolor[HTML]{FFFFFF}41.70\%} &
  41.60\% \\ \cline{1-7} \cline{9-14} 
\end{tabular}

}
\end{table*}
\begin{figure*}
	\centering
	\subfloat[Elapsed Time]{\resizebox{46mm}{42mm}{\begin{tikzpicture}
    \begin{axis}[
    ybar,
    width=0.56\textwidth,
    height=.42\textwidth,
            bar width=.18cm,
            symbolic x coords={Gemini-1.5, Command-r7b, Claude 3.5, Mistral, Granite3.1, Gpt-4o}, xtick=data, 
            legend columns=1,
            legend pos=north east,legend style={font=\fontsize{8}{8}\selectfont},
	ylabel=Time Elapsed (seconds), ylabel style={font=\fontsize{10}{10}\selectfont},ymajorgrids=true,   grid style=dashed,ymin=0,
	 xticklabel style={rotate=90, anchor=east}, 
]
\addplot[fill=black] table[x=interval,y=ElapsedEL]{\agentperparlowhigh};
\addplot[fill=lightgray] table[x=interval,y=ElapsedEH]{\agentperparlowhigh};
\legend{Low,High}
\end{axis}  
\end{tikzpicture} }}\label{Fig41a}
\ \ 
\subfloat[CPU Utilization]{\resizebox{46mm}{42mm}{\begin{tikzpicture}
    \begin{axis}[
    ybar,
    width=0.56\textwidth,
    height=.42\textwidth,
            bar width=.18cm,
            symbolic x coords={Gemini-1.5, Command-r7b, Claude 3.5, Mistral, Granite3.1, Gpt-4o}, xtick=data, 
            legend columns=1,
            legend pos=north west,legend style={font=\fontsize{6}{6}\selectfont},
	ylabel=CPU Utilization (\%), ylabel style={font=\fontsize{10}{10}\selectfont},ymajorgrids=true,   grid style=dashed,
	 xticklabel style={rotate=90, anchor=east}, 
]
\addplot[fill=black] table[x=interval,y=CPUEL]{\agentperparlowhigh};
\addplot[fill=lightgray] table[x=interval,y=CPUEH]{\agentperparlowhigh};
\legend{Low,High}
\end{axis}  
\end{tikzpicture} }}\label{Fig41b}
\ \
\subfloat[Memory Utilization]{\resizebox{46mm}{42mm}{\begin{tikzpicture}
    \begin{axis}[
    ybar,
    width=0.56\textwidth,
    height=.42\textwidth,
            bar width=.18cm,
            symbolic x coords={Gemini-1.5, Command-r7b, Claude 3.5, Mistral, Granite3.1, Gpt-4o}, xtick=data, 
            legend columns=1,
            legend pos= north west,legend style={font=\fontsize{6}{6}\selectfont},
	ylabel= Memory Utilization (\%), ylabel style={font=\fontsize{10}{10}\selectfont},ymajorgrids=true,   grid style=dashed,
	 xticklabel style={rotate=90, anchor=east}, 
]
\addplot[fill=black] table[x=interval,y=MemoryEL]{\agentperparlowhigh};
\addplot[fill=lightgray] table[x=interval,y=MemoryEH]{\agentperparlowhigh};
\legend{Low,High}
\end{axis}  
\end{tikzpicture} }}\label{Fig41c}
\caption{Evaluation of Low-urgency and High-urgency Agent Performance at the Edge.}\label{Fig41}
\end{figure*}
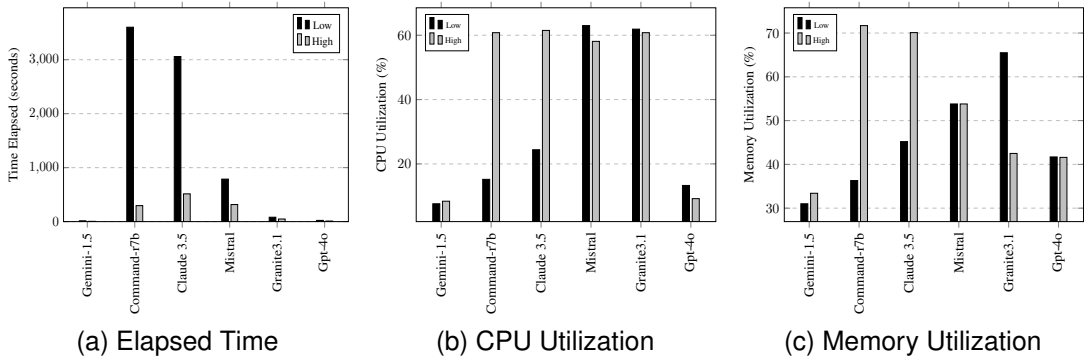
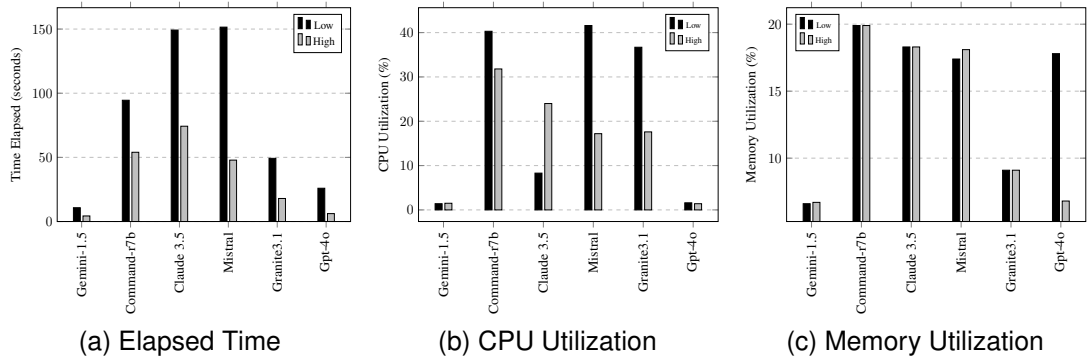
\begin{figure*}
	\centering
	\subfloat[Elapsed Time]{\resizebox{46mm}{42mm}{\begin{tikzpicture}
    \begin{axis}[
    ybar,
    width=0.56\textwidth,
    height=.42\textwidth,
            bar width=.18cm,
            symbolic x coords={Gemini-1.5, Command-r7b, Claude 3.5, Mistral, Granite3.1, Gpt-4o}, xtick=data, 
            legend columns=1,
            legend pos=north east,legend style={font=\fontsize{8}{8}\selectfont},
	ylabel=Time Elapsed (seconds), ylabel style={font=\fontsize{10}{10}\selectfont},ymajorgrids=true,   grid style=dashed,ymin=0,
	 xticklabel style={rotate=90, anchor=east}, 
]
\addplot[fill=black] table[x=interval,y=ElapsedCL]{\agentperparlowhigh};
\addplot[fill=lightgray] table[x=interval,y=ElapsedCH]{\agentperparlowhigh};
\legend{Low,High}
\end{axis}  
\end{tikzpicture} }}\label{Fig42a}
\ \ 
\subfloat[CPU Utilization]{\resizebox{46mm}{42mm}{\begin{tikzpicture}
    \begin{axis}[
    ybar,
    width=0.56\textwidth,
    height=.42\textwidth,
            bar width=.18cm,
            symbolic x coords={Gemini-1.5, Command-r7b, Claude 3.5, Mistral, Granite3.1, Gpt-4o}, xtick=data, 
            legend columns=1,
            legend pos=north east,legend style={font=\fontsize{6}{6}\selectfont},
	ylabel=CPU Utilization (\%), ylabel style={font=\fontsize{10}{10}\selectfont},ymajorgrids=true,   grid style=dashed,
	 xticklabel style={rotate=90, anchor=east}, 
]
\addplot[fill=black] table[x=interval,y=CPUCL]{\agentperparlowhigh};
\addplot[fill=lightgray] table[x=interval,y=CPUCH]{\agentperparlowhigh};
\legend{Low,High}
\end{axis}  
\end{tikzpicture} }}\label{Fig42b}
\ \ 
\subfloat[Memory Utilization]{\resizebox{46mm}{42mm}{\begin{tikzpicture}
    \begin{axis}[
    ybar,
    width=0.56\textwidth,
    height=.42\textwidth,
            bar width=.18cm,
            symbolic x coords={Gemini-1.5, Command-r7b, Claude 3.5, Mistral, Granite3.1, Gpt-4o}, xtick=data, 
            legend columns=1,
            legend pos= north west,legend style={font=\fontsize{6}{6}\selectfont},
	ylabel= Memory Utilization (\%), ylabel style={font=\fontsize{10}{10}\selectfont},ymajorgrids=true,   grid style=dashed,
	 xticklabel style={rotate=90, anchor=east}, 
]
\addplot[fill=black] table[x=interval,y=MemoryCL]{\agentperparlowhigh};
\addplot[fill=lightgray] table[x=interval,y=MemoryCH]{\agentperparlowhigh};
\legend{Low,High}
\end{axis}  
\end{tikzpicture} }}\label{Fig42c}

\caption{Evaluation of Low-urgency High-urgency Agent Performance at the Cloud.}\label{Fig42}
\end{figure*}

%% file: tables/table-pareto.tex
\begin{table*}[]
\centering
\caption{Evaluation of Low-Level Agent Performance and Pareto Analysis.}
\label{tab:pareto}
\resizebox{!}{.17\paperheight}{%
\begin{tabular}{lcccccclccccccc}
\hline
\rowcolor[HTML]{FFFFFF} 
\multicolumn{15}{c}{\cellcolor[HTML]{FFFFFF}\textbf{Experiment}} \\ \hline
\rowcolor[HTML]{FFFFFF} 
\multicolumn{4}{c|}{\cellcolor[HTML]{FFFFFF}} &
  \multicolumn{3}{c|}{\cellcolor[HTML]{FFFFFF}\textbf{Metrics}} &
  \multicolumn{1}{l|}{\cellcolor[HTML]{FFFFFF}} &
  \multicolumn{3}{c|}{\cellcolor[HTML]{FFFFFF}\textbf{Cloud Server}} &
  \multicolumn{1}{c|}{\cellcolor[HTML]{FFFFFF}} &
  \multicolumn{3}{c|}{\cellcolor[HTML]{FFFFFF}\textbf{Edge Device}} \\ \cline{1-7} \cline{9-11} \cline{13-15} 
\rowcolor[HTML]{FFFFFF} 
\multicolumn{1}{|c|}{\cellcolor[HTML]{FFFFFF}\textbf{Model Name}} &
  \multicolumn{1}{c|}{\cellcolor[HTML]{FFFFFF}\textbf{\begin{tabular}[c]{@{}c@{}}Urgency\\ Level\end{tabular}}} &
  \multicolumn{1}{c|}{\cellcolor[HTML]{FFFFFF}\textbf{\begin{tabular}[c]{@{}c@{}}LM\\ Call\\ Count\end{tabular}}} &
  \multicolumn{1}{c|}{\cellcolor[HTML]{FFFFFF}\textbf{\begin{tabular}[c]{@{}c@{}}Hierarchy\\ Depth\\ Count\end{tabular}}} &
  \multicolumn{1}{c|}{\cellcolor[HTML]{FFFFFF}\textbf{\begin{tabular}[c]{@{}c@{}}Similarity\\ Score\end{tabular}}} &
  \multicolumn{1}{c|}{\cellcolor[HTML]{FFFFFF}\textbf{\begin{tabular}[c]{@{}c@{}}LM\\ Call\\ Usage \\ Cost\end{tabular}}} &
  \multicolumn{1}{c|}{\cellcolor[HTML]{FFFFFF}\textbf{\begin{tabular}[c]{@{}c@{}}Precision\\ Score\end{tabular}}} &
  \multicolumn{1}{l|}{\cellcolor[HTML]{FFFFFF}} &
  \multicolumn{1}{c|}{\cellcolor[HTML]{FFFFFF}\textbf{\begin{tabular}[c]{@{}c@{}}Elapsed\\ Time\end{tabular}}} &
  \multicolumn{1}{c|}{\cellcolor[HTML]{FFFFFF}\textbf{\begin{tabular}[c]{@{}c@{}}CPU\\ Utilization\end{tabular}}} &
  \multicolumn{1}{c|}{\cellcolor[HTML]{FFFFFF}\textbf{\begin{tabular}[c]{@{}c@{}}Memory\\ Utilization\end{tabular}}} &
  \multicolumn{1}{c|}{\cellcolor[HTML]{FFFFFF}} &
  \multicolumn{1}{c|}{\cellcolor[HTML]{FFFFFF}\textbf{\begin{tabular}[c]{@{}c@{}}Elapsed\\ Time\end{tabular}}} &
  \multicolumn{1}{c|}{\cellcolor[HTML]{FFFFFF}\textbf{\begin{tabular}[c]{@{}c@{}}CPU\\ Utilization\end{tabular}}} &
  \multicolumn{1}{c|}{\cellcolor[HTML]{FFFFFF}\textbf{\begin{tabular}[c]{@{}c@{}}Memory\\ Utilization\end{tabular}}} \\ \cline{1-7} \cline{9-11} \cline{13-15} 
\rowcolor[HTML]{EFEFEF} 
\multicolumn{1}{|l|}{\cellcolor[HTML]{FFFFFF}} &
  \multicolumn{1}{c|}{\cellcolor[HTML]{FFFFFF}} &
  \multicolumn{1}{c|}{\cellcolor[HTML]{EFEFEF}3} &
  \multicolumn{1}{c|}{\cellcolor[HTML]{EFEFEF}3} &
  \multicolumn{1}{c|}{\cellcolor[HTML]{EFEFEF}0.8704} &
  \multicolumn{1}{c|}{\cellcolor[HTML]{EFEFEF}0.5276} &
  \multicolumn{1}{c|}{\cellcolor[HTML]{EFEFEF}0.3636} &
  \multicolumn{1}{l|}{\cellcolor[HTML]{FFFFFF}} &
  \multicolumn{1}{c|}{\cellcolor[HTML]{EFEFEF}23.87156} &
  \multicolumn{1}{c|}{\cellcolor[HTML]{EFEFEF}6.10\%} &
  \multicolumn{1}{c|}{\cellcolor[HTML]{EFEFEF}13.90\%} &
  \multicolumn{1}{c|}{\cellcolor[HTML]{FFFFFF}} &
  \multicolumn{1}{c|}{\cellcolor[HTML]{EFEFEF}3.67961} &
  \multicolumn{1}{c|}{\cellcolor[HTML]{EFEFEF}10.40\%} &
  \multicolumn{1}{c|}{\cellcolor[HTML]{EFEFEF}51.60\%} \\ \cline{3-7} \cline{9-11} \cline{13-15} 
\rowcolor[HTML]{EFEFEF} 
\multicolumn{1}{|l|}{\cellcolor[HTML]{FFFFFF}} &
  \multicolumn{1}{c|}{\cellcolor[HTML]{FFFFFF}} &
  \multicolumn{1}{c|}{\cellcolor[HTML]{EFEFEF}\textbf{4}} &
  \multicolumn{1}{c|}{\cellcolor[HTML]{EFEFEF}\textbf{3}} &
  \multicolumn{1}{c|}{\cellcolor[HTML]{EFEFEF}\textbf{0.8659}} &
  \multicolumn{1}{c|}{\cellcolor[HTML]{EFEFEF}\textbf{0.6321}} &
  \multicolumn{1}{c|}{\cellcolor[HTML]{EFEFEF}\textbf{0.4091}} &
  \multicolumn{1}{l|}{\cellcolor[HTML]{FFFFFF}} &
  \multicolumn{1}{c|}{\cellcolor[HTML]{EFEFEF}\textbf{15.7927}} &
  \multicolumn{1}{c|}{\cellcolor[HTML]{EFEFEF}\textbf{4.80\%}} &
  \multicolumn{1}{c|}{\cellcolor[HTML]{EFEFEF}\textbf{14.00\%}} &
  \multicolumn{1}{c|}{\cellcolor[HTML]{FFFFFF}} &
  \multicolumn{1}{c|}{\cellcolor[HTML]{EFEFEF}\textbf{3.8502}} &
  \multicolumn{1}{c|}{\cellcolor[HTML]{EFEFEF}\textbf{7.00\%}} &
  \multicolumn{1}{c|}{\cellcolor[HTML]{EFEFEF}\textbf{36.70\%}} \\ \cline{3-7} \cline{9-11} \cline{13-15} 
\rowcolor[HTML]{FFFFFF} 
\multicolumn{1}{|l|}{\cellcolor[HTML]{FFFFFF}} &
  \multicolumn{1}{c|}{\multirow{-3}{*}{\cellcolor[HTML]{FFFFFF}\textbf{Low}}} &
  \multicolumn{1}{c|}{\cellcolor[HTML]{FFFFFF}3} &
  \multicolumn{1}{c|}{\cellcolor[HTML]{FFFFFF}3} &
  \multicolumn{1}{c|}{\cellcolor[HTML]{FFFFFF}0.8602} &
  \multicolumn{1}{c|}{\cellcolor[HTML]{FFFFFF}0.5276} &
  \multicolumn{1}{c|}{\cellcolor[HTML]{FFFFFF}0.4091} &
  \multicolumn{1}{l|}{\cellcolor[HTML]{FFFFFF}} &
  \multicolumn{1}{c|}{\cellcolor[HTML]{FFFFFF}17.1799} &
  \multicolumn{1}{c|}{\cellcolor[HTML]{FFFFFF}4.60\%} &
  \multicolumn{1}{c|}{\cellcolor[HTML]{FFFFFF}14.00\%} &
  \multicolumn{1}{c|}{\cellcolor[HTML]{FFFFFF}} &
  \multicolumn{1}{c|}{\cellcolor[HTML]{FFFFFF}3.5883} &
  \multicolumn{1}{c|}{\cellcolor[HTML]{FFFFFF}8.80\%} &
  \multicolumn{1}{c|}{\cellcolor[HTML]{FFFFFF}36.40\%} \\ \cline{2-7} \cline{9-11} \cline{13-15} 
\rowcolor[HTML]{FFFFFF} 
\multicolumn{1}{|l|}{\multirow{-4}{*}{\cellcolor[HTML]{FFFFFF}\textbf{Gemini-1.5 flash}}} &
  \multicolumn{1}{c|}{\cellcolor[HTML]{FFFFFF}\textbf{High}} &
  \multicolumn{1}{c|}{\cellcolor[HTML]{FFFFFF}2} &
  \multicolumn{1}{c|}{\cellcolor[HTML]{FFFFFF}2} &
  \multicolumn{1}{c|}{\cellcolor[HTML]{FFFFFF}0.9039} &
  \multicolumn{1}{c|}{\cellcolor[HTML]{FFFFFF}-} &
  \multicolumn{1}{c|}{\cellcolor[HTML]{FFFFFF}0.3636} &
  \multicolumn{1}{l|}{\cellcolor[HTML]{FFFFFF}} &
  \multicolumn{1}{c|}{\cellcolor[HTML]{FFFFFF}3.7146} &
  \multicolumn{1}{c|}{\cellcolor[HTML]{FFFFFF}5.60\%} &
  \multicolumn{1}{c|}{\cellcolor[HTML]{FFFFFF}14.00\%} &
  \multicolumn{1}{c|}{\cellcolor[HTML]{FFFFFF}} &
  \multicolumn{1}{c|}{\cellcolor[HTML]{FFFFFF}3.439471} &
  \multicolumn{1}{c|}{\cellcolor[HTML]{FFFFFF}15.50\%} &
  \multicolumn{1}{c|}{\cellcolor[HTML]{FFFFFF}36.30\%} \\ \cline{1-7} \cline{9-11} \cline{13-15} 
\rowcolor[HTML]{FFFFFF} 
\multicolumn{6}{l}{\cellcolor[HTML]{FFFFFF}\textbf{}} &
   &
  \multicolumn{1}{l|}{\cellcolor[HTML]{FFFFFF}} &
   &
   &
   &
  \multicolumn{1}{c|}{\cellcolor[HTML]{FFFFFF}} &
   &
   &
   \\ \cline{1-7} \cline{9-11} \cline{13-15} 
\rowcolor[HTML]{EFEFEF} 
\multicolumn{1}{|l|}{\cellcolor[HTML]{FFFFFF}} &
  \multicolumn{1}{c|}{\cellcolor[HTML]{FFFFFF}} &
  \multicolumn{1}{c|}{\cellcolor[HTML]{EFEFEF}\textbf{2}} &
  \multicolumn{1}{c|}{\cellcolor[HTML]{EFEFEF}\textbf{2}} &
  \multicolumn{1}{c|}{\cellcolor[HTML]{EFEFEF}\textbf{0.8913}} &
  \multicolumn{1}{c|}{\cellcolor[HTML]{EFEFEF}\textbf{0.6321}} &
  \multicolumn{1}{c|}{\cellcolor[HTML]{EFEFEF}\textbf{0.4545}} &
  \multicolumn{1}{l|}{\cellcolor[HTML]{FFFFFF}} &
  \multicolumn{1}{c|}{\cellcolor[HTML]{EFEFEF}\textbf{498.1659}} &
  \multicolumn{1}{c|}{\cellcolor[HTML]{EFEFEF}\textbf{30.00\%}} &
  \multicolumn{1}{c|}{\cellcolor[HTML]{EFEFEF}\textbf{15.60\%}} &
  \multicolumn{1}{c|}{\cellcolor[HTML]{FFFFFF}} &
  \multicolumn{1}{c|}{\cellcolor[HTML]{EFEFEF}\textbf{625.2602}} &
  \multicolumn{1}{c|}{\cellcolor[HTML]{EFEFEF}\textbf{64.00\%}} &
  \multicolumn{1}{c|}{\cellcolor[HTML]{EFEFEF}\textbf{73.10\%}} \\ \cline{3-7} \cline{9-11} \cline{13-15} 
\rowcolor[HTML]{FFFFFF} 
\multicolumn{1}{|l|}{\cellcolor[HTML]{FFFFFF}} &
  \multicolumn{1}{c|}{\multirow{-2}{*}{\cellcolor[HTML]{FFFFFF}\textbf{Low}}} &
  \multicolumn{1}{c|}{\cellcolor[HTML]{FFFFFF}2} &
  \multicolumn{1}{c|}{\cellcolor[HTML]{FFFFFF}2} &
  \multicolumn{1}{c|}{\cellcolor[HTML]{FFFFFF}0.8860} &
  \multicolumn{1}{c|}{\cellcolor[HTML]{FFFFFF}0.6321} &
  \multicolumn{1}{c|}{\cellcolor[HTML]{FFFFFF}0.2727} &
  \multicolumn{1}{l|}{\cellcolor[HTML]{FFFFFF}} &
  \multicolumn{1}{c|}{\cellcolor[HTML]{FFFFFF}492.8408} &
  \multicolumn{1}{c|}{\cellcolor[HTML]{FFFFFF}49.10\%} &
  \multicolumn{1}{c|}{\cellcolor[HTML]{FFFFFF}15.70\%} &
  \multicolumn{1}{c|}{\cellcolor[HTML]{FFFFFF}} &
  \multicolumn{1}{c|}{\cellcolor[HTML]{FFFFFF}843.0399} &
  \multicolumn{1}{c|}{\cellcolor[HTML]{FFFFFF}60.80\%} &
  \multicolumn{1}{c|}{\cellcolor[HTML]{FFFFFF}72.00\%} \\ \cline{2-7} \cline{9-11} \cline{13-15} 
\rowcolor[HTML]{FFFFFF} 
\multicolumn{1}{|l|}{\multirow{-3}{*}{\cellcolor[HTML]{FFFFFF}\textbf{command-r7b}}} &
  \multicolumn{1}{c|}{\cellcolor[HTML]{FFFFFF}\textbf{High}} &
  \multicolumn{1}{c|}{\cellcolor[HTML]{FFFFFF}2} &
  \multicolumn{1}{c|}{\cellcolor[HTML]{FFFFFF}2} &
  \multicolumn{1}{c|}{\cellcolor[HTML]{FFFFFF}0.9183} &
  \multicolumn{1}{c|}{\cellcolor[HTML]{FFFFFF}-} &
  \multicolumn{1}{c|}{\cellcolor[HTML]{FFFFFF}0.5455} &
  \multicolumn{1}{l|}{\cellcolor[HTML]{FFFFFF}} &
  \multicolumn{1}{c|}{\cellcolor[HTML]{FFFFFF}676.5058} &
  \multicolumn{1}{c|}{\cellcolor[HTML]{FFFFFF}49.80\%} &
  \multicolumn{1}{c|}{\cellcolor[HTML]{FFFFFF}15.60\%} &
  \multicolumn{1}{c|}{\cellcolor[HTML]{FFFFFF}} &
  \multicolumn{1}{c|}{\cellcolor[HTML]{FFFFFF}616.2926} &
  \multicolumn{1}{c|}{\cellcolor[HTML]{FFFFFF}60.50\%} &
  \multicolumn{1}{c|}{\cellcolor[HTML]{FFFFFF}67.60\%} \\ \cline{1-7} \cline{9-11} \cline{13-15} 
\rowcolor[HTML]{FFFFFF} 
\multicolumn{7}{l}{\cellcolor[HTML]{FFFFFF}\textbf{}} &
  \multicolumn{1}{l|}{\cellcolor[HTML]{FFFFFF}} &
  \multicolumn{3}{c}{\cellcolor[HTML]{FFFFFF}} &
  \multicolumn{1}{c|}{\cellcolor[HTML]{FFFFFF}} &
  \multicolumn{3}{c}{\cellcolor[HTML]{FFFFFF}} \\ \cline{1-7} \cline{9-11} \cline{13-15} 
\rowcolor[HTML]{FFFFFF} 
\multicolumn{1}{|l|}{\cellcolor[HTML]{FFFFFF}} &
  \multicolumn{1}{c|}{\cellcolor[HTML]{FFFFFF}} &
  \multicolumn{1}{c|}{\cellcolor[HTML]{FFFFFF}5} &
  \multicolumn{1}{c|}{\cellcolor[HTML]{FFFFFF}5} &
  \multicolumn{1}{c|}{\cellcolor[HTML]{FFFFFF}0.8702} &
  \multicolumn{1}{c|}{\cellcolor[HTML]{FFFFFF}0.6321} &
  \multicolumn{1}{c|}{\cellcolor[HTML]{FFFFFF}0.3636} &
  \multicolumn{1}{l|}{\cellcolor[HTML]{FFFFFF}} &
  \multicolumn{1}{c|}{\cellcolor[HTML]{FFFFFF}1689.107} &
  \multicolumn{1}{c|}{\cellcolor[HTML]{FFFFFF}42.70\%} &
  \multicolumn{1}{c|}{\cellcolor[HTML]{FFFFFF}15.30\%} &
  \multicolumn{1}{c|}{\cellcolor[HTML]{FFFFFF}} &
  \multicolumn{1}{c|}{\cellcolor[HTML]{FFFFFF}2265.092} &
  \multicolumn{1}{c|}{\cellcolor[HTML]{FFFFFF}59.10\%} &
  \multicolumn{1}{c|}{\cellcolor[HTML]{FFFFFF}67.70\%} \\ \cline{3-7} \cline{9-11} \cline{13-15} 
\rowcolor[HTML]{EFEFEF} 
\multicolumn{1}{|l|}{\cellcolor[HTML]{FFFFFF}} &
  \multicolumn{1}{c|}{\cellcolor[HTML]{FFFFFF}} &
  \multicolumn{1}{c|}{\cellcolor[HTML]{EFEFEF}\textbf{4}} &
  \multicolumn{1}{c|}{\cellcolor[HTML]{EFEFEF}\textbf{4}} &
  \multicolumn{1}{c|}{\cellcolor[HTML]{EFEFEF}\textbf{0.8488}} &
  \multicolumn{1}{c|}{\cellcolor[HTML]{EFEFEF}\textbf{0.5507}} &
  \multicolumn{1}{c|}{\cellcolor[HTML]{EFEFEF}\textbf{0.2727}} &
  \multicolumn{1}{l|}{\cellcolor[HTML]{FFFFFF}} &
  \multicolumn{1}{c|}{\cellcolor[HTML]{EFEFEF}\textbf{2034.623}} &
  \multicolumn{1}{c|}{\cellcolor[HTML]{EFEFEF}\textbf{51.10\%}} &
  \multicolumn{1}{c|}{\cellcolor[HTML]{EFEFEF}\textbf{14.90\%}} &
  \multicolumn{1}{c|}{\cellcolor[HTML]{FFFFFF}} &
  \multicolumn{1}{c|}{\cellcolor[HTML]{EFEFEF}\textbf{1621.683}} &
  \multicolumn{1}{c|}{\cellcolor[HTML]{EFEFEF}\textbf{57.50\%}} &
  \multicolumn{1}{c|}{\cellcolor[HTML]{EFEFEF}\textbf{68.00\%}} \\ \cline{3-7} \cline{9-11} \cline{13-15} 
\rowcolor[HTML]{FFFFFF} 
\multicolumn{1}{|l|}{\cellcolor[HTML]{FFFFFF}} &
  \multicolumn{1}{c|}{\multirow{-3}{*}{\cellcolor[HTML]{FFFFFF}\textbf{Low}}} &
  \multicolumn{1}{c|}{\cellcolor[HTML]{FFFFFF}4} &
  \multicolumn{1}{c|}{\cellcolor[HTML]{FFFFFF}4} &
  \multicolumn{1}{c|}{\cellcolor[HTML]{FFFFFF}0.8421} &
  \multicolumn{1}{c|}{\cellcolor[HTML]{FFFFFF}0.5507} &
  \multicolumn{1}{c|}{\cellcolor[HTML]{FFFFFF}0.2727} &
  \multicolumn{1}{l|}{\cellcolor[HTML]{FFFFFF}} &
  \multicolumn{1}{c|}{\cellcolor[HTML]{FFFFFF}1569.989} &
  \multicolumn{1}{c|}{\cellcolor[HTML]{FFFFFF}52.60\%} &
  \multicolumn{1}{c|}{\cellcolor[HTML]{FFFFFF}14.90\%} &
  \multicolumn{1}{c|}{\cellcolor[HTML]{FFFFFF}} &
  \multicolumn{1}{c|}{\cellcolor[HTML]{FFFFFF}1869.202} &
  \multicolumn{1}{c|}{\cellcolor[HTML]{FFFFFF}57.80\%} &
  \multicolumn{1}{c|}{\cellcolor[HTML]{FFFFFF}69.10\%} \\ \cline{2-7} \cline{9-11} \cline{13-15} 
\rowcolor[HTML]{FFFFFF} 
\multicolumn{1}{|l|}{\multirow{-4}{*}{\cellcolor[HTML]{FFFFFF}\textbf{Claude 3.5 Sonnet}}} &
  \multicolumn{1}{c|}{\cellcolor[HTML]{FFFFFF}\textbf{High}} &
  \multicolumn{1}{c|}{\cellcolor[HTML]{FFFFFF}2} &
  \multicolumn{1}{c|}{\cellcolor[HTML]{FFFFFF}2} &
  \multicolumn{1}{c|}{\cellcolor[HTML]{FFFFFF}0.8809} &
  \multicolumn{1}{c|}{\cellcolor[HTML]{FFFFFF}-} &
  \multicolumn{1}{c|}{\cellcolor[HTML]{FFFFFF}0.3636} &
  \multicolumn{1}{l|}{\cellcolor[HTML]{FFFFFF}} &
  \multicolumn{1}{c|}{\cellcolor[HTML]{FFFFFF}643.4445} &
  \multicolumn{1}{c|}{\cellcolor[HTML]{FFFFFF}50.00\%} &
  \multicolumn{1}{c|}{\cellcolor[HTML]{FFFFFF}14.40\%} &
  \multicolumn{1}{c|}{\cellcolor[HTML]{FFFFFF}} &
  \multicolumn{1}{c|}{\cellcolor[HTML]{FFFFFF}666.3432} &
  \multicolumn{1}{c|}{\cellcolor[HTML]{FFFFFF}57.50\%} &
  \multicolumn{1}{c|}{\cellcolor[HTML]{FFFFFF}68.90\%} \\ \cline{1-7} \cline{9-11} \cline{13-15} 
\rowcolor[HTML]{FFFFFF} 
\multicolumn{7}{l}{\cellcolor[HTML]{FFFFFF}\textbf{}} &
  \multicolumn{1}{l|}{\cellcolor[HTML]{FFFFFF}} &
  \multicolumn{3}{c}{\cellcolor[HTML]{FFFFFF}} &
  \multicolumn{1}{c|}{\cellcolor[HTML]{FFFFFF}} &
  \multicolumn{3}{c}{\cellcolor[HTML]{FFFFFF}} \\ \cline{1-7} \cline{9-11} \cline{13-15} 
\rowcolor[HTML]{FFFFFF} 
\multicolumn{1}{|l|}{\cellcolor[HTML]{FFFFFF}} &
  \multicolumn{1}{c|}{\cellcolor[HTML]{FFFFFF}} &
  \multicolumn{1}{c|}{\cellcolor[HTML]{FFFFFF}3} &
  \multicolumn{1}{c|}{\cellcolor[HTML]{FFFFFF}3} &
  \multicolumn{1}{c|}{\cellcolor[HTML]{FFFFFF}0.8733} &
  \multicolumn{1}{c|}{\cellcolor[HTML]{FFFFFF}0.6321} &
  \multicolumn{1}{c|}{\cellcolor[HTML]{FFFFFF}0.4091} &
  \multicolumn{1}{l|}{\cellcolor[HTML]{FFFFFF}} &
  \multicolumn{1}{c|}{\cellcolor[HTML]{FFFFFF}802.7946} &
  \multicolumn{1}{c|}{\cellcolor[HTML]{FFFFFF}41.70\%} &
  \multicolumn{1}{c|}{\cellcolor[HTML]{FFFFFF}14.20\%} &
  \multicolumn{1}{c|}{\cellcolor[HTML]{FFFFFF}} &
  \multicolumn{1}{c|}{\cellcolor[HTML]{FFFFFF}949.0744} &
  \multicolumn{1}{c|}{\cellcolor[HTML]{FFFFFF}57.60\%} &
  \multicolumn{1}{c|}{\cellcolor[HTML]{FFFFFF}66.70\%} \\ \cline{3-7} \cline{9-11} \cline{13-15} 
\rowcolor[HTML]{FFFFFF} 
\multicolumn{1}{|l|}{\cellcolor[HTML]{FFFFFF}} &
  \multicolumn{1}{c|}{\cellcolor[HTML]{FFFFFF}} &
  \multicolumn{1}{c|}{\cellcolor[HTML]{FFFFFF}3} &
  \multicolumn{1}{c|}{\cellcolor[HTML]{FFFFFF}3} &
  \multicolumn{1}{c|}{\cellcolor[HTML]{FFFFFF}0.8624} &
  \multicolumn{1}{c|}{\cellcolor[HTML]{FFFFFF}0.6321} &
  \multicolumn{1}{c|}{\cellcolor[HTML]{FFFFFF}0.3636} &
  \multicolumn{1}{l|}{\cellcolor[HTML]{FFFFFF}} &
  \multicolumn{1}{c|}{\cellcolor[HTML]{FFFFFF}786.7635} &
  \multicolumn{1}{c|}{\cellcolor[HTML]{FFFFFF}51.60\%} &
  \multicolumn{1}{c|}{\cellcolor[HTML]{FFFFFF}14.00\%} &
  \multicolumn{1}{c|}{\cellcolor[HTML]{FFFFFF}} &
  \multicolumn{1}{c|}{\cellcolor[HTML]{FFFFFF}946.0731} &
  \multicolumn{1}{c|}{\cellcolor[HTML]{FFFFFF}57.50\%} &
  \multicolumn{1}{c|}{\cellcolor[HTML]{FFFFFF}66.80\%} \\ \cline{3-7} \cline{9-11} \cline{13-15} 
\rowcolor[HTML]{EFEFEF} 
\multicolumn{1}{|l|}{\cellcolor[HTML]{FFFFFF}} &
  \multicolumn{1}{c|}{\multirow{-3}{*}{\cellcolor[HTML]{FFFFFF}\textbf{Low}}} &
  \multicolumn{1}{c|}{\cellcolor[HTML]{EFEFEF}\textbf{2}} &
  \multicolumn{1}{c|}{\cellcolor[HTML]{EFEFEF}\textbf{2}} &
  \multicolumn{1}{c|}{\cellcolor[HTML]{EFEFEF}\textbf{0.8752}} &
  \multicolumn{1}{c|}{\cellcolor[HTML]{EFEFEF}\textbf{0.4866}} &
  \multicolumn{1}{c|}{\cellcolor[HTML]{EFEFEF}\textbf{0.3636}} &
  \multicolumn{1}{l|}{\cellcolor[HTML]{FFFFFF}} &
  \multicolumn{1}{c|}{\cellcolor[HTML]{EFEFEF}\textbf{491.5169}} &
  \multicolumn{1}{c|}{\cellcolor[HTML]{EFEFEF}\textbf{40.30\%}} &
  \multicolumn{1}{c|}{\cellcolor[HTML]{EFEFEF}\textbf{14.10\%}} &
  \multicolumn{1}{c|}{\cellcolor[HTML]{FFFFFF}} &
  \multicolumn{1}{c|}{\cellcolor[HTML]{EFEFEF}\textbf{627.3577}} &
  \multicolumn{1}{c|}{\cellcolor[HTML]{EFEFEF}\textbf{57.90\%}} &
  \multicolumn{1}{c|}{\cellcolor[HTML]{EFEFEF}\textbf{63.00\%}} \\ \cline{2-7} \cline{9-11} \cline{13-15} 
\rowcolor[HTML]{FFFFFF} 
\multicolumn{1}{|l|}{\multirow{-4}{*}{\cellcolor[HTML]{FFFFFF}\textbf{Mistral}}} &
  \multicolumn{1}{c|}{\cellcolor[HTML]{FFFFFF}\textbf{High}} &
  \multicolumn{1}{c|}{\cellcolor[HTML]{FFFFFF}2} &
  \multicolumn{1}{c|}{\cellcolor[HTML]{FFFFFF}2} &
  \multicolumn{1}{c|}{\cellcolor[HTML]{FFFFFF}0.8993} &
  \multicolumn{1}{c|}{\cellcolor[HTML]{FFFFFF}-} &
  \multicolumn{1}{c|}{\cellcolor[HTML]{FFFFFF}0.4091} &
  \multicolumn{1}{l|}{\cellcolor[HTML]{FFFFFF}} &
  \multicolumn{1}{c|}{\cellcolor[HTML]{FFFFFF}514.6962} &
  \multicolumn{1}{c|}{\cellcolor[HTML]{FFFFFF}49.80\%} &
  \multicolumn{1}{c|}{\cellcolor[HTML]{FFFFFF}14.00\%} &
  \multicolumn{1}{c|}{\cellcolor[HTML]{FFFFFF}} &
  \multicolumn{1}{c|}{\cellcolor[HTML]{FFFFFF}801.5293} &
  \multicolumn{1}{c|}{\cellcolor[HTML]{FFFFFF}58.70\%} &
  \multicolumn{1}{c|}{\cellcolor[HTML]{FFFFFF}64.60\%} \\ \cline{1-7} \cline{9-11} \cline{13-15} 
\rowcolor[HTML]{FFFFFF} 
\multicolumn{7}{l}{\cellcolor[HTML]{FFFFFF}\textbf{}} &
  \multicolumn{1}{l|}{\cellcolor[HTML]{FFFFFF}} &
  \multicolumn{3}{c}{\cellcolor[HTML]{FFFFFF}} &
  \multicolumn{1}{c|}{\cellcolor[HTML]{FFFFFF}} &
  \multicolumn{3}{c}{\cellcolor[HTML]{FFFFFF}} \\ \cline{1-7} \cline{9-11} \cline{13-15} 
\rowcolor[HTML]{EFEFEF} 
\multicolumn{1}{|l|}{\cellcolor[HTML]{FFFFFF}} &
  \multicolumn{1}{c|}{\cellcolor[HTML]{FFFFFF}} &
  \multicolumn{1}{c|}{\cellcolor[HTML]{EFEFEF}\textbf{3}} &
  \multicolumn{1}{c|}{\cellcolor[HTML]{EFEFEF}\textbf{3}} &
  \multicolumn{1}{c|}{\cellcolor[HTML]{EFEFEF}\textbf{0.8282}} &
  \multicolumn{1}{c|}{\cellcolor[HTML]{EFEFEF}\textbf{0.6321}} &
  \multicolumn{1}{c|}{\cellcolor[HTML]{EFEFEF}\textbf{0.3636}} &
  \multicolumn{1}{l|}{\cellcolor[HTML]{FFFFFF}} &
  \multicolumn{1}{c|}{\cellcolor[HTML]{EFEFEF}\textbf{202.6593}} &
  \multicolumn{1}{c|}{\cellcolor[HTML]{EFEFEF}\textbf{37.20\%}} &
  \multicolumn{1}{c|}{\cellcolor[HTML]{EFEFEF}\textbf{9.20\%}} &
  \multicolumn{1}{c|}{\cellcolor[HTML]{FFFFFF}} &
  \multicolumn{1}{c|}{\cellcolor[HTML]{EFEFEF}\textbf{285.3483}} &
  \multicolumn{1}{c|}{\cellcolor[HTML]{EFEFEF}\textbf{56.00\%}} &
  \multicolumn{1}{c|}{\cellcolor[HTML]{EFEFEF}\textbf{51.60\%}} \\ \cline{3-7} \cline{9-11} \cline{13-15} 
\rowcolor[HTML]{FFFFFF} 
\multicolumn{1}{|l|}{\cellcolor[HTML]{FFFFFF}} &
  \multicolumn{1}{c|}{\cellcolor[HTML]{FFFFFF}} &
  \multicolumn{1}{c|}{\cellcolor[HTML]{FFFFFF}3} &
  \multicolumn{1}{c|}{\cellcolor[HTML]{FFFFFF}3} &
  \multicolumn{1}{c|}{\cellcolor[HTML]{FFFFFF}0.8270} &
  \multicolumn{1}{c|}{\cellcolor[HTML]{FFFFFF}0.6321} &
  \multicolumn{1}{c|}{\cellcolor[HTML]{FFFFFF}0.3182} &
  \multicolumn{1}{l|}{\cellcolor[HTML]{FFFFFF}} &
  \multicolumn{1}{c|}{\cellcolor[HTML]{FFFFFF}179.3239} &
  \multicolumn{1}{c|}{\cellcolor[HTML]{FFFFFF}43.60\%} &
  \multicolumn{1}{c|}{\cellcolor[HTML]{FFFFFF}9.20\%} &
  \multicolumn{1}{c|}{\cellcolor[HTML]{FFFFFF}} &
  \multicolumn{1}{c|}{\cellcolor[HTML]{FFFFFF}277.4193} &
  \multicolumn{1}{c|}{\cellcolor[HTML]{FFFFFF}55.70\%} &
  \multicolumn{1}{c|}{\cellcolor[HTML]{FFFFFF}51.80\%} \\ \cline{3-7} \cline{9-11} \cline{13-15} 
\rowcolor[HTML]{FFFFFF} 
\multicolumn{1}{|l|}{\cellcolor[HTML]{FFFFFF}} &
  \multicolumn{1}{c|}{\cellcolor[HTML]{FFFFFF}} &
  \multicolumn{1}{c|}{\cellcolor[HTML]{FFFFFF}3} &
  \multicolumn{1}{c|}{\cellcolor[HTML]{FFFFFF}3} &
  \multicolumn{1}{c|}{\cellcolor[HTML]{FFFFFF}0.8144} &
  \multicolumn{1}{c|}{\cellcolor[HTML]{FFFFFF}0.6321} &
  \multicolumn{1}{c|}{\cellcolor[HTML]{FFFFFF}0.2727} &
  \multicolumn{1}{l|}{\cellcolor[HTML]{FFFFFF}} &
  \multicolumn{1}{c|}{\cellcolor[HTML]{FFFFFF}124.8736} &
  \multicolumn{1}{c|}{\cellcolor[HTML]{FFFFFF}41.90\%} &
  \multicolumn{1}{c|}{\cellcolor[HTML]{FFFFFF}9.20\%} &
  \multicolumn{1}{c|}{\cellcolor[HTML]{FFFFFF}} &
  \multicolumn{1}{c|}{\cellcolor[HTML]{FFFFFF}111.0891} &
  \multicolumn{1}{c|}{\cellcolor[HTML]{FFFFFF}56.50\%} &
  \multicolumn{1}{c|}{\cellcolor[HTML]{FFFFFF}52.10\%} \\ \cline{3-7} \cline{9-11} \cline{13-15} 
\rowcolor[HTML]{FFFFFF} 
\multicolumn{1}{|l|}{\cellcolor[HTML]{FFFFFF}} &
  \multicolumn{1}{c|}{\multirow{-4}{*}{\cellcolor[HTML]{FFFFFF}\textbf{Low}}} &
  \multicolumn{1}{c|}{\cellcolor[HTML]{FFFFFF}3} &
  \multicolumn{1}{c|}{\cellcolor[HTML]{FFFFFF}3} &
  \multicolumn{1}{c|}{\cellcolor[HTML]{FFFFFF}0.8191} &
  \multicolumn{1}{c|}{\cellcolor[HTML]{FFFFFF}0.6321} &
  \multicolumn{1}{c|}{\cellcolor[HTML]{FFFFFF}0.3636} &
  \multicolumn{1}{l|}{\cellcolor[HTML]{FFFFFF}} &
  \multicolumn{1}{c|}{\cellcolor[HTML]{FFFFFF}196.0647} &
  \multicolumn{1}{c|}{\cellcolor[HTML]{FFFFFF}47\%} &
  \multicolumn{1}{c|}{\cellcolor[HTML]{FFFFFF}9.30\%} &
  \multicolumn{1}{c|}{\cellcolor[HTML]{FFFFFF}} &
  \multicolumn{1}{c|}{\cellcolor[HTML]{FFFFFF}202.7976} &
  \multicolumn{1}{c|}{\cellcolor[HTML]{FFFFFF}56.40\%} &
  \multicolumn{1}{c|}{\cellcolor[HTML]{FFFFFF}51.90\%} \\ \cline{2-7} \cline{9-11} \cline{13-15} 
\rowcolor[HTML]{FFFFFF} 
\multicolumn{1}{|l|}{\multirow{-5}{*}{\cellcolor[HTML]{FFFFFF}\textbf{granite3.1-MoE}}} &
  \multicolumn{1}{c|}{\cellcolor[HTML]{FFFFFF}\textbf{High}} &
  \multicolumn{1}{c|}{\cellcolor[HTML]{FFFFFF}4} &
  \multicolumn{1}{c|}{\cellcolor[HTML]{FFFFFF}4} &
  \multicolumn{1}{c|}{\cellcolor[HTML]{FFFFFF}0.8858} &
  \multicolumn{1}{c|}{\cellcolor[HTML]{FFFFFF}-} &
  \multicolumn{1}{c|}{\cellcolor[HTML]{FFFFFF}0.4545} &
  \multicolumn{1}{l|}{\cellcolor[HTML]{FFFFFF}} &
  \multicolumn{1}{c|}{\cellcolor[HTML]{FFFFFF}199.4555} &
  \multicolumn{1}{c|}{\cellcolor[HTML]{FFFFFF}46.50\%} &
  \multicolumn{1}{c|}{\cellcolor[HTML]{FFFFFF}9.30\%} &
  \multicolumn{1}{c|}{\cellcolor[HTML]{FFFFFF}} &
  \multicolumn{1}{c|}{\cellcolor[HTML]{FFFFFF}297.2756} &
  \multicolumn{1}{c|}{\cellcolor[HTML]{FFFFFF}56.70\%} &
  \multicolumn{1}{c|}{\cellcolor[HTML]{FFFFFF}52.00\%} \\ \cline{1-7} \cline{9-11} \cline{13-15} 
\rowcolor[HTML]{FFFFFF} 
\multicolumn{7}{l}{\cellcolor[HTML]{FFFFFF}\textbf{}} &
  \multicolumn{1}{l|}{\cellcolor[HTML]{FFFFFF}} &
  \multicolumn{3}{c}{\cellcolor[HTML]{FFFFFF}} &
  \multicolumn{1}{c|}{\cellcolor[HTML]{FFFFFF}} &
  \multicolumn{3}{c}{\cellcolor[HTML]{FFFFFF}} \\ \cline{1-7} \cline{9-11} \cline{13-15} 
\rowcolor[HTML]{FFFFFF} 
\multicolumn{1}{|l|}{\cellcolor[HTML]{FFFFFF}} &
  \multicolumn{1}{c|}{\cellcolor[HTML]{FFFFFF}} &
  \multicolumn{1}{c|}{\cellcolor[HTML]{FFFFFF}3} &
  \multicolumn{1}{c|}{\cellcolor[HTML]{FFFFFF}3} &
  \multicolumn{1}{c|}{\cellcolor[HTML]{FFFFFF}0.7741} &
  \multicolumn{1}{c|}{\cellcolor[HTML]{FFFFFF}0.6321} &
  \multicolumn{1}{c|}{\cellcolor[HTML]{FFFFFF}0.4737} &
  \multicolumn{1}{l|}{\cellcolor[HTML]{FFFFFF}} &
  \multicolumn{1}{c|}{\cellcolor[HTML]{FFFFFF}3.5381} &
  \multicolumn{1}{c|}{\cellcolor[HTML]{FFFFFF}2.70\%} &
  \multicolumn{1}{c|}{\cellcolor[HTML]{FFFFFF}3.60\%} &
  \multicolumn{1}{c|}{\cellcolor[HTML]{FFFFFF}} &
  \multicolumn{1}{c|}{\cellcolor[HTML]{FFFFFF}7.6907} &
  \multicolumn{1}{c|}{\cellcolor[HTML]{FFFFFF}3.10\%} &
  \multicolumn{1}{c|}{\cellcolor[HTML]{FFFFFF}52.90\%} \\ \cline{3-7} \cline{9-11} \cline{13-15} 
\rowcolor[HTML]{FFFFFF} 
\multicolumn{1}{|l|}{\cellcolor[HTML]{FFFFFF}} &
  \multicolumn{1}{c|}{\cellcolor[HTML]{FFFFFF}} &
  \multicolumn{1}{c|}{\cellcolor[HTML]{FFFFFF}3} &
  \multicolumn{1}{c|}{\cellcolor[HTML]{FFFFFF}3} &
  \multicolumn{1}{c|}{\cellcolor[HTML]{FFFFFF}0.7872} &
  \multicolumn{1}{c|}{\cellcolor[HTML]{FFFFFF}0.6321} &
  \multicolumn{1}{c|}{\cellcolor[HTML]{FFFFFF}0.4737} &
  \multicolumn{1}{l|}{\cellcolor[HTML]{FFFFFF}} &
  \multicolumn{1}{c|}{\cellcolor[HTML]{FFFFFF}2.1682} &
  \multicolumn{1}{c|}{\cellcolor[HTML]{FFFFFF}3.30\%} &
  \multicolumn{1}{c|}{\cellcolor[HTML]{FFFFFF}3.60\%} &
  \multicolumn{1}{c|}{\cellcolor[HTML]{FFFFFF}} &
  \multicolumn{1}{c|}{\cellcolor[HTML]{FFFFFF}6.5495} &
  \multicolumn{1}{c|}{\cellcolor[HTML]{FFFFFF}4.40\%} &
  \multicolumn{1}{c|}{\cellcolor[HTML]{FFFFFF}53.40\%} \\ \cline{3-7} \cline{9-11} \cline{13-15} 
\rowcolor[HTML]{EFEFEF} 
\multicolumn{1}{|l|}{\cellcolor[HTML]{FFFFFF}} &
  \multicolumn{1}{c|}{\cellcolor[HTML]{FFFFFF}} &
  \multicolumn{1}{c|}{\cellcolor[HTML]{EFEFEF}2} &
  \multicolumn{1}{c|}{\cellcolor[HTML]{EFEFEF}2} &
  \multicolumn{1}{c|}{\cellcolor[HTML]{EFEFEF}0.8288} &
  \multicolumn{1}{c|}{\cellcolor[HTML]{EFEFEF}0.4866} &
  \multicolumn{1}{c|}{\cellcolor[HTML]{EFEFEF}0.4211} &
  \multicolumn{1}{l|}{\cellcolor[HTML]{FFFFFF}} &
  \multicolumn{1}{c|}{\cellcolor[HTML]{EFEFEF}11.3214} &
  \multicolumn{1}{c|}{\cellcolor[HTML]{EFEFEF}2.40\%} &
  \multicolumn{1}{c|}{\cellcolor[HTML]{EFEFEF}3.60\%} &
  \multicolumn{1}{c|}{\cellcolor[HTML]{FFFFFF}} &
  \multicolumn{1}{c|}{\cellcolor[HTML]{EFEFEF}3.6525} &
  \multicolumn{1}{c|}{\cellcolor[HTML]{EFEFEF}4.70\%} &
  \multicolumn{1}{c|}{\cellcolor[HTML]{EFEFEF}52.90\%} \\ \cline{3-7} \cline{9-11} \cline{13-15} 
\rowcolor[HTML]{EFEFEF} 
\multicolumn{1}{|l|}{\cellcolor[HTML]{FFFFFF}} &
  \multicolumn{1}{c|}{\cellcolor[HTML]{FFFFFF}} &
  \multicolumn{1}{c|}{\cellcolor[HTML]{EFEFEF}\textbf{3}} &
  \multicolumn{1}{c|}{\cellcolor[HTML]{EFEFEF}\textbf{3}} &
  \multicolumn{1}{c|}{\cellcolor[HTML]{EFEFEF}\textbf{0.7937}} &
  \multicolumn{1}{c|}{\cellcolor[HTML]{EFEFEF}\textbf{0.6321}} &
  \multicolumn{1}{c|}{\cellcolor[HTML]{EFEFEF}\textbf{0.5263}} &
  \multicolumn{1}{l|}{\cellcolor[HTML]{FFFFFF}} &
  \multicolumn{1}{c|}{\cellcolor[HTML]{EFEFEF}\textbf{13.8917}} &
  \multicolumn{1}{c|}{\cellcolor[HTML]{EFEFEF}\textbf{2.00\%}} &
  \multicolumn{1}{c|}{\cellcolor[HTML]{EFEFEF}\textbf{3.60\%}} &
  \multicolumn{1}{c|}{\cellcolor[HTML]{FFFFFF}} &
  \multicolumn{1}{c|}{\cellcolor[HTML]{EFEFEF}\textbf{9.4319}} &
  \multicolumn{1}{c|}{\cellcolor[HTML]{EFEFEF}\textbf{2.30\%}} &
  \multicolumn{1}{c|}{\cellcolor[HTML]{EFEFEF}\textbf{53.20\%}} \\ \cline{3-7} \cline{9-11} \cline{13-15} 
\rowcolor[HTML]{FFFFFF} 
\multicolumn{1}{|l|}{\cellcolor[HTML]{FFFFFF}} &
  \multicolumn{1}{c|}{\multirow{-5}{*}{\cellcolor[HTML]{FFFFFF}\textbf{Low}}} &
  \multicolumn{1}{c|}{\cellcolor[HTML]{FFFFFF}3} &
  \multicolumn{1}{c|}{\cellcolor[HTML]{FFFFFF}3} &
  \multicolumn{1}{c|}{\cellcolor[HTML]{FFFFFF}0.7938} &
  \multicolumn{1}{c|}{\cellcolor[HTML]{FFFFFF}0.6321} &
  \multicolumn{1}{c|}{\cellcolor[HTML]{FFFFFF}0.4211} &
  \multicolumn{1}{l|}{\cellcolor[HTML]{FFFFFF}} &
  \multicolumn{1}{c|}{\cellcolor[HTML]{FFFFFF}11.6523} &
  \multicolumn{1}{c|}{\cellcolor[HTML]{FFFFFF}3.30\%} &
  \multicolumn{1}{c|}{\cellcolor[HTML]{FFFFFF}3.60\%} &
  \multicolumn{1}{c|}{\cellcolor[HTML]{FFFFFF}} &
  \multicolumn{1}{c|}{\cellcolor[HTML]{FFFFFF}3.4147} &
  \multicolumn{1}{c|}{\cellcolor[HTML]{FFFFFF}3.10\%} &
  \multicolumn{1}{c|}{\cellcolor[HTML]{FFFFFF}52.60\%} \\ \cline{2-7} \cline{9-11} \cline{13-15} 
\rowcolor[HTML]{FFFFFF} 
\multicolumn{1}{|l|}{\multirow{-6}{*}{\cellcolor[HTML]{FFFFFF}\textbf{Gpt-4o}}} &
  \multicolumn{1}{c|}{\cellcolor[HTML]{FFFFFF}\textbf{High}} &
  \multicolumn{1}{c|}{\cellcolor[HTML]{FFFFFF}2} &
  \multicolumn{1}{c|}{\cellcolor[HTML]{FFFFFF}2} &
  \multicolumn{1}{c|}{\cellcolor[HTML]{FFFFFF}0.8364} &
  \multicolumn{1}{c|}{\cellcolor[HTML]{FFFFFF}-} &
  \multicolumn{1}{c|}{\cellcolor[HTML]{FFFFFF}0.6315} &
  \multicolumn{1}{l|}{\multirow{-33}{*}{\cellcolor[HTML]{FFFFFF}}} &
  \multicolumn{1}{c|}{\cellcolor[HTML]{FFFFFF}1.9784} &
  \multicolumn{1}{c|}{\cellcolor[HTML]{FFFFFF}3.40\%} &
  \multicolumn{1}{c|}{\cellcolor[HTML]{FFFFFF}3.60\%} &
  \multicolumn{1}{c|}{\multirow{-33}{*}{\cellcolor[HTML]{FFFFFF}\textbf{}}} &
  \multicolumn{1}{c|}{\cellcolor[HTML]{FFFFFF}1.3812} &
  \multicolumn{1}{c|}{\cellcolor[HTML]{FFFFFF}5.60\%} &
  \multicolumn{1}{c|}{\cellcolor[HTML]{FFFFFF}49.00\%} \\ \cline{1-7} \cline{9-11} \cline{13-15} 
\end{tabular}
}
\end{table*}

\begin{figure*}
	\centering
\subfloat[]{\resizebox{48mm}{42mm}{\begin{tikzpicture}
    \begin{axis}[
    ybar,
    width=0.56\textwidth,
    height=.42\textwidth,
            bar width=.18cm,
            symbolic x coords={0.5,1,1.5,2,2.5,3,3.5}, xtick={0.5,1,1.5,2,2.5,3,3.5},  
        xticklabels={\ , 3,\ , 4,\ , 3, \ }, 	 %
            legend columns=3,ymin=0.0,ymax=0.9,
            legend pos=south east,legend style={font=\fontsize{5}{5}\selectfont},
	ylabel= \%, ylabel style={font=\fontsize{10}{10}\selectfont},ymajorgrids=true,   grid style=dashed, xlabel={LM Call Counts},
	 xticklabel style={rotate=0, anchor=east}, 
]
\addplot[fill=black] table[x=interval,y=Similarity]{\agentperparmodela};
\addplot[fill=darkgray, pattern=horizontal lines light gray] table[x=interval,y=LMUsage]{\agentperparmodela};
\addplot[fill=lightgray] table[x=interval,y=Precision]{\agentperparmodela};
\draw[fill=gray, opacity=0.3] (axis cs:0.5,0) rectangle (axis cs:2.5,9); 
\draw[dotted,ultra thick, red](axis cs:1.5,0) rectangle (axis cs:2.5,9);

\legend{Similarity, LM Usage, Precision}
\end{axis}  
\end{tikzpicture} } \label{Fig51a}}
\subfloat[]{\resizebox{48mm}{42mm}{\begin{tikzpicture}
    \begin{axis}[
    ybar,
    width=0.56\textwidth,
    height=.42\textwidth,
            bar width=.18cm,
            symbolic x coords={0.5,1,1.5,2,2.5}, xtick={0.5,1,1.5,2,2.5},  
        xticklabels={\ , 2,\ , 2,\ }, 	 %
            legend columns=3,ymin=0.0,
            legend pos=south east,legend style={font=\fontsize{5}{5}\selectfont},
	ylabel= \%, ylabel style={font=\fontsize{10}{10}\selectfont},ymajorgrids=true,   grid style=dashed, xlabel={LM Call Counts},
	 xticklabel style={rotate=0, anchor=east}, 
]
\addplot[fill=black] table[x=interval,y=Similarity]{\agentperparmodelb};
\addplot[fill=darkgray, pattern=horizontal lines light gray] table[x=interval,y=LMUsage]{\agentperparmodelb};
\addplot[fill=lightgray] table[x=interval,y=Precision]{\agentperparmodelb};
\draw[fill=gray, opacity=0.3] (axis cs:0.5,0) rectangle (axis cs:1.5,1); 
\draw[dotted,ultra thick, red](axis cs:0.5,0) rectangle (axis cs:1.5,1);

\legend{Similarity, LM Usage, Precision}
\end{axis}  
\end{tikzpicture} }\label{Fig51b}}
\subfloat[]{\resizebox{48mm}{42mm}{\begin{tikzpicture}
    \begin{axis}[
    ybar,
    width=0.56\textwidth,
    height=.42\textwidth,
            bar width=.18cm,
            symbolic x coords={0.5,1,1.5,2,2.5,3,3.5}, xtick={0.5,1,1.5,2,2.5,3,3.5},  
        xticklabels={\ , 5,\ , 4,\ , 4, \ }, 	 	 %
            legend columns=3,ymin=0.0,
            legend pos=south east,legend style={font=\fontsize{5}{5}\selectfont},
	ylabel= \%, ylabel style={font=\fontsize{10}{10}\selectfont},ymajorgrids=true,   grid style=dashed, xlabel={LM Call Counts},
	 xticklabel style={rotate=0, anchor=east}, 
]
\addplot[fill=black] table[x=interval,y=Similarity]{\agentperparmodelc};
\addplot[fill=darkgray, pattern=horizontal lines light gray] table[x=interval,y=LMUsage]{\agentperparmodelc};
\addplot[fill=lightgray] table[x=interval,y=Precision]{\agentperparmodelc};
\draw[fill=gray, opacity=0.3] (axis cs:1.5,0) rectangle (axis cs:2.5,1); 
\draw[dotted,ultra thick, red](axis cs:1.5,0) rectangle (axis cs:2.5,1);

\legend{Similarity, LM Usage, Precision}
\end{axis}  
\end{tikzpicture} }\label{Fig51c}}\\
\subfloat[]{\resizebox{48mm}{42mm}{\begin{tikzpicture}
    \begin{axis}[
    ybar,
    width=0.56\textwidth,
    height=.42\textwidth,
            bar width=.18cm,
            symbolic x coords={0.5,1,1.5,2,2.5,3,3.5}, xtick={0.5,1,1.5,2,2.5,3,3.5},  
        xticklabels={\ , 3,\ , 2,\ , 2, \ },  	 %
            legend columns=3,ymin=0.0,ymax=0.85,
            legend pos=south east,legend style={font=\fontsize{5}{5}\selectfont},
	ylabel= \%, ylabel style={font=\fontsize{10}{10}\selectfont},ymajorgrids=true,   grid style=dashed, xlabel={LM Call Counts},
	 xticklabel style={rotate=0, anchor=east}, 
]
\addplot[fill=black] table[x=interval,y=Similarity]{\agentperparmodeld};
\addplot[fill=darkgray, pattern=horizontal lines light gray] table[x=interval,y=LMUsage]{\agentperparmodeld};
\addplot[fill=lightgray] table[x=interval,y=Precision]{\agentperparmodeld};
\draw[fill=gray, opacity=0.3] (axis cs:2.5,0) rectangle (axis cs:3.5,1); 
\draw[dotted,ultra thick, red](axis cs:2.5,0) rectangle (axis cs:3.5,1);

\legend{Similarity, LM Usage, Precision}
\end{axis}  
\end{tikzpicture} }\label{Fig51d}}
\subfloat[]{\resizebox{48mm}{42mm}{\begin{tikzpicture}
    \begin{axis}[
    ybar,
    width=0.56\textwidth,
    height=.42\textwidth,
            bar width=.18cm,
            symbolic x coords={0.5,1,1.5,2,2.5,3,3.5,4}, xtick={0.5,1,1.5,2,2.5,3,3.5,4},  
        xticklabels={\ , 3,\ , 3,\ , 3,\ , 3 }, 	 %
            legend columns=3,ymin=0.0,ymax=0.9,
            legend pos=south east,legend style={font=\fontsize{5}{5}\selectfont},
	ylabel= \%, ylabel style={font=\fontsize{10}{10}\selectfont},ymajorgrids=true,   grid style=dashed, xlabel={LM Call Counts},
	 xticklabel style={rotate=0, anchor=east}, 
]
\addplot[fill=black] table[x=interval,y=Similarity]{\agentperparmodele};
\addplot[fill=darkgray, pattern=horizontal lines light gray] table[x=interval,y=LMUsage]{\agentperparmodele};
\addplot[fill=lightgray] table[x=interval,y=Precision]{\agentperparmodele};
\draw[fill=gray, opacity=0.3] (axis cs:0.5,0) rectangle (axis cs:1.5,0.9); 
\draw[dotted,ultra thick, red](axis cs:0.5,0) rectangle (axis cs:1.5,0.9);

\legend{Similarity, LM Usage, Precision}
\end{axis}  
\end{tikzpicture} }\label{Fig51e}}
\subfloat[]{\resizebox{48mm}{42mm}{\begin{tikzpicture}
    \begin{axis}[
    ybar,
    width=0.56\textwidth,
    height=.42\textwidth,
            bar width=.18cm,
            symbolic x coords={1,1.5,2,2.5,3,3.5,4,4.5,5}, xtick={1,1.5,2,2.5,3,3.5,4,4.5,5},  
        xticklabels={3,\ , 3,\ , 2,\ , 3,\ , 3}, 	 %
            legend columns=3,ymin=0.0,ymax=0.85,
            legend pos=south east,legend style={font=\fontsize{5}{5}\selectfont},
	ylabel= \%, ylabel style={font=\fontsize{10}{10}\selectfont},ymajorgrids=true,   grid style=dashed, xlabel={LM Call Counts},
	 xticklabel style={rotate=0, anchor=east}, 
]
\addplot[fill=black] table[x=interval,y=Similarity]{\agentperparmodelf};
\addplot[fill=darkgray, pattern=horizontal lines light gray] table[x=interval,y=LMUsage]{\agentperparmodelf};
\addplot[fill=lightgray] table[x=interval,y=Precision]{\agentperparmodelf};
\draw[fill=gray, opacity=0.3] (axis cs:1.5,0) rectangle (axis cs:3.5,0.85); 
\draw[dotted,ultra thick, red](axis cs:2.5,0) rectangle (axis cs:3.5,0.85);

\legend{Similarity, LM Usage, Precision}
\end{axis}  
\end{tikzpicture} }\label{Fig51f}}
\caption{Evaluation of Low-Urgency Agent Performance Metrics and Pareto Analysis. \protect\subref{Fig51a}  Gemini-1.5 flash \protect\subref{Fig51b} Command-r7b \protect\subref{Fig51c} Claude 3.5 Sonnet \protect\subref{Fig51d} Mistral \protect\subref{Fig51e} Granite3.1-MoE \protect\subref{Fig51f} Gpt-4o}
\end{figure*}
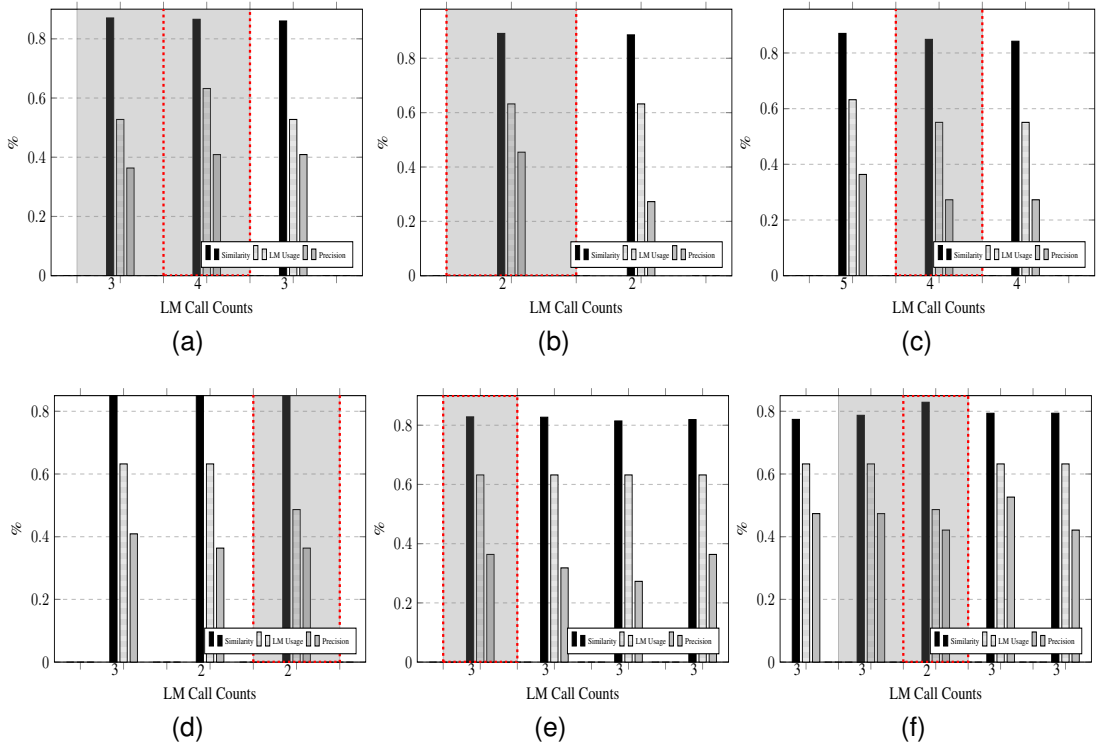

\begin{figure*} 
	\centering
\subfloat[]{\resizebox{40mm}{34mm}{\begin{tikzpicture}
\pgfplotsset{
    scale only axis,
}
  \begin{axis}[
    ybar,
    width=0.12\textwidth,
    height=.22\textwidth,
    bar width=.18cm,
    symbolic x coords={Low1,Low2,High},
    xtick=data, legend style={font=\fontsize{4}{3}\selectfont},
    ylabel=CPU Utilization (\%),
    axis y line*=left,
    axis x line*=bottom,
    ymajorgrids=true,
    grid style=dashed,
    xticklabel style={rotate=90, anchor=east},
    legend style={at={(0.5,1.05)}, anchor=south, legend columns=-1},
    ]
    \addplot[fill=black, pattern=horizontal lines light gray, bar width=0.42cm, opacity=0.7] table[x=interval, y=CPUC]{\agentperparmodelaa};\label{plot_twoa}
  \end{axis}
  \begin{axis}[
    ybar,
    width=0.12\textwidth,
    height=.22\textwidth,
    bar width=.18cm,
     symbolic x coords={Low1,Low2,High},
    xtick=\empty, legend columns=2, legend pos=north east, legend style={font=\fontsize{4}{3}\selectfont},
    ylabel=Elapsed Time (s),
    axis y line*=right,
    axis x line=none,
    ymajorgrids=false,
    legend style={at={(0.5,1.05)}, anchor=south, legend columns=-1},
    ]
    \addplot[fill=gray] table[x=interval, y=ElapsedC]{\agentperparmodelaa};
    \addlegendimage{/pgfplots/refstyle=plot_twoa}\addlegendentry{plot 1} \legend{CPU, Elapsed}
  \end{axis}
\end{tikzpicture}\label{Fig52a}}}
\subfloat[]{\resizebox{40mm}{34mm}{\begin{tikzpicture}
\pgfplotsset{
    scale only axis,
}
  \begin{axis}[
    ybar,
    width=0.12\textwidth,
    height=.22\textwidth,
    bar width=.18cm,
    symbolic x coords={Low,High},
    xtick=data, legend style={font=\fontsize{4}{3}\selectfont},
    ylabel=CPU Utilization (\%),
    axis y line*=left,
    axis x line*=bottom,
    ymajorgrids=true,
    grid style=dashed,
    xticklabel style={rotate=90, anchor=east},
    legend style={at={(0.5,1.05)}, anchor=south, legend columns=-1},
    ]
    \addplot[fill=black, pattern=horizontal lines light gray, bar width=0.42cm, opacity=0.7] table[x=interval, y=CPUC]{\agentperparmodelbb};\label{plot_twob}
  \end{axis}
  \begin{axis}[
    ybar,
    width=0.12\textwidth,
    height=.22\textwidth,
    bar width=.18cm,
     symbolic x coords={Low,High},
    xtick=\empty, legend columns=2, legend pos=north east, legend style={font=\fontsize{4}{3}\selectfont},
    ylabel=Elapsed Time (s),
    axis y line*=right,
    axis x line=none,
    ymajorgrids=false,
    legend style={at={(0.5,1.05)}, anchor=south, legend columns=-1},
    ]
    \addplot[fill=gray] table[x=interval, y=ElapsedC]{\agentperparmodelbb};
    \addlegendimage{/pgfplots/refstyle=plot_twob}\addlegendentry{plot 1} \legend{CPU, Elapsed}
  \end{axis}
\end{tikzpicture}\label{Fig52b}}}
\subfloat[]{\resizebox{40mm}{34mm}{\begin{tikzpicture}
\pgfplotsset{
    scale only axis,
}
  \begin{axis}[
    ybar,
    width=0.12\textwidth,
    height=.22\textwidth,
    bar width=.18cm,
    symbolic x coords={Low,High},
    xtick=data, legend style={font=\fontsize{4}{3}\selectfont},
    ylabel=CPU Utilization (\%),
    axis y line*=left,
    axis x line*=bottom,
    ymajorgrids=true,
    grid style=dashed,
    xticklabel style={rotate=90, anchor=east},
    legend style={at={(0.5,1.05)}, anchor=south, legend columns=-1},
    ]
    \addplot[fill=black, pattern=horizontal lines light gray, bar width=0.42cm, opacity=0.7] table[x=interval, y=CPUC]{\agentperparmodelcc};\label{plot_twoc}
  \end{axis}
  \begin{axis}[
    ybar,
    width=0.12\textwidth,
    height=.22\textwidth,
    bar width=.18cm,
     symbolic x coords={Low,High},
    xtick=\empty, legend columns=2, legend pos=north east, legend style={font=\fontsize{4}{3}\selectfont},
    ylabel=Elapsed Time (s),
    axis y line*=right,
    axis x line=none,
    ymajorgrids=false,
    legend style={at={(0.5,1.05)}, anchor=south, legend columns=-1},
    ]
    \addplot[fill=gray] table[x=interval, y=ElapsedC]{\agentperparmodelcc};
    \addlegendimage{/pgfplots/refstyle=plot_twoc}\addlegendentry{plot 3} \legend{CPU, Elapsed}
  \end{axis}
\end{tikzpicture}\label{Fig52c}}}
\subfloat[]{\resizebox{40mm}{34mm}{\begin{tikzpicture}
\pgfplotsset{
    scale only axis,
}
  \begin{axis}[
    ybar,
    width=0.12\textwidth,
    height=.22\textwidth,
    bar width=.18cm,
    symbolic x coords={Low,High},
    xtick=data, legend style={font=\fontsize{4}{3}\selectfont},
    ylabel=CPU Utilization (\%),
    axis y line*=left,
    axis x line*=bottom,
    ymajorgrids=true,
    grid style=dashed,
    xticklabel style={rotate=90, anchor=east},
    legend style={at={(0.5,1.05)}, anchor=south, legend columns=-1},
    ]
    \addplot[fill=black, pattern=horizontal lines light gray, bar width=0.42cm, opacity=0.7] table[x=interval, y=CPUC]{\agentperparmodeldd};\label{plot_twod}
  \end{axis}
  \begin{axis}[
    ybar,
    width=0.12\textwidth,
    height=.22\textwidth,
    bar width=.18cm,
     symbolic x coords={Low,High},
    xtick=\empty, legend columns=2, legend pos=north east, legend style={font=\fontsize{4}{3}\selectfont},
    ylabel=Elapsed Time (s),
    axis y line*=right,
    axis x line=none,
    ymajorgrids=false,
    legend style={at={(0.5,1.05)}, anchor=south, legend columns=-1},
    ]
    \addplot[fill=gray] table[x=interval, y=ElapsedC]{\agentperparmodeldd};
    \addlegendimage{/pgfplots/refstyle=plot_twod}\addlegendentry{plot 4} \legend{CPU, Elapsed}
  \end{axis}
\end{tikzpicture}\label{Fig52d}}}

\subfloat[]{\resizebox{40mm}{34mm}{\begin{tikzpicture}
\pgfplotsset{
    scale only axis,
}
  \begin{axis}[
    ybar,
    width=0.12\textwidth,
    height=.22\textwidth,
    bar width=.18cm,
    symbolic x coords={Low,High},
    xtick=data, legend style={font=\fontsize{4}{3}\selectfont},
    ylabel=CPU Utilization (\%),
    axis y line*=left,
    axis x line*=bottom,
    ymajorgrids=true,
    grid style=dashed,
    xticklabel style={rotate=90, anchor=east},
    legend style={at={(0.5,1.05)}, anchor=south, legend columns=-1},
    ]
    \addplot[fill=black, pattern=horizontal lines light gray, bar width=0.42cm, opacity=0.7] table[x=interval, y=CPUC]{\agentperparmodelee};\label{plot_twoe}
  \end{axis}
  \begin{axis}[
    ybar,
    width=0.12\textwidth,
    height=.22\textwidth,
    bar width=.18cm,
     symbolic x coords={Low,High},
    xtick=\empty, legend columns=2, legend pos=north east, legend style={font=\fontsize{4}{3}\selectfont},
    ylabel=Elapsed Time (s),
    axis y line*=right,
    axis x line=none,
    ymajorgrids=false,
    legend style={at={(0.5,1.05)}, anchor=south, legend columns=-1},
    ]
    \addplot[fill=gray] table[x=interval, y=ElapsedC]{\agentperparmodelee};
    \addlegendimage{/pgfplots/refstyle=plot_twoe}\addlegendentry{plot 5} \legend{CPU, Elapsed}
  \end{axis}
\end{tikzpicture}\label{Fig52e}}}
\subfloat[]{\resizebox{40mm}{34mm}{\begin{tikzpicture}
\pgfplotsset{
    scale only axis,
}
  \begin{axis}[
    ybar,
    width=0.12\textwidth,
    height=.22\textwidth,
    bar width=.18cm,
    symbolic x coords={Low1,Low2,High},
    xtick=data, legend style={font=\fontsize{4}{3}\selectfont},
    ylabel=CPU Utilization (\%),
    axis y line*=left,
    axis x line*=bottom,
    ymajorgrids=true,
    grid style=dashed,
    xticklabel style={rotate=90, anchor=east},
    legend style={at={(0.5,1.05)}, anchor=south, legend columns=-1},
    ]
    \addplot[fill=black, pattern=horizontal lines light gray, bar width=0.42cm, opacity=0.7] table[x=interval, y=CPUC]{\agentperparmodelff};\label{plot_twof}
  \end{axis}
  \begin{axis}[
    ybar,
    width=0.12\textwidth,
    height=.22\textwidth,
    bar width=.18cm,
     symbolic x coords={Low1,Low2,High},
    xtick=\empty, legend columns=2, legend pos=north east, legend style={font=\fontsize{4}{3}\selectfont},
    ylabel=Elapsed Time (s),
    axis y line*=right,
    axis x line=none,
    ymajorgrids=false,
    legend style={at={(0.5,1.05)}, anchor=south, legend columns=-1},
    ]
    \addplot[fill=gray] table[x=interval, y=ElapsedC]{\agentperparmodelff};
    \addlegendimage{/pgfplots/refstyle=plot_twof}\addlegendentry{plot 6} \legend{CPU, Elapsed}
  \end{axis}
\end{tikzpicture}\label{Fig52f}}}
\subfloat[]{\resizebox{40mm}{34mm}{\begin{tikzpicture}
\pgfplotsset{
    scale only axis,
}
  \begin{axis}[
    ybar,
    width=0.12\textwidth,
    height=.22\textwidth,
    bar width=.18cm,
    symbolic x coords={Low1,Low2,High},
    xtick=data, legend style={font=\fontsize{4}{3}\selectfont},
    ylabel=CPU Utilization (\%),
    axis y line*=left,
    axis x line*=bottom,
    ymajorgrids=true,
    grid style=dashed,
    xticklabel style={rotate=90, anchor=east},
    legend style={at={(0.5,1.05)}, anchor=south, legend columns=-1},
    ]
    \addplot[fill=black, pattern=horizontal lines light gray, bar width=0.42cm, opacity=0.7] table[x=interval, y=CPUE]{\agentperparmodelaa};\label{plot_twog}
  \end{axis}
  \begin{axis}[
    ybar,
    width=0.12\textwidth,
    height=.22\textwidth,
    bar width=.18cm,
     symbolic x coords={Low1,Low2,High},
    xtick=\empty, legend columns=2, legend pos=north east, legend style={font=\fontsize{4}{3}\selectfont},
    ylabel=Elapsed Time (s),
    axis y line*=right,
    axis x line=none,
    ymajorgrids=false,
    legend style={at={(0.5,1.05)}, anchor=south, legend columns=-1},
    ]
    \addplot[fill=gray] table[x=interval, y=ElapsedE]{\agentperparmodelaa};
    \addlegendimage{/pgfplots/refstyle=plot_twog}\addlegendentry{plot 7} \legend{CPU, Elapsed}
  \end{axis}
\end{tikzpicture}\label{Fig52g}}}
\subfloat[]{\resizebox{40mm}{34mm}{\begin{tikzpicture}
\pgfplotsset{
    scale only axis,
}
  \begin{axis}[
    ybar,
    width=0.12\textwidth,
    height=.22\textwidth,
    bar width=.18cm,
    symbolic x coords={Low,High},
    xtick=data, legend style={font=\fontsize{4}{3}\selectfont},
    ylabel=CPU Utilization (\%),
    axis y line*=left,
    axis x line*=bottom,
    ymajorgrids=true,
    grid style=dashed,
    xticklabel style={rotate=90, anchor=east},
    legend style={at={(0.5,1.05)}, anchor=south, legend columns=-1},
    ]
    \addplot[fill=black, pattern=horizontal lines light gray, bar width=0.42cm, opacity=0.7] table[x=interval, y=CPUE]{\agentperparmodelbb};\label{plot_twoh}
  \end{axis}
  \begin{axis}[
    ybar,
    width=0.12\textwidth,
    height=.22\textwidth,
    bar width=.18cm,
     symbolic x coords={Low,High},
    xtick=\empty, legend columns=2, legend pos=north east, legend style={font=\fontsize{4}{3}\selectfont},
    ylabel=Elapsed Time (s),
    axis y line*=right,
    axis x line=none,
    ymajorgrids=false,
    legend style={at={(0.5,1.05)}, anchor=south, legend columns=-1},
    ]
    \addplot[fill=gray] table[x=interval, y=ElapsedE]{\agentperparmodelbb};
    \addlegendimage{/pgfplots/refstyle=plot_twoh}\addlegendentry{plot 8} \legend{CPU, Elapsed}
  \end{axis}
\end{tikzpicture}\label{Fig52h}}}

\subfloat[]{\resizebox{40mm}{34mm}{\begin{tikzpicture}
\pgfplotsset{
    scale only axis,
}
  \begin{axis}[
    ybar,
    width=0.12\textwidth,
    height=.22\textwidth,
    bar width=.18cm,
    symbolic x coords={Low,High},
    xtick=data, legend style={font=\fontsize{4}{3}\selectfont},
    ylabel=CPU Utilization (\%),
    axis y line*=left,
    axis x line*=bottom,
    ymajorgrids=true,
    grid style=dashed,
    xticklabel style={rotate=90, anchor=east},
    legend style={at={(0.5,1.05)}, anchor=south, legend columns=-1},
    ]
    \addplot[fill=black, pattern=horizontal lines light gray, bar width=0.42cm, opacity=0.7] table[x=interval, y=CPUE]{\agentperparmodelcc};\label{plot_twoi}
  \end{axis}
  \begin{axis}[
    ybar,
    width=0.12\textwidth,
    height=.22\textwidth,
    bar width=.18cm,
     symbolic x coords={Low,High},
    xtick=\empty, legend columns=2, legend pos=north east, legend style={font=\fontsize{4}{3}\selectfont},
    ylabel=Elapsed Time (s),
    axis y line*=right,
    axis x line=none,
    ymajorgrids=false,
    legend style={at={(0.5,1.05)}, anchor=south, legend columns=-1},
    ]
    \addplot[fill=gray] table[x=interval, y=ElapsedE]{\agentperparmodelcc};
    \addlegendimage{/pgfplots/refstyle=plot_twoi}\addlegendentry{plot 9} \legend{CPU, Elapsed}
  \end{axis}
\end{tikzpicture}\label{Fig52i}}}
\subfloat[]{\resizebox{40mm}{34mm}{\begin{tikzpicture}
\pgfplotsset{
    scale only axis,
}
  \begin{axis}[
    ybar,
    width=0.12\textwidth,
    height=.22\textwidth,
    bar width=.18cm,
    symbolic x coords={Low,High},
    xtick=data, legend style={font=\fontsize{4}{3}\selectfont},
    ylabel=CPU Utilization (\%),
    axis y line*=left,
    axis x line*=bottom,
    ymajorgrids=true,
    grid style=dashed,
    xticklabel style={rotate=90, anchor=east},
    legend style={at={(0.5,1.05)}, anchor=south, legend columns=-1},
    ]
    \addplot[fill=black, pattern=horizontal lines light gray, bar width=0.42cm, opacity=0.7] table[x=interval, y=CPUE]{\agentperparmodeldd};\label{plot_twoj}
  \end{axis}
  \begin{axis}[
    ybar,
    width=0.12\textwidth,
    height=.22\textwidth,
    bar width=.18cm,
     symbolic x coords={Low,High},
    xtick=\empty, legend columns=2, legend pos=north east, legend style={font=\fontsize{4}{3}\selectfont},
    ylabel=Elapsed Time (s),
    axis y line*=right,
    axis x line=none,
    ymajorgrids=false,
    legend style={at={(0.5,1.05)}, anchor=south, legend columns=-1},
    ]
    \addplot[fill=gray] table[x=interval, y=ElapsedE]{\agentperparmodeldd};
    \addlegendimage{/pgfplots/refstyle=plot_twoj}\addlegendentry{plot 10} \legend{CPU, Elapsed}
  \end{axis}
\end{tikzpicture}\label{Fig52j}}}
\subfloat[]{\resizebox{40mm}{34mm}{\begin{tikzpicture}
\pgfplotsset{
    scale only axis,
}
  \begin{axis}[
    ybar,
    width=0.12\textwidth,
    height=.22\textwidth,
    bar width=.18cm,
    symbolic x coords={Low,High},
    xtick=data, legend style={font=\fontsize{4}{3}\selectfont},
    ylabel=CPU Utilization (\%),
    axis y line*=left,
    axis x line*=bottom,
    ymajorgrids=true,
    grid style=dashed,
    xticklabel style={rotate=90, anchor=east},
    legend style={at={(0.5,1.05)}, anchor=south, legend columns=-1},
    ]
    \addplot[fill=black, pattern=horizontal lines light gray, bar width=0.42cm, opacity=0.7] table[x=interval, y=CPUE]{\agentperparmodelee};\label{plot_twok}
  \end{axis}
  \begin{axis}[
    ybar,
    width=0.12\textwidth,
    height=.22\textwidth,
    bar width=.18cm,
     symbolic x coords={Low,High},
    xtick=\empty, legend columns=2, legend pos=north east, legend style={font=\fontsize{4}{3}\selectfont},
    ylabel=Elapsed Time (s),
    axis y line*=right,
    axis x line=none,
    ymajorgrids=false,
    legend style={at={(0.5,1.05)}, anchor=south, legend columns=-1},
    ]
    \addplot[fill=gray] table[x=interval, y=ElapsedE]{\agentperparmodelee};
    \addlegendimage{/pgfplots/refstyle=plot_twok}\addlegendentry{plot 11} \legend{CPU, Elapsed}
  \end{axis}
\end{tikzpicture}\label{Fig52k}}}
\subfloat[]{\resizebox{40mm}{34mm}{\begin{tikzpicture}
\pgfplotsset{
    scale only axis,
}
  \begin{axis}[
    ybar,
    width=0.12\textwidth,
    height=.22\textwidth,
    bar width=.18cm,
    symbolic x coords={Low1,Low2,High},
    xtick=data, legend style={font=\fontsize{4}{3}\selectfont},
    ylabel=CPU Utilization (\%),
    axis y line*=left,
    axis x line*=bottom,
    ymajorgrids=true,
    grid style=dashed,
    xticklabel style={rotate=90, anchor=east},
    legend style={at={(0.5,1.05)}, anchor=south, legend columns=-1},
    ]
    \addplot[fill=black, pattern=horizontal lines light gray, bar width=0.42cm, opacity=0.7] table[x=interval, y=CPUE]{\agentperparmodelff};\label{plot_twol}
  \end{axis}
  \begin{axis}[
    ybar,
    width=0.12\textwidth,
    height=.22\textwidth,
    bar width=.18cm,
     symbolic x coords={Low1,Low2,High},
    xtick=\empty, legend columns=2, legend pos=north east, legend style={font=\fontsize{4}{3}\selectfont},
    ylabel=Elapsed Time (s),
    axis y line*=right,
    axis x line=none,
    ymajorgrids=false,
    legend style={at={(0.5,1.05)}, anchor=south, legend columns=-1},
    ]
    \addplot[fill=gray] table[x=interval, y=ElapsedE]{\agentperparmodelff};
    \addlegendimage{/pgfplots/refstyle=plot_twol}\addlegendentry{plot 12} \legend{CPU, Elapsed}
  \end{axis}
\end{tikzpicture}\label{Fig52l}}}

\caption{Evaluation of Low-Level Agents Performance at Cloud: \protect\subref{Fig52a}  Gemini-1.5 flash \protect\subref{Fig52b} Command-r7b \protect\subref{Fig52c} Claude 3.5 Sonnet \protect\subref{Fig52d} Mistral \protect\subref{Fig52e} Granite3.1-MoE \protect\subref{Fig52f} Gpt-4o; Evaluation of Low-Level Agents Performance at the Edge: \protect\subref{Fig52g}  Gemini-1.5 flash \protect\subref{Fig52h} Command-r7b \protect\subref{Fig52i} Claude 3.5 Sonnet \protect\subref{Fig52j} Mistral \protect\subref{Fig52k} Granite3.1-MoE \protect\subref{Fig52l} Gpt-4o;} 
\end{figure*}
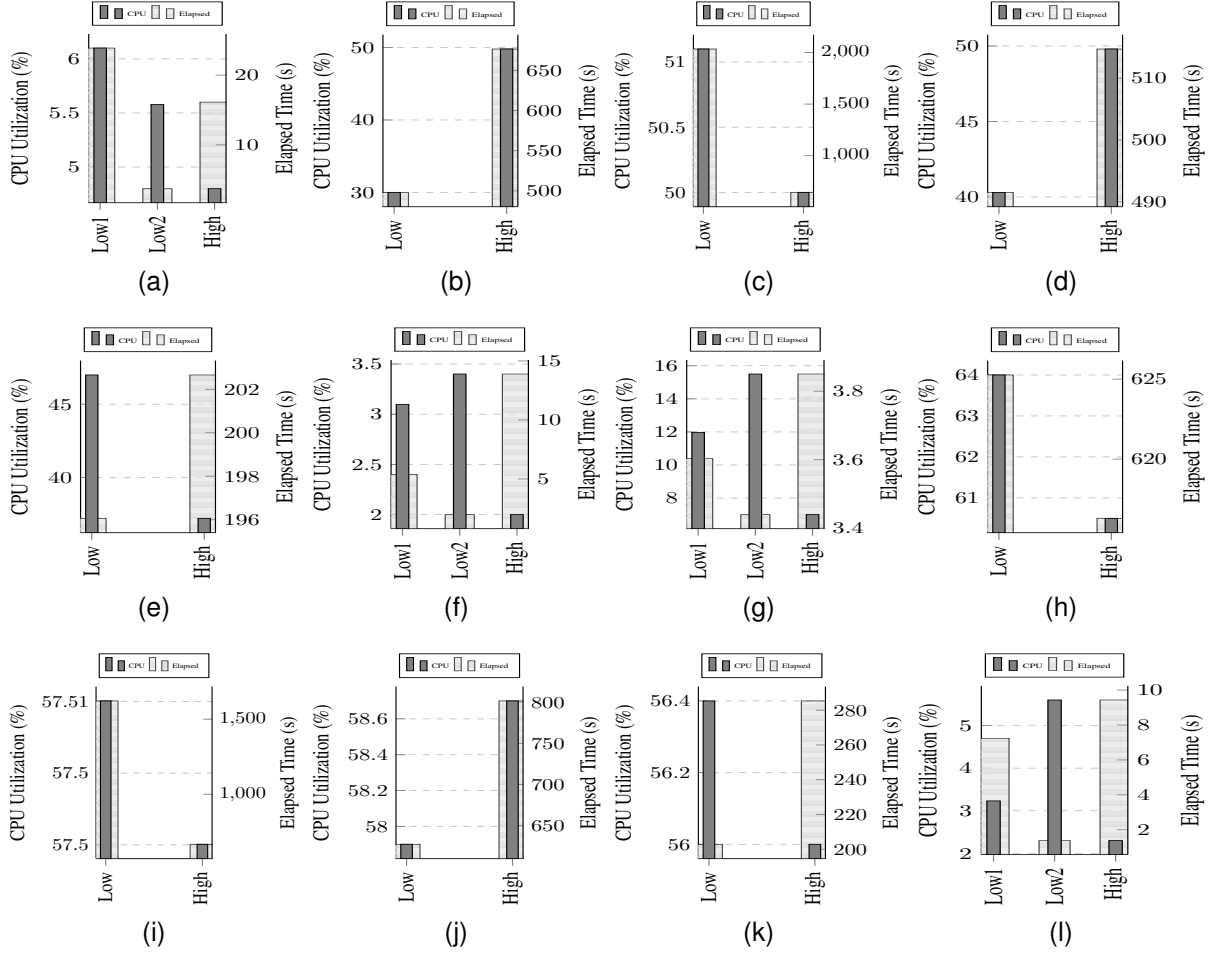

%% file: tables/table-learning.tex
\begin{table*}[]
\centering
\caption{Low-urgency Learning Performance Evaluation.}
\label{tab:learn}
\resizebox{!}{.15\paperheight}{%

\begin{tabular}{ccccccccccc}
\hline
\rowcolor[HTML]{FFFFFF} 
\multicolumn{11}{c}{\cellcolor[HTML]{FFFFFF}\textbf{Before Learning}} \\ \hline
\rowcolor[HTML]{FFFFFF} 
\multicolumn{4}{c|}{\cellcolor[HTML]{FFFFFF}\textbf{}} &
  \multicolumn{3}{c|}{\cellcolor[HTML]{FFFFFF}\textbf{Metrics}} &
  \multicolumn{1}{c|}{\cellcolor[HTML]{FFFFFF}} &
  \multicolumn{3}{c|}{\cellcolor[HTML]{FFFFFF}\textbf{Cloud Server}} \\ \cline{1-7} \cline{9-11} 
\rowcolor[HTML]{FFFFFF} 
\multicolumn{1}{|c|}{\cellcolor[HTML]{FFFFFF}\textbf{\begin{tabular}[c]{@{}c@{}}Model \\ Name\end{tabular}}} &
  \multicolumn{1}{c|}{\cellcolor[HTML]{FFFFFF}\textbf{\begin{tabular}[c]{@{}c@{}}Urgency\\ Level\end{tabular}}} &
  \multicolumn{1}{c|}{\cellcolor[HTML]{FFFFFF}\textbf{\begin{tabular}[c]{@{}c@{}}LM\\ Call\\ Count\end{tabular}}} &
  \multicolumn{1}{c|}{\cellcolor[HTML]{FFFFFF}\textbf{\begin{tabular}[c]{@{}c@{}}Hierarchy\\ Depth\\ Count\end{tabular}}} &
  \multicolumn{1}{c|}{\cellcolor[HTML]{FFFFFF}\textbf{\begin{tabular}[c]{@{}c@{}}Similarity \\ Score\end{tabular}}} &
  \multicolumn{1}{c|}{\cellcolor[HTML]{FFFFFF}\textbf{\begin{tabular}[c]{@{}c@{}}LM\\ Call\\ Usage\\ Cost\end{tabular}}} &
  \multicolumn{1}{c|}{\cellcolor[HTML]{FFFFFF}\textbf{\begin{tabular}[c]{@{}c@{}}Precision\\ Score\end{tabular}}} &
  \multicolumn{1}{c|}{\cellcolor[HTML]{FFFFFF}} &
  \multicolumn{1}{c|}{\cellcolor[HTML]{FFFFFF}\textbf{\begin{tabular}[c]{@{}c@{}}Elapsed\\ Time\end{tabular}}} &
  \multicolumn{1}{c|}{\cellcolor[HTML]{FFFFFF}\textbf{\begin{tabular}[c]{@{}c@{}}CPU\\ Utilization\end{tabular}}} &
  \multicolumn{1}{c|}{\cellcolor[HTML]{FFFFFF}\textbf{\begin{tabular}[c]{@{}c@{}}Memory\\ Utilization\end{tabular}}} \\ \cline{1-7} \cline{9-11} 
\rowcolor[HTML]{FFFFFF} 
\multicolumn{1}{|c|}{\cellcolor[HTML]{FFFFFF}} &
  \multicolumn{1}{c|}{\cellcolor[HTML]{FFFFFF}} &
  \multicolumn{1}{c|}{\cellcolor[HTML]{FFFFFF}3} &
  \multicolumn{1}{c|}{\cellcolor[HTML]{FFFFFF}3} &
  \multicolumn{1}{c|}{\cellcolor[HTML]{FFFFFF}0.7741} &
  \multicolumn{1}{c|}{\cellcolor[HTML]{FFFFFF}0.6321} &
  \multicolumn{1}{c|}{\cellcolor[HTML]{FFFFFF}0.4737} &
  \multicolumn{1}{c|}{\cellcolor[HTML]{FFFFFF}} &
  \multicolumn{1}{c|}{\cellcolor[HTML]{FFFFFF}3.5381} &
  \multicolumn{1}{c|}{\cellcolor[HTML]{FFFFFF}2.70\%} &
  \multicolumn{1}{c|}{\cellcolor[HTML]{FFFFFF}3.60\%} \\ \cline{3-7} \cline{9-11} 
\rowcolor[HTML]{FFFFFF} 
\multicolumn{1}{|c|}{\cellcolor[HTML]{FFFFFF}} &
  \multicolumn{1}{c|}{\cellcolor[HTML]{FFFFFF}} &
  \multicolumn{1}{c|}{\cellcolor[HTML]{FFFFFF}3} &
  \multicolumn{1}{c|}{\cellcolor[HTML]{FFFFFF}3} &
  \multicolumn{1}{c|}{\cellcolor[HTML]{FFFFFF}0.7872} &
  \multicolumn{1}{c|}{\cellcolor[HTML]{FFFFFF}0.6321} &
  \multicolumn{1}{c|}{\cellcolor[HTML]{FFFFFF}0.4737} &
  \multicolumn{1}{c|}{\cellcolor[HTML]{FFFFFF}} &
  \multicolumn{1}{c|}{\cellcolor[HTML]{FFFFFF}2.1682} &
  \multicolumn{1}{c|}{\cellcolor[HTML]{FFFFFF}3.30\%} &
  \multicolumn{1}{c|}{\cellcolor[HTML]{FFFFFF}3.60\%} \\ \cline{3-7} \cline{9-11} 
\rowcolor[HTML]{EFEFEF} 
\multicolumn{1}{|c|}{\cellcolor[HTML]{FFFFFF}} &
  \multicolumn{1}{c|}{\cellcolor[HTML]{FFFFFF}} &
  \multicolumn{1}{c|}{\cellcolor[HTML]{EFEFEF}2} &
  \multicolumn{1}{c|}{\cellcolor[HTML]{EFEFEF}2} &
  \multicolumn{1}{c|}{\cellcolor[HTML]{EFEFEF}0.8288} &
  \multicolumn{1}{c|}{\cellcolor[HTML]{EFEFEF}0.4866} &
  \multicolumn{1}{c|}{\cellcolor[HTML]{EFEFEF}0.4211} &
  \multicolumn{1}{c|}{\cellcolor[HTML]{FFFFFF}} &
  \multicolumn{1}{c|}{\cellcolor[HTML]{EFEFEF}11.3214} &
  \multicolumn{1}{c|}{\cellcolor[HTML]{EFEFEF}2.40\%} &
  \multicolumn{1}{c|}{\cellcolor[HTML]{EFEFEF}3.60\%} \\ \cline{3-7} \cline{9-11} 
\rowcolor[HTML]{EFEFEF} 
\multicolumn{1}{|c|}{\cellcolor[HTML]{FFFFFF}} &
  \multicolumn{1}{c|}{\cellcolor[HTML]{FFFFFF}} &
  \multicolumn{1}{c|}{\cellcolor[HTML]{EFEFEF}{\color[HTML]{000000} \textbf{3}}} &
  \multicolumn{1}{c|}{\cellcolor[HTML]{EFEFEF}{\color[HTML]{000000} \textbf{3}}} &
  \multicolumn{1}{c|}{\cellcolor[HTML]{EFEFEF}{\color[HTML]{000000} \textbf{0.7937}}} &
  \multicolumn{1}{c|}{\cellcolor[HTML]{EFEFEF}{\color[HTML]{000000} \textbf{0.6321}}} &
  \multicolumn{1}{c|}{\cellcolor[HTML]{EFEFEF}{\color[HTML]{000000} \textbf{0.5263}}} &
  \multicolumn{1}{c|}{\cellcolor[HTML]{FFFFFF}} &
  \multicolumn{1}{c|}{\cellcolor[HTML]{EFEFEF}{\color[HTML]{000000} \textbf{13.8917}}} &
  \multicolumn{1}{c|}{\cellcolor[HTML]{EFEFEF}{\color[HTML]{000000} \textbf{2.00\%}}} &
  \multicolumn{1}{c|}{\cellcolor[HTML]{EFEFEF}{\color[HTML]{000000} \textbf{3.60\%}}} \\ \cline{3-7} \cline{9-11} 
\rowcolor[HTML]{FFFFFF} 
\multicolumn{1}{|c|}{\multirow{-5}{*}{\cellcolor[HTML]{FFFFFF}\textbf{Gpt-4o}}} &
  \multicolumn{1}{c|}{\multirow{-5}{*}{\cellcolor[HTML]{FFFFFF}Low}} &
  \multicolumn{1}{c|}{\cellcolor[HTML]{FFFFFF}3} &
  \multicolumn{1}{c|}{\cellcolor[HTML]{FFFFFF}3} &
  \multicolumn{1}{c|}{\cellcolor[HTML]{FFFFFF}0.7938} &
  \multicolumn{1}{c|}{\cellcolor[HTML]{FFFFFF}0.6321} &
  \multicolumn{1}{c|}{\cellcolor[HTML]{FFFFFF}0.4211} &
  \multicolumn{1}{c|}{\cellcolor[HTML]{FFFFFF}} &
  \multicolumn{1}{c|}{\cellcolor[HTML]{FFFFFF}11.6523} &
  \multicolumn{1}{c|}{\cellcolor[HTML]{FFFFFF}3.30\%} &
  \multicolumn{1}{c|}{\cellcolor[HTML]{FFFFFF}3.60\%} \\ \cline{1-7} \cline{9-11} 
\rowcolor[HTML]{FFFFFF} 
\multicolumn{7}{c}{\cellcolor[HTML]{FFFFFF}} &
  \multicolumn{1}{c|}{\multirow{-8}{*}{\cellcolor[HTML]{FFFFFF}\textbf{}}} &
  \multicolumn{3}{c}{\cellcolor[HTML]{FFFFFF}} \\ \hline
\rowcolor[HTML]{FFFFFF} 
\multicolumn{11}{c}{\cellcolor[HTML]{FFFFFF}\textbf{After Learning}} \\ \hline
\rowcolor[HTML]{FFFFFF} 
\multicolumn{4}{c|}{\cellcolor[HTML]{FFFFFF}\textbf{}} &
  \multicolumn{3}{c|}{\cellcolor[HTML]{FFFFFF}\textbf{Metrics}} &
  \multicolumn{1}{c|}{\cellcolor[HTML]{FFFFFF}} &
  \multicolumn{3}{c|}{\cellcolor[HTML]{FFFFFF}\textbf{Cloud Server}} \\ \cline{1-7} \cline{9-11} 
\rowcolor[HTML]{FFFFFF} 
\multicolumn{1}{|c|}{\cellcolor[HTML]{FFFFFF}\textbf{\begin{tabular}[c]{@{}c@{}}Model\\ Name\end{tabular}}} &
  \multicolumn{1}{c|}{\cellcolor[HTML]{FFFFFF}\textbf{\begin{tabular}[c]{@{}c@{}}Urgency\\ Level\end{tabular}}} &
  \multicolumn{1}{c|}{\cellcolor[HTML]{FFFFFF}\textbf{\begin{tabular}[c]{@{}c@{}}LM\\ Call\\ Count\end{tabular}}} &
  \multicolumn{1}{c|}{\cellcolor[HTML]{FFFFFF}\textbf{\begin{tabular}[c]{@{}c@{}}Hierarchy\\ Depth\\ Count\end{tabular}}} &
  \multicolumn{1}{c|}{\cellcolor[HTML]{FFFFFF}\textbf{\begin{tabular}[c]{@{}c@{}}Similarity\\ Score\end{tabular}}} &
  \multicolumn{1}{c|}{\cellcolor[HTML]{FFFFFF}\textbf{\begin{tabular}[c]{@{}c@{}}LM\\ Call\\ Usage\\ Cost\end{tabular}}} &
  \multicolumn{1}{c|}{\cellcolor[HTML]{FFFFFF}\textbf{\begin{tabular}[c]{@{}c@{}}Precision\\ Score\end{tabular}}} &
  \multicolumn{1}{c|}{\cellcolor[HTML]{FFFFFF}} &
  \multicolumn{1}{c|}{\cellcolor[HTML]{FFFFFF}\textbf{\begin{tabular}[c]{@{}c@{}}Elapsed\\ Time\end{tabular}}} &
  \multicolumn{1}{c|}{\cellcolor[HTML]{FFFFFF}\textbf{\begin{tabular}[c]{@{}c@{}}CPU\\ Utilization\end{tabular}}} &
  \multicolumn{1}{c|}{\cellcolor[HTML]{FFFFFF}\textbf{\begin{tabular}[c]{@{}c@{}}Memory\\ Utilization\end{tabular}}} \\ \cline{1-7} \cline{9-11} 
\rowcolor[HTML]{FFFFFF} 
\multicolumn{1}{|c|}{\cellcolor[HTML]{FFFFFF}} &
  \multicolumn{1}{c|}{\cellcolor[HTML]{FFFFFF}} &
  \multicolumn{1}{c|}{\cellcolor[HTML]{FFFFFF}3} &
  \multicolumn{1}{c|}{\cellcolor[HTML]{FFFFFF}3} &
  \multicolumn{1}{c|}{\cellcolor[HTML]{FFFFFF}0.8027} &
  \multicolumn{1}{c|}{\cellcolor[HTML]{FFFFFF}0.6321} &
  \multicolumn{1}{c|}{\cellcolor[HTML]{FFFFFF}0.3673} &
  \multicolumn{1}{c|}{\cellcolor[HTML]{FFFFFF}} &
  \multicolumn{1}{c|}{\cellcolor[HTML]{FFFFFF}16.5101} &
  \multicolumn{1}{c|}{\cellcolor[HTML]{FFFFFF}4.60\%} &
  \multicolumn{1}{c|}{\cellcolor[HTML]{FFFFFF}3.40\%} \\ \cline{3-7} \cline{9-11} 
\rowcolor[HTML]{EFEFEF} 
\multicolumn{1}{|c|}{\cellcolor[HTML]{FFFFFF}} &
  \multicolumn{1}{c|}{\cellcolor[HTML]{FFFFFF}} &
  \multicolumn{1}{c|}{\cellcolor[HTML]{EFEFEF}{\color[HTML]{000000} \textbf{3}}} &
  \multicolumn{1}{c|}{\cellcolor[HTML]{EFEFEF}{\color[HTML]{000000} \textbf{3}}} &
  \multicolumn{1}{c|}{\cellcolor[HTML]{EFEFEF}{\color[HTML]{000000} \textbf{0.8421}}} &
  \multicolumn{1}{c|}{\cellcolor[HTML]{EFEFEF}{\color[HTML]{000000} \textbf{0.6321}}} &
  \multicolumn{1}{c|}{\cellcolor[HTML]{EFEFEF}{\color[HTML]{000000} \textbf{0.3636}}} &
  \multicolumn{1}{c|}{\cellcolor[HTML]{FFFFFF}} &
  \multicolumn{1}{c|}{\cellcolor[HTML]{EFEFEF}{\color[HTML]{000000} \textbf{23.8452}}} &
  \multicolumn{1}{c|}{\cellcolor[HTML]{EFEFEF}{\color[HTML]{000000} \textbf{3.40\%}}} &
  \multicolumn{1}{c|}{\cellcolor[HTML]{EFEFEF}{\color[HTML]{000000} \textbf{3.40\%}}} \\ \cline{3-7} \cline{9-11} 
\rowcolor[HTML]{FFFFFF} 
\multicolumn{1}{|c|}{\cellcolor[HTML]{FFFFFF}} &
  \multicolumn{1}{c|}{\cellcolor[HTML]{FFFFFF}} &
  \multicolumn{1}{c|}{\cellcolor[HTML]{FFFFFF}3} &
  \multicolumn{1}{c|}{\cellcolor[HTML]{FFFFFF}3} &
  \multicolumn{1}{c|}{\cellcolor[HTML]{FFFFFF}0.8048} &
  \multicolumn{1}{c|}{\cellcolor[HTML]{FFFFFF}0.6321} &
  \multicolumn{1}{c|}{\cellcolor[HTML]{FFFFFF}0.2909} &
  \multicolumn{1}{c|}{\cellcolor[HTML]{FFFFFF}} &
  \multicolumn{1}{c|}{\cellcolor[HTML]{FFFFFF}30.2243} &
  \multicolumn{1}{c|}{\cellcolor[HTML]{FFFFFF}3.50\%} &
  \multicolumn{1}{c|}{\cellcolor[HTML]{FFFFFF}3.60\%} \\ \cline{3-7} \cline{9-11} 
\rowcolor[HTML]{FFFFFF} 
\multicolumn{1}{|c|}{\cellcolor[HTML]{FFFFFF}} &
  \multicolumn{1}{c|}{\cellcolor[HTML]{FFFFFF}} &
  \multicolumn{1}{c|}{\cellcolor[HTML]{FFFFFF}3} &
  \multicolumn{1}{c|}{\cellcolor[HTML]{FFFFFF}2} &
  \multicolumn{1}{c|}{\cellcolor[HTML]{FFFFFF}0.8294} &
  \multicolumn{1}{c|}{\cellcolor[HTML]{FFFFFF}0.6321} &
  \multicolumn{1}{c|}{\cellcolor[HTML]{FFFFFF}0.3934} &
  \multicolumn{1}{c|}{\cellcolor[HTML]{FFFFFF}} &
  \multicolumn{1}{c|}{\cellcolor[HTML]{FFFFFF}19.5182} &
  \multicolumn{1}{c|}{\cellcolor[HTML]{FFFFFF}3.50\%} &
  \multicolumn{1}{c|}{\cellcolor[HTML]{FFFFFF}3.50\%} \\ \cline{3-7} \cline{9-11} 
\rowcolor[HTML]{FFFFFF} 
\multicolumn{1}{|c|}{\multirow{-5}{*}{\cellcolor[HTML]{FFFFFF}\textbf{Gpt-4o}}} &
  \multicolumn{1}{c|}{\multirow{-5}{*}{\cellcolor[HTML]{FFFFFF}Low}} &
  \multicolumn{1}{c|}{\cellcolor[HTML]{FFFFFF}3} &
  \multicolumn{1}{c|}{\cellcolor[HTML]{FFFFFF}2} &
  \multicolumn{1}{c|}{\cellcolor[HTML]{FFFFFF}0.8109} &
  \multicolumn{1}{c|}{\cellcolor[HTML]{FFFFFF}0.6321} &
  \multicolumn{1}{c|}{\cellcolor[HTML]{FFFFFF}0.3157} &
  \multicolumn{1}{c|}{\multirow{-7}{*}{\cellcolor[HTML]{FFFFFF}\textbf{}}} &
  \multicolumn{1}{c|}{\cellcolor[HTML]{FFFFFF}21.6117} &
  \multicolumn{1}{c|}{\cellcolor[HTML]{FFFFFF}3.70\%} &
  \multicolumn{1}{c|}{\cellcolor[HTML]{FFFFFF}3.40\%} \\ \cline{1-7} \cline{9-11} 
\end{tabular}

}
\end{table*}

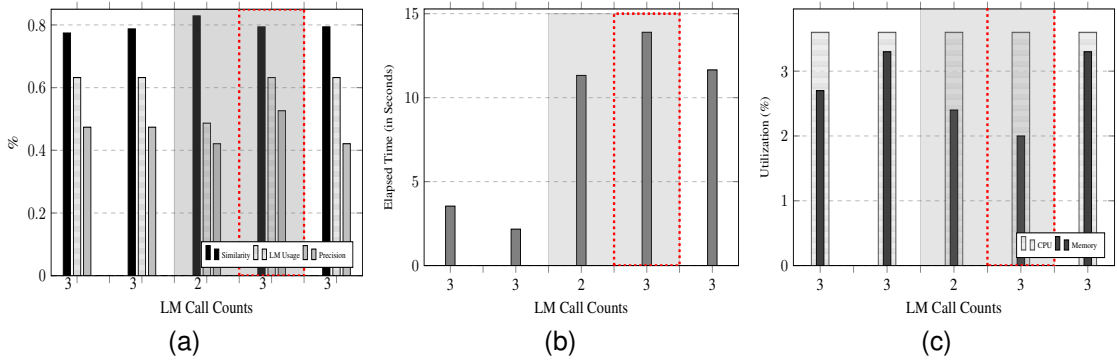
\begin{figure*}
	\centering
\subfloat[]{\resizebox{48mm}{42mm}{\begin{tikzpicture}
    \begin{axis}[
    ybar,
    width=0.56\textwidth,
    height=.42\textwidth,
            bar width=.18cm,
            symbolic x coords={1,1.5,2,2.5,3,3.5,4,4.5,5}, xtick={1,1.5,2,2.5,3,3.5,4,4.5,5},  
        xticklabels={3,\ , 3,\ , 2,\ , 3,\ , 3}, 	 %
            legend columns=3,ymin=0.0,ymax=0.85,
            legend pos=south east,legend style={font=\fontsize{5}{5}\selectfont},
	ylabel= \%, ylabel style={font=\fontsize{10}{10}\selectfont},ymajorgrids=true,   grid style=dashed, xlabel={LM Call Counts},
	 xticklabel style={rotate=0, anchor=east}, 
]
\addplot[fill=black] table[x=interval,y=Similarity]{\beforelearninga};
\addplot[fill=darkgray, pattern=horizontal lines light gray] table[x=interval,y=LMUsage]{\beforelearninga};
\addplot[fill=lightgray] table[x=interval,y=Precision]{\beforelearninga};
\draw[fill=gray, opacity=0.3] (axis cs:2.5,0) rectangle (axis cs:4.5,0.85); 
\draw[dotted,ultra thick, red](axis cs:3.5,0) rectangle (axis cs:4.5,0.85);
\legend{Similarity, LM Usage, Precision}
\end{axis}  
\end{tikzpicture} }\label{Fig6a}}
\ \ 
\subfloat[]{\resizebox{48mm}{42mm}{\begin{tikzpicture}
    \begin{axis}[
    ybar,
    width=0.56\textwidth,
    height=.42\textwidth,
            bar width=.23cm,
            symbolic x coords={1,1.5,2,2.5,3,3.5,4,4.5,5}, xtick={1,1.5,2,2.5,3,3.5,4,4.5,5},  
        xticklabels={3,\ , 3,\ , 2,\ , 3,\ , 3}, 	 %
            legend columns=3,ymin=0.0,
            legend pos=south east,legend style={font=\fontsize{5}{5}\selectfont},
	ylabel= Elapsed Time (in Seconds), ylabel style={font=\fontsize{8}{8}\selectfont},ymajorgrids=true,   grid style=dashed,xlabel={LM Call Counts},
]
\addplot[fill=gray] table[x=interval,y=Elapsed]{\beforelearninga};
\draw[fill=gray, opacity=0.2] (axis cs:2.5,0) rectangle (axis cs:4.5,15); 
\draw[dotted,ultra thick, red](axis cs:3.5,0) rectangle (axis cs:4.5,15);

\end{axis}  
\end{tikzpicture} }\label{Fig6b}}
\ \ 
\subfloat[]{\resizebox{48mm}{42mm}{\begin{tikzpicture}
    \begin{axis}[
    ybar,
    width=0.56\textwidth,
    height=.42\textwidth,
            bar width=.18cm,
            symbolic x coords={1,1.5,2,2.5,3,3.5,4,4.5,5}, xtick={1,1.5,2,2.5,3,3.5,4,4.5,5},  
        xticklabels={3,\ , 3,\ , 2,\ , 3,\ , 3}, 	       
            legend columns=3,ymin=0.0,
            legend pos=south east,legend style={font=\fontsize{5}{5}\selectfont},
	ylabel= Utilization (\%), ylabel style={font=\fontsize{8}{8}\selectfont},ymajorgrids=true,   grid style=dashed,xlabel={LM Call Counts},
]
\addplot[fill=black, pattern=horizontal lines light gray, bar width=0.42cm, opacity=0.7,bar shift=0.0pt] table[x=interval,y=Memory]{\beforelearninga};
\addplot[fill=darkgray,bar shift=0.0pt] table[x=interval,y=CPU]{\beforelearninga};
\draw[fill=gray, opacity=0.2] (axis cs:2.5,0) rectangle (axis cs:4.5,15); 
\draw[dotted,ultra thick, red](axis cs:3.5,0) rectangle (axis cs:4.5,15);
\legend{CPU, Memory}
\end{axis}  
\end{tikzpicture} }\label{Fig6c}}

\caption{Low-urgency Learning Performance Evaluation before learning \protect \subref{Fig6a} Response of Metrics \protect\subref{Fig6b} Elapsed time before learning \protect\subref{Fig6c} CPU and Memory Utilization}\label{Fig6-1}
\end{figure*}

\begin{figure*}
	\centering
\subfloat[]{\resizebox{48mm}{42mm}{\begin{tikzpicture}
    \begin{axis}[
    ybar,
    width=0.56\textwidth,
    height=.42\textwidth,
            bar width=.18cm,
            symbolic x coords={1,1.5,2,2.5,3,3.5,4,4.5,5}, xtick={1,1.5,2,2.5,3,3.5,4,4.5,5},  
        xticklabels={3,\ , 3,\ , 3,\ , 3,\ , 3}, 	 %
            legend columns=3,ymin=0.0,ymax=0.85,
            legend pos=south east,legend style={font=\fontsize{5}{5}\selectfont},
	ylabel= \%, ylabel style={font=\fontsize{10}{10}\selectfont},ymajorgrids=true,   grid style=dashed, xlabel={LM Call Counts},
	 xticklabel style={rotate=0, anchor=east}, 
]
\addplot[fill=black] table[x=interval,y=Similarity]{\beforelearningb};
\addplot[fill=darkgray, pattern=horizontal lines light gray] table[x=interval,y=LMUsage]{\beforelearningb};
\addplot[fill=lightgray] table[x=interval,y=Precision]{\beforelearningb};
\draw[fill=gray, opacity=0.3] (axis cs:1.5,0) rectangle (axis cs:2.5,1); 
\draw[dotted,ultra thick, red](axis cs:1.5,0) rectangle (axis cs:2.5,1);

\legend{Similarity, LM Usage, Precision}
\end{axis}  
\end{tikzpicture} }\label{Fig6d}}
\subfloat[]{\resizebox{48mm}{42mm}{\begin{tikzpicture}
    \begin{axis}[
    ybar,
    width=0.56\textwidth,
    height=.42\textwidth,
            bar width=.23cm,
            symbolic x coords={1,1.5,2,2.5,3,3.5,4,4.5,5}, xtick={1,1.5,2,2.5,3,3.5,4,4.5,5},  
        xticklabels={3,\ , 3,\ , 3,\ , 3,\ , 3}, 	 %
            legend columns=3,ymin=0.0,
            legend pos=south east,legend style={font=\fontsize{5}{5}\selectfont},
	ylabel= Elapsed Time (in Seconds), ylabel style={font=\fontsize{8}{8}\selectfont},ymajorgrids=true,   grid style=dashed,xlabel={LM Call Counts},
]
\addplot[fill=black] table[x=interval,y=Elapsed]{\beforelearningb};
\draw[fill=gray, opacity=0.3] (axis cs:1.5,0) rectangle (axis cs:2.5,30); 
\draw[dotted,ultra thick, red](axis cs:1.5,0) rectangle (axis cs:2.5,30);

\end{axis}  
\end{tikzpicture} }\label{Fig6e}}
\ \ 
\subfloat[]{\resizebox{48mm}{42mm}{\begin{tikzpicture}
    \begin{axis}[
    ybar,
    width=0.56\textwidth,
    height=.42\textwidth,
            bar width=.18cm,
            symbolic x coords={1,1.5,2,2.5,3,3.5,4,4.5,5}, xtick={1,1.5,2,2.5,3,3.5,4,4.5,5},  
        xticklabels={3,\ , 3,\ , 3,\ , 3,\ , 3}, 	 %
            legend columns=2,ymin=0.0,
            legend pos=north east,legend style={font=\fontsize{5}{5}\selectfont},
	ylabel= Utilization (\%), ylabel style={font=\fontsize{8}{8}\selectfont},ymajorgrids=true,   grid style=dashed,xlabel={LM Call Counts},
]
\addplot[fill=black, pattern=horizontal lines light gray, bar width=0.42cm, opacity=0.7,bar shift=0.0pt] table[x=interval,y=Memory]{\beforelearningb};
\addplot[fill=darkgray,bar shift=0.0pt] table[x=interval,y=CPU]{\beforelearningb};
\draw[fill=gray, opacity=0.3] (axis cs:1.5,0) rectangle (axis cs:2.5,30); 
\draw[dotted,ultra thick, red](axis cs:1.5,0) rectangle (axis cs:2.5,5);

\legend{CPU, Memory}
\end{axis}  
\end{tikzpicture} }\label{Fig6f}}
\caption{Low-urgency Learning Performance Evaluation after learning \protect \subref{Fig6d} Response of Metrics  \protect\subref{Fig6e} Elapsed time \protect\subref{Fig6f} CPU and Memory utilization}\label{Fig6-2}
\end{figure*}
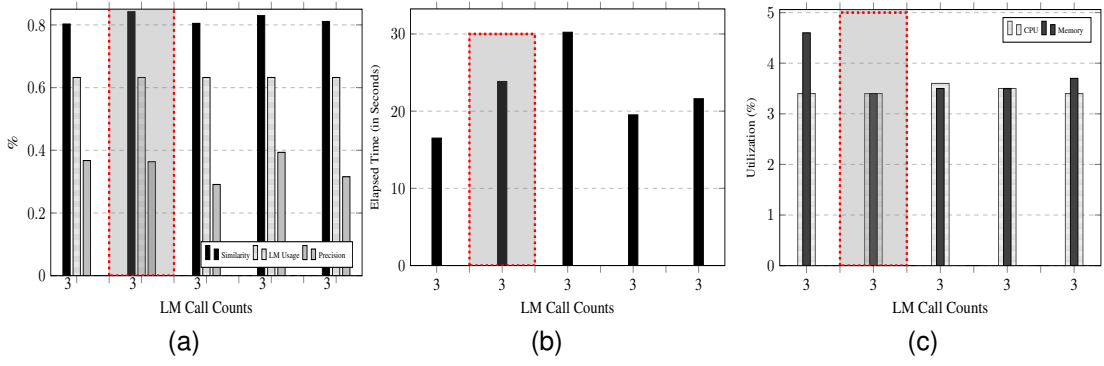

%% file: tables/table-monitor.tex
\begin{table*}[]
\centering
\caption{Environment Agent Performance Evaluation.}
\label{tab:monitor}
\resizebox{!}{.06\paperheight}{%

\begin{tabular}{|
>{\columncolor[HTML]{FFFFFF}}l |
>{\columncolor[HTML]{FFFFFF}}c 
>{\columncolor[HTML]{FFFFFF}}c 
>{\columncolor[HTML]{FFFFFF}}c |
>{\columncolor[HTML]{FFFFFF}}c |
>{\columncolor[HTML]{FFFFFF}}c 
>{\columncolor[HTML]{FFFFFF}}c 
>{\columncolor[HTML]{FFFFFF}}c |
>{\columncolor[HTML]{FFFFFF}}l |
>{\columncolor[HTML]{FFFFFF}}c |}
\cline{1-4} \cline{6-8} \cline{10-10}
\multicolumn{1}{|c|}{\cellcolor[HTML]{FFFFFF}} &
  \multicolumn{3}{c|}{\cellcolor[HTML]{FFFFFF}\textbf{Cloud Server}} &
  \cellcolor[HTML]{FFFFFF} &
  \multicolumn{3}{c|}{\cellcolor[HTML]{FFFFFF}\textbf{Edge Device}} &
  \cellcolor[HTML]{FFFFFF} &
  \cellcolor[HTML]{FFFFFF} \\ \cline{2-4} \cline{6-8}
\multicolumn{1}{|c|}{\multirow{-2}{*}{\cellcolor[HTML]{FFFFFF}\textbf{Model Name}}} &
  \multicolumn{1}{c|}{\cellcolor[HTML]{FFFFFF}\textbf{\begin{tabular}[c]{@{}c@{}}Elapsed\\ Time\end{tabular}}} &
  \multicolumn{1}{c|}{\cellcolor[HTML]{FFFFFF}\textbf{\begin{tabular}[c]{@{}c@{}}CPU\\ Utilization\end{tabular}}} &
  \textbf{\begin{tabular}[c]{@{}c@{}}Memory\\ Utilization\end{tabular}} &
  \cellcolor[HTML]{FFFFFF} &
  \multicolumn{1}{c|}{\cellcolor[HTML]{FFFFFF}\textbf{\begin{tabular}[c]{@{}c@{}}Elapsed\\ Time\end{tabular}}} &
  \multicolumn{1}{c|}{\cellcolor[HTML]{FFFFFF}\textbf{\begin{tabular}[c]{@{}c@{}}CPU\\ Utilization\end{tabular}}} &
  \textbf{\begin{tabular}[c]{@{}c@{}}Memory\\ Utilization\end{tabular}} &
  \cellcolor[HTML]{FFFFFF} &
  \multirow{-2}{*}{\cellcolor[HTML]{FFFFFF}\textbf{Accuracy}} \\ \cline{1-4} \cline{6-8} \cline{10-10} 
\textbf{Gemini-1.5 flash} &
  \multicolumn{1}{c|}{\cellcolor[HTML]{FFFFFF}2.7076} &
  \multicolumn{1}{c|}{\cellcolor[HTML]{FFFFFF}5.60\%} &
  3.70\% &
  \cellcolor[HTML]{FFFFFF} &
  \multicolumn{1}{c|}{\cellcolor[HTML]{FFFFFF}3.5179} &
  \multicolumn{1}{c|}{\cellcolor[HTML]{FFFFFF}14.40\%} &
  35.70\% &
  \cellcolor[HTML]{FFFFFF} &
  1 \\ \cline{1-4} \cline{6-8} \cline{10-10} 
\textbf{command-r7b} &
  \multicolumn{1}{c|}{\cellcolor[HTML]{FFFFFF}220.2311} &
  \multicolumn{1}{c|}{\cellcolor[HTML]{FFFFFF}35.00\%} &
  15.40\% &
  \cellcolor[HTML]{FFFFFF} &
  \multicolumn{1}{c|}{\cellcolor[HTML]{FFFFFF}192.1994} &
  \multicolumn{1}{c|}{\cellcolor[HTML]{FFFFFF}57.60\%} &
  64.90\% &
  \cellcolor[HTML]{FFFFFF} &
  1 \\ \cline{1-4} \cline{6-8} \cline{10-10} 
\textbf{Claude 3.5 Sonnet} &
  \multicolumn{1}{c|}{\cellcolor[HTML]{FFFFFF}298.506} &
  \multicolumn{1}{c|}{\cellcolor[HTML]{FFFFFF}45.20\%} &
  15.10\% &
  \cellcolor[HTML]{FFFFFF} &
  \multicolumn{1}{c|}{\cellcolor[HTML]{FFFFFF}284.1143} &
  \multicolumn{1}{c|}{\cellcolor[HTML]{FFFFFF}57.30\%} &
  61.70\% &
  \cellcolor[HTML]{FFFFFF} &
  0 \\ \cline{1-4} \cline{6-8} \cline{10-10} 
\textbf{Mistral} &
  \multicolumn{1}{c|}{\cellcolor[HTML]{FFFFFF}251.7816} &
  \multicolumn{1}{c|}{\cellcolor[HTML]{FFFFFF}44.20\%} &
  13.70\% &
  \cellcolor[HTML]{FFFFFF} &
  \multicolumn{1}{c|}{\cellcolor[HTML]{FFFFFF}242.4539} &
  \multicolumn{1}{c|}{\cellcolor[HTML]{FFFFFF}57.10\%} &
  57.80\% &
  \cellcolor[HTML]{FFFFFF} &
  0.5 \\ \cline{1-4} \cline{6-8} \cline{10-10} 
\textbf{granite3.1-MoE} &
  \multicolumn{1}{c|}{\cellcolor[HTML]{FFFFFF}38.577} &
  \multicolumn{1}{c|}{\cellcolor[HTML]{FFFFFF}10.20\%} &
  9.00\% &
  \cellcolor[HTML]{FFFFFF} &
  \multicolumn{1}{c|}{\cellcolor[HTML]{FFFFFF}35.5997} &
  \multicolumn{1}{c|}{\cellcolor[HTML]{FFFFFF}52.90\%} &
  77.30\% &
  \cellcolor[HTML]{FFFFFF} &
  0 \\ \cline{1-4} \cline{6-8} \cline{10-10} 
\textbf{Gpt-4o} &
  \multicolumn{1}{c|}{\cellcolor[HTML]{FFFFFF}8.3605} &
  \multicolumn{1}{c|}{\cellcolor[HTML]{FFFFFF}2.50\%} &
  3.80\% &
  \multirow{-8}{*}{\cellcolor[HTML]{FFFFFF}\textbf{}} &
  \multicolumn{1}{c|}{\cellcolor[HTML]{FFFFFF}10.1599} &
  \multicolumn{1}{c|}{\cellcolor[HTML]{FFFFFF}10.80\%} &
  36.50\% &
  \multirow{-8}{*}{\cellcolor[HTML]{FFFFFF}} &
  1 \\ \cline{1-4} \cline{6-8} \cline{10-10} 
\end{tabular}

}
\end{table*}

\begin{figure*}
	\centering
	\subfloat[]{\resizebox{38mm}{32mm}{\begin{tikzpicture}
    \begin{axis}[
    ybar,
    width=0.56\textwidth,
    height=.42\textwidth,
            bar width=.18cm,
            symbolic x coords={Gemini-1.5, command-r7b, Claude 3.5, Mistral, Granite3.1, Gpt-4o}, xtick=data, 
            legend columns=1,
            legend pos=north east,legend style={font=\fontsize{8}{8}\selectfont},
	ylabel=Time Elapsed (seconds), ylabel style={font=\fontsize{10}{10}\selectfont},ymajorgrids=true,   grid style=dashed,
	 xticklabel style={rotate=90, anchor=east}, 
]
\addplot[fill=black] table[x=interval,y=Edge]{\EnvAgenta};
\addplot[fill=lightgray] table[x=interval,y=Cloud]{\EnvAgenta};
\legend{Cloud,Edge}
\end{axis}  
\end{tikzpicture} }}\label{Fig7a}
\ \ \ \ 
\subfloat[]{\resizebox{38mm}{32mm}{\begin{tikzpicture}
    \begin{axis}[
    ybar,
    width=0.56\textwidth,
    height=.42\textwidth,
            bar width=.18cm,
            symbolic x coords={Gemini-1.5, command-r7b, Claude 3.5, Mistral, Granite3.1, Gpt-4o}, xtick=data, 
            legend columns=1,
            legend pos=north east,legend style={font=\fontsize{8}{8}\selectfont},
	ylabel=CPU Utilization (\%), ylabel style={font=\fontsize{10}{10}\selectfont},ymajorgrids=true,   grid style=dashed,
	 xticklabel style={rotate=90, anchor=east}, 
]
\addplot[fill=black] table[x=interval,y=Edge]{\EnvAgentb};
\addplot[fill=lightgray] table[x=interval,y=Cloud]{\EnvAgentb};
\legend{Cloud,Edge}
\end{axis}  
\end{tikzpicture} }}\label{Fig7b}
\ \ \ \ 
\subfloat[]{\resizebox{38mm}{32mm}{\begin{tikzpicture}
    \begin{axis}[
    ybar,
    width=0.56\textwidth,
    height=.42\textwidth,
            bar width=.18cm,
            symbolic x coords={Gemini-1.5, command-r7b, Claude 3.5, Mistral, Granite3.1, Gpt-4o}, xtick=data, 
            legend columns=1,
            legend pos=north east,legend style={font=\fontsize{8}{8}\selectfont},
	ylabel= Memory Utilization (\%), ylabel style={font=\fontsize{10}{10}\selectfont},ymajorgrids=true,   grid style=dashed,
	 xticklabel style={rotate=90, anchor=east}, 
]
\addplot[fill=black] table[x=interval,y=Edge]{\EnvAgentc};
\addplot[fill=lightgray] table[x=interval,y=Cloud]{\EnvAgentc};
\legend{Cloud,Edge}
\end{axis}  
\end{tikzpicture} }}\label{Fig7c}
\ \ \ \ 
\subfloat[]{\resizebox{38mm}{32mm}{\begin{tikzpicture}
    \begin{axis}[
    width=0.56\textwidth,
    height=.42\textwidth,
            bar width=.18cm,
            symbolic x coords={Gemini-1.5, command-r7b, Claude 3.5, Mistral, Granite3.1, Gpt-4o}, xtick=data, 
            legend columns=1,
            legend pos=north east,legend style={font=\fontsize{8}{8}\selectfont},
	ylabel= Accuracy, ylabel style={font=\fontsize{10}{10}\selectfont},ymajorgrids=true,   grid style=dashed,
	 xticklabel style={rotate=90, anchor=east}, 
]
\addplot+[jump mark mid, color=black, mark size=4pt, mark options={mark color=darkgray}]  table[x=interval,y=Accuracy]{\EnvAgentd};
\end{axis}  
\end{tikzpicture} }}\label{Fig7d}
\caption{Environment Agent Performance Evaluation.}\label{Fig7}
\end{figure*}
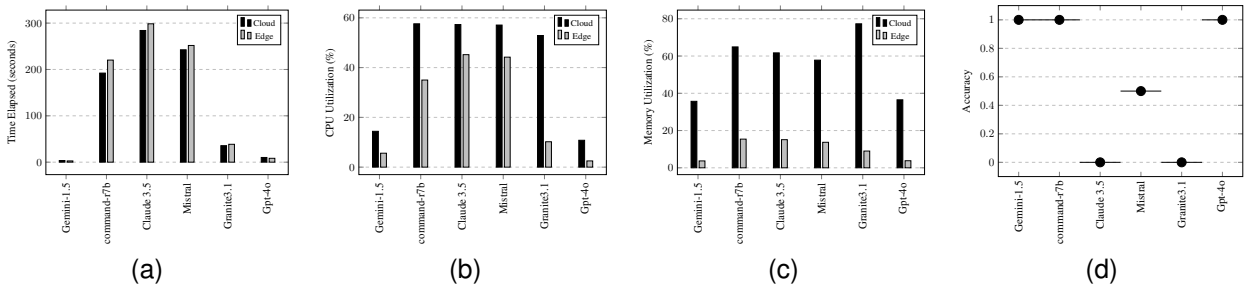

%% file: sections/Discussion.tex
\section{Discussion}\label{sec:Discussion}
This section discusses the main insights from the experimental results and the techniques used in the \texttt{\textbf{UserCentrix}} framework. It highlights how the framework’s design choices influence performance, efficiency, and adaptability in smart spaces, and explores both the strengths and areas for improvement. The discussion is organized through the following key questions:

\textbf{Q1: \textit{How does on-device AI, incorporating CBR and memory recall, influence intent precision and personalization over time?}}\\
The results, as illustrated in Table~\ref{tab:user}, demonstrate that the LM-powered AI agent effectively extracts plans and preferences while retrieving missing information from memory based on similar past cases to fulfill user intent with high accuracy. This capability facilitates more personalized responses, thereby enhancing user experience and aligning system outputs more closely with individual preferences.

\textbf{Q2: \textit{How does auto-scaling the number of AI agents (LM calls) affect user experience?}}\\
Higher LM call counts may increase execution time and resource consumption, potentially delaying responses in latency-sensitive scenarios. To preserve a high-quality user experience, \texttt{\textbf{UserCentrix}} balances decision quality and response speed through urgency-based agent selection and pareto filtering. A classification agent, guided by user intent value, activates urgency-aware scaling to ensure timely responses, adjusting the number of LM calls according to required reasoning depth while accounting for parallel execution. As shown in Table~\ref{tab:pareto}, pareto optimization identifies configurations that sustain high accuracy while minimizing unnecessary calls.

\textbf{Q3: \textit{How does in-context learning contribute to improving the solution quality of an AI agent?}}\\
The results, as illustrated in Table~\ref{tab:learn}, indicate that in-context learning enables agents to generalize from prior contextual hints without retraining, thereby enhancing performance on new intents through prompt evolution with previously acquired hints. For \textit{gpt-4o}, the low-urgency agent leveraged highly similar prior cases stored in memory to generate solutions with high similarity $\approx$6.1\%.

\textbf{Q4: \textit{How did the environment AI agent’s corrective commands influence the end-user experience and overall system performance?}}\\
As shown in Table~\ref{tab:monitor}, the LM-powered environment agent maintains a strong balance between execution speed, low resource consumption, and high accuracy. This balance enables efficient real-time monitoring and timely generation of corrective commands, thereby improving user experience through responsive system behavior while maintaining minimal computational overhead and ensuring fault tolerance without the need for human intervention.

\textbf{Q5: \textit{What are the main limitations and future directions of the proposed framework?}}\\
Although the framework integrates evaluation mechanisms within the agents’ reasoning processes, the correctness of both intermediate reasoning steps and final outputs remains strongly dependent on the accuracy and reasoning capabilities of the underlying LMs. This reliance may affect the consistency and reliability of decision-making. A promising future direction is dynamic agent optimization through real-time discovery, selection, and reconfiguration of agents and models based on intent complexity and domain relevance.

%% file: sections/Conclusion.tex
\section{Conclusion}\label{sec:Conclusion}
This paper introduced \texttt{\textbf{UserCentrix}}, a hybrid agentic orchestration framework that integrates personalized LM agents with scalable memory and self-evaluation, in-context learning for prompt evolution, urgency-aware auto-scaling, and RPs-based negotiation, and command scheduling and regulation. The framework dynamically aligns system behavior with user intent while balancing reasoning depth, response latency, and resource utilization across both cloud and edge environments. The experimental results demonstrate that appropriate model selection within \texttt{\textbf{UserCentrix}} enables substantial gains in computational efficiency and reasoning quality. By integrating agentic orchestration with performance-aware model selection and learning-driven optimization, the framework advances the intelligence, adaptability, and efficiency of AI-driven smart spaces without human intervention.

%% file: sections/Acknowledgments.tex
\section{Acknowledgments}\label{sec:acknowledges}
This research is funded by the Research Council of Finland through the evoS3 (Grant Number 362594) and the 6G Flagship (Grant Number 369116) projects, and by Business Finland through the Neural Pub/Sub research project (Diary Number 8754/31/2022).

%% file: references.bib
@inproceedings{snell2024scaling,
title={Scaling {LLM} Test-Time Compute Optimally Can be More Effective than Scaling Parameters for Reasoning},
author={Charlie Victor Snell and Jaehoon Lee and Kelvin Xu and Aviral Kumar},
booktitle={The Thirteenth International Conference on Learning Representations},
year={2025},
url={https://openreview.net/forum?id=4FWAwZtd2n}
}

@inproceedings{christakopoulou2024agents,
title={Agents Thinking Fast and Slow: A Talker-Reasoner Architecture},
author={Konstantina Christakopoulou and Shibl Mourad and Maja Mataric},
booktitle={NeurIPS 2024 Workshop on Open-World Agents},
year={2024},
url={https://openreview.net/forum?id=xPhcP6rbI4}
}

@inproceedings{wu2024thinking,
title={Thinking {LLM}s: General Instruction Following with Thought Generation},
author={Tianhao Wu and Janice Lan and Weizhe Yuan and Jiantao Jiao and Jason E Weston and Sainbayar Sukhbaatar},
booktitle={Forty-second International Conference on Machine Learning},
year={2025},
url={https://openreview.net/forum?id=z6SrgYCdey}
}

@inproceedings{hou2024my,
  title={" my agent understands me better": Integrating dynamic human-like memory recall and consolidation in llm-based agents},
  author={Hou, Yuki and Tamoto, Haruki and Miyashita, Homei},
  booktitle={Extended Abstracts of the CHI Conference on Human Factors in Computing Systems},
  pages={1--7},
  year={2024}
}

@inproceedings{10.1145/3706599.3719853,
author = {Lee, Christine P. and Choi, Jihye and Mutlu, Bilge},
title = {MAP: Multi-user Personalization with Collaborative LLM-powered Agents},
year = {2025},
isbn = {9798400713958},
publisher = {Association for Computing Machinery},
address = {New York, NY, USA},
url = {https://doi.org/10.1145/3706599.3719853},
doi = {10.1145/3706599.3719853},
booktitle = {Proceedings of the Extended Abstracts of the CHI Conference on Human Factors in Computing Systems},
articleno = {384},
numpages = {11},
location = {
},
series = {CHI EA '25}
}

@inproceedings{qi2024mutual,
title={Mutual Reasoning Makes Smaller {LLM}s Stronger Problem-Solver},
author={Zhenting Qi and Mingyuan MA and Jiahang Xu and Li Lyna Zhang and Fan Yang and Mao Yang},
booktitle={The Thirteenth International Conference on Learning Representations},
year={2025},
url={https://openreview.net/forum?id=6aHUmotXaw}
}

@inproceedings{zelikman2403quiet,
title={Quiet-{ST}aR: Language Models Can Teach Themselves to Think Before Speaking},
author={Eric Zelikman and Georges Raif Harik and Yijia Shao and Varuna Jayasiri and Nick Haber and Noah Goodman},
booktitle={First Conference on Language Modeling},
year={2024},
url={https://openreview.net/forum?id=oRXPiSOGH9}
}

@inproceedings{
wang2406mixture,
title={Mixture-of-Agents Enhances Large Language Model Capabilities},
author={Junlin Wang and Jue WANG and Ben Athiwaratkun and Ce Zhang and James Zou},
booktitle={The Thirteenth International Conference on Learning Representations},
year={2025},
url={https://openreview.net/forum?id=h0ZfDIrj7T}
}

@book{russell2020artificial,
  title={Artificial intelligence: a Modern Approach, Fourth Edition},
  author={Russell, Stuart J and Norvig, Peter},
  year={2020},
  publisher={Pearson}
}

@inproceedings{guo2024embodied,
title={Embodied {LLM} Agents Learn to Cooperate in Organized Teams},
author={Xudong Guo and Kaixuan Huang and Jiale Liu and Wenhui Fan and Natalia V{\'e}lez and Qingyun Wu and Huazheng Wang and Thomas L. Griffiths and Mengdi Wang},
booktitle={Language Gamification - NeurIPS 2024 Workshop},
year={2024},
url={https://openreview.net/forum?id=VKlrzygQlT}
}

@article{saleh2024follow,
author = {Saleh, Alaa and Donta, Praveen Kumar and Morabito, Roberto and Hossein Motlagh, Naser and Tarkoma, Sasu and Lovén, Lauri},
year = {2025},
month = {02},
pages = {1-10},
title = {{Follow-Me AI: Energy-Efficient User Interaction with Smart Environments}},
journal = {IEEE Pervasive Computing},
doi = {10.1109/MPRV.2025.3539421}
}

@article{saleh2023pub,
author = {Saleh, Alaa and Morabito, Roberto and Dustdar, Schahram and Tarkoma, Sasu and Pirttikangas, Susanna and Lov\'{e}n, Lauri},
title = {Towards Message Brokers for Generative AI: Survey, Challenges, and Opportunities},
year = {2025},
publisher = {Association for Computing Machinery},
address = {New York, NY, USA},
issn = {0360-0300},
doi = {10.1145/3742891},
note = {Just Accepted},
journal = {ACM Comput. Surv.},
month = jun
}

@misc{gu2024survey,
      title={A Survey on LLM-as-a-Judge}, 
      author={Jiawei Gu and Xuhui Jiang and Zhichao Shi and Hexiang Tan and Xuehao Zhai and Chengjin Xu and Wei Li and Yinghan Shen and Shengjie Ma and Honghao Liu and Saizhuo Wang and Kun Zhang and Yuanzhuo Wang and Wen Gao and Lionel Ni and Jian Guo},
      year={2025},
      eprint={2411.15594},
      archivePrefix={arXiv},
      primaryClass={cs.CL},
      url={https://arxiv.org/abs/2411.15594}, 
}

@article{WU2022364,
title = {A survey of human-in-the-loop for machine learning},
journal = {Future Generation Computer Systems},
volume = {135},
pages = {364-381},
year = {2022},
issn = {0167-739X},
doi = {https://doi.org/10.1016/j.future.2022.05.014},
url = {https://www.sciencedirect.com/science/article/pii/S0167739X22001790},
author = {Xingjiao Wu and Luwei Xiao and Yixuan Sun and Junhang Zhang and Tianlong Ma and Liang He},
}

@article{meuser2024revisiting,
  title={{Revisiting Edge AI: Opportunities and Challenges}},
  author={Meuser, Tobias and Lov{\'e}n, Lauri and Bhuyan, Monowar and Patil, Shishir G and Dustdar, Schahram and Aral, Atakan and Bayhan, Suzan and Becker, Christian and de Lara, Eyal and Ding, Aaron Yi and others},
  journal={IEEE Internet Computing},
  volume={28},
  number={4},
  pages={49--59},
  year={2024},
  publisher={IEEE}
}

@article{shen2024large,
  title={Large language models empowered autonomous edge AI for connected intelligence},
  author={Shen, Yifei and Shao, Jiawei and Zhang, Xinjie and Lin, Zehong and Pan, Hao and Li, Dongsheng and Zhang, Jun and Letaief, Khaled B},
  journal={IEEE Communications Magazine},
  year={2024},
  publisher={IEEE}
}

@article{zhang2024edgeshard,
  title={Edgeshard: Efficient llm inference via collaborative edge computing},
  author={Zhang, Mingjin and Shen, Xiaoming and Cao, Jiannong and Cui, Zeyang and Jiang, Shan},
  journal={IEEE Internet of Things Journal},
  year={2024},
  publisher={IEEE}
}

@inproceedings{hao2024hybrid,
  title={Hybrid slm and llm for edge-cloud collaborative inference},
  author={Hao, Zixu and Jiang, Huiqiang and Jiang, Shiqi and Ren, Ju and Cao, Ting},
  booktitle={Proceedings of the Workshop on Edge and Mobile Foundation Models},
  pages={36--41},
  year={2024}
}

@inproceedings{yu2024edge,
  title={Edge-llm: Enabling efficient large language model adaptation on edge devices via unified compression and adaptive layer voting},
  author={Yu, Zhongzhi and Wang, Zheng and Li, Yuhan and Gao, Ruijie and Zhou, Xiaoya and Bommu, Sreenidhi Reddy and Zhao, Yang and Lin, Yingyan},
  booktitle={Proceedings of the 61st ACM/IEEE Design Automation Conference},
  pages={1--6},
  year={2024}
}

@inproceedings{ding2024hybrid,
title={Hybrid {LLM}: Cost-Efficient and Quality-Aware Query Routing},
author={Dujian Ding and Ankur Mallick and Chi Wang and Robert Sim and Subhabrata Mukherjee and Victor R{\"u}hle and Laks V. S. Lakshmanan and Ahmed Hassan Awadallah},
booktitle={The Twelfth International Conference on Learning Representations},
year={2024},
url={https://openreview.net/forum?id=02f3mUtqnM}
}

@ARTICLE{xu2024cached,
  author={Xu, Minrui and Niyato, Dusit and Zhang, Hongliang and Kang, Jiawen and Xiong, Zehui and Mao, Shiwen and Han, Zhu},
  journal={IEEE Transactions on Networking}, 
  title={Cached Model-as-a-Resource: Provisioning Large Language Model Agents for Edge Intelligence in Space–Air–Ground Integrated Networks}, 
  year={2026},
  volume={34},
  number={},
  pages={2850-2864},
  doi={10.1109/TON.2025.3649068}}

@inproceedings{gao2024dlora,
title = "DLoRA: Distributed Parameter-Efficient Fine-Tuning Solution for Large Language Model",
author = "Chao Gao and Zhang, {Sai Qian}",
note = "Publisher Copyright: {\textcopyright} 2024 Association for Computational Linguistics.; 2024 Findings of the Association for Computational Linguistics, EMNLP 2024 ; Conference date: 12-11-2024 Through 16-11-2024",
year = "2024",
doi = "10.18653/v1/2024.findings-emnlp.802",
language = "English (US)",
series = "EMNLP 2024 - 2024 Conference on Empirical Methods in Natural Language Processing, Findings of EMNLP 2024",
publisher = "Association for Computational Linguistics (ACL)",
pages = "13703--13714",
editor = "Yaser Al-Onaizan and Mohit Bansal and Yun-Nung Chen",
booktitle = "EMNLP 2024 - 2024 Conference on Empirical Methods in Natural Language Processing, Findings of EMNLP 2024",
}

@article{10.1145/3641289,
author = {Chang, Yupeng and Wang, Xu and Wang, Jindong and Wu, Yuan and Yang, Linyi and Zhu, Kaijie and Chen, Hao and Yi, Xiaoyuan and Wang, Cunxiang and Wang, Yidong and Ye, Wei and Zhang, Yue and Chang, Yi and Yu, Philip S. and Yang, Qiang and Xie, Xing},
title = {A Survey on Evaluation of Large Language Models},
year = {2024},
issue_date = {June 2024},
publisher = {Association for Computing Machinery},
address = {New York, NY, USA},
volume = {15},
number = {3},
issn = {2157-6904},
url = {https://doi.org/10.1145/3641289},
doi = {10.1145/3641289},
journal = {ACM Trans. Intell. Syst. Technol.},
month = mar,
articleno = {39},
numpages = {45}
}

@article{10.1145/3712701,
author = {Lee, Seungpil and Sim, Woochang and Shin, Donghyeon and Seo, Wongyu and Park, Jiwon and Lee, Seokki and Hwang, Sanha and Kim, Sejin and Kim, Sundong},
title = {Reasoning Abilities of Large Language Models: In-Depth Analysis on the Abstraction and Reasoning Corpus},
year = {2025},
publisher = {Association for Computing Machinery},
address = {New York, NY, USA},
issn = {2157-6904},
url = {https://doi.org/10.1145/3712701},
doi = {10.1145/3712701},
note = {Just Accepted},
journal = {ACM Trans. Intell. Syst. Technol.},
month = jan
}

@article{10.1145/3716629,
author = {Zheng, Junhao and Qiu, Shengjie and Shi, Chengming and Ma, Qianli},
title = {Towards Lifelong Learning of Large Language Models: A Survey},
year = {2025},
issue_date = {August 2025},
publisher = {Association for Computing Machinery},
address = {New York, NY, USA},
volume = {57},
number = {8},
issn = {0360-0300},
url = {https://doi.org/10.1145/3716629},
doi = {10.1145/3716629},
journal = {ACM Comput. Surv.},
month = mar,
articleno = {193},
numpages = {35}
}

@article{10.1145/3706418,
author = {Xu, Mengwei and Cai, Dongqi and Yin, Wangsong and Wang, Shangguang and Jin, Xin and Liu, Xuanzhe},
title = {Resource-efficient Algorithms and Systems of Foundation Models: A Survey},
year = {2025},
issue_date = {May 2025},
publisher = {Association for Computing Machinery},
address = {New York, NY, USA},
volume = {57},
number = {5},
issn = {0360-0300},
url = {https://doi.org/10.1145/3706418},
doi = {10.1145/3706418},
journal = {ACM Comput. Surv.},
month = jan,
articleno = {110},
numpages = {39}
}

@article{motlagh2023digital,
  title={Digital twins for smart spaces—beyond IoT analytics},
  author={Motlagh, Naser Hossein and Zaidan, Martha Arbayani and Lov{\'e}n, Lauri and Fung, Pak Lun and H{\"a}nninen, Tuomo and Morabito, Roberto and Nurmi, Petteri and Tarkoma, Sasu},
  journal={IEEE internet of things journal},
  volume={11},
  number={1},
  pages={573--583},
  year={2023},
  publisher={IEEE}
}

@article{watson1994case,
  title={Case-based reasoning: A review},
  author={Watson, Ian and Marir, Farhi},
  journal={The knowledge engineering review},
  volume={9},
  number={4},
  pages={327--354},
  year={1994},
  publisher={Cambridge University Press}
}

@inproceedings{motlagh2021monitoring,
  title={Monitoring social distancing in smart spaces using infrastructure-based sensors},
  author={Motlagh, Naser Hossein and Toivonen, Pupu and Zaidan, Martha Arbayani and Lagerspetz, Eemil and Peltonen, Ella and Gilman, Ekaterina and Nurmi, Petteri and Tarkoma, Sasu},
  booktitle={2021 IEEE 7th World Forum on Internet of Things (WF-IoT)},
  pages={124--129},
  year={2021},
  organization={IEEE}
}

@misc{10.23729/b9adb0a2-7381-45db-b32f-7e78ae1bc9e3,
author = {University of Oulu},
title = {Smart Campus Oulu indoor climate, air-quality and motion},
howpublished = {\url{https://doi.org/10.23729/b9adb0a2-7381-45db-b32f-7e78ae1bc9e3}},
month = {6},
year = {2021},
note = {University of Oulu, CWC - Verkot ja järjestelmät}
}

@inproceedings{10.1145/3706598.3713606,
author = {G\"{o}ldi, Andreas and Rietsche, Roman and Ungar, Lyle},
title = {Efficient Management of LLM-Based Coaching Agents' Reasoning While Maintaining Interaction Quality and Speed},
year = {2025},
isbn = {9798400713941},
publisher = {Association for Computing Machinery},
address = {New York, NY, USA},
url = {https://doi.org/10.1145/3706598.3713606},
doi = {10.1145/3706598.3713606},
booktitle = {Proceedings of the 2025 CHI Conference on Human Factors in Computing Systems},
articleno = {992},
numpages = {18},
location = {
},
series = {CHI '25}
}

@inproceedings{10.1145/3640794.3665572,
author = {Lee, Jungjae and Choi, Yubin and Song, Minhyuk and Park, Sanghyun},
title = {ChatFive: Enhancing User Experience in Likert Scale Personality Test through Interactive Conversation with LLM Agents},
year = {2024},
isbn = {9798400705113},
publisher = {Association for Computing Machinery},
address = {New York, NY, USA},
url = {https://doi.org/10.1145/3640794.3665572},
doi = {10.1145/3640794.3665572},
booktitle = {Proceedings of the 6th ACM Conference on Conversational User Interfaces},
articleno = {36},
numpages = {8},
location = {Luxembourg, Luxembourg},
series = {CUI '24}
}

@article{10.1145/3643505,
author = {King, Evan and Yu, Haoxiang and Lee, Sangsu and Julien, Christine},
title = {Sasha: Creative Goal-Oriented Reasoning in Smart Homes with Large Language Models},
year = {2024},
issue_date = {March 2024},
publisher = {Association for Computing Machinery},
address = {New York, NY, USA},
volume = {8},
number = {1},
url = {https://doi.org/10.1145/3643505},
doi = {10.1145/3643505},
journal = {Proc. ACM Interact. Mob. Wearable Ubiquitous Technol.},
month = mar,
articleno = {12},
numpages = {38}
}

@inproceedings{10.1145/3686215.3688372,
author = {Nandy, Palash and Adalgeirsson, Sigurdur Orn and Sinha, Anoop K. and Kraljic, Tanya and Cleron, Mike and Shi, Lei and Singh, Angad and Chaudhary, Ashish and Ganti, Ashwin and Melancon, Christopher A and Zhang, Shudi and Robishaw, David and Ciurdar, Horia and Secor, Justin and Robertsen, Kenneth Aleksander and Climer, Kirsten and Le, Madison and Venkatesan, Mathangi and Chi, Peggy and Li, Peixin and McDermott, Peter F and Shim, Rachel and Onsan, Selcen and Vaishnav, Shilp and Guam\'{a}n, Stephanie},
title = {Bespoke: Using LLM agents to generate just-in-time interfaces by reasoning about user intent},
year = {2024},
isbn = {9798400704635},
publisher = {Association for Computing Machinery},
address = {New York, NY, USA},
url = {https://doi.org/10.1145/3686215.3688372},
doi = {10.1145/3686215.3688372},
booktitle = {Companion Proceedings of the 26th International Conference on Multimodal Interaction},
pages = {78–81},
numpages = {4},
location = {San Jose, Costa Rica},
series = {ICMI Companion '24}
}

@article{10.1145/3698145,
author = {Zeng, Xin and Wang, Xiaoyu and Zhang, Tengxiang and Yu, Chun and Zhao, Shengdong and Chen, Yiqiang},
title = {GestureGPT: Toward Zero-Shot Free-Form Hand Gesture Understanding with Large Language Model Agents},
year = {2024},
issue_date = {December 2024},
publisher = {Association for Computing Machinery},
address = {New York, NY, USA},
volume = {8},
number = {ISS},
url = {https://doi.org/10.1145/3698145},
doi = {10.1145/3698145},
journal = {Proc. ACM Hum.-Comput. Interact.},
month = oct,
articleno = {545},
numpages = {38}
}

@inproceedings{10.1145/3701716.3717734,
author = {Wo\'{z}niak, Stanis\l{}aw and Duszenko, Jacek and Koco\'{n}, Jan and Kazienko, Przemysaw},
title = {Improving LLM-Based Recommender Systems with User-Controllable Profiles},
year = {2025},
isbn = {9798400713316},
publisher = {Association for Computing Machinery},
address = {New York, NY, USA},
url = {https://doi.org/10.1145/3701716.3717734},
doi = {10.1145/3701716.3717734},
booktitle = {Companion Proceedings of the ACM on Web Conference 2025},
pages = {2102–2111},
numpages = {10},
location = {Sydney NSW, Australia},
series = {WWW '25}
}

@article{10.1145/3731446,
author = {Huang, Xu and Lian, Jianxun and Lei, Yuxuan and Yao, Jing and Lian, Defu and Xie, Xing},
title = {Recommender AI Agent: Integrating Large Language Models for Interactive Recommendations},
year = {2025},
issue_date = {July 2025},
publisher = {Association for Computing Machinery},
address = {New York, NY, USA},
volume = {43},
number = {4},
issn = {1046-8188},
url = {https://doi.org/10.1145/3731446},
doi = {10.1145/3731446},
journal = {ACM Trans. Inf. Syst.},
month = jun,
articleno = {96},
numpages = {33}
}

@article{10.1145/3715114,
author = {Chen, Xiaohong and Chen, Shi and Jin, Zhi and Bian, Han and Chen, Zihan and Li, Haotian},
title = {Expressing the Needs in Smart Home: What Is the End Users’ Favorite Way},
year = {2025},
issue_date = {April 2025},
publisher = {Association for Computing Machinery},
address = {New York, NY, USA},
volume = {32},
number = {2},
issn = {1073-0516},
url = {https://doi.org/10.1145/3715114},
doi = {10.1145/3715114},
journal = {ACM Trans. Comput.-Hum. Interact.},
month = apr,
articleno = {16},
numpages = {38}
}

@article{10.1145/3716132,
author = {Huang, Tian and Yu, Chun and Shi, Weinan and Peng, Zijian and Yang, David and Sun, Weiqi and Shi, Yuanchun},
title = {Prompt2Task: Automating UI Tasks on Smartphones from Textual Prompts},
year = {2025},
issue_date = {June 2025},
publisher = {Association for Computing Machinery},
address = {New York, NY, USA},
volume = {32},
number = {3},
issn = {1073-0516},
url = {https://doi.org/10.1145/3716132},
doi = {10.1145/3716132},
journal = {ACM Trans. Comput.-Hum. Interact.},
month = jun,
articleno = {25},
numpages = {45}
}

@article{10.1145/3748302,
author = {Zhang, Zeyu and Dai, Quanyu and Bo, Xiaohe and Ma, Chen and Li, Rui and Chen, Xu and Zhu, Jieming and Dong, Zhenhua and Wen, Ji-Rong},
title = {A Survey on the Memory Mechanism of Large Language Model based Agents},
year = {2025},
publisher = {Association for Computing Machinery},
address = {New York, NY, USA},
issn = {1046-8188},
url = {https://doi.org/10.1145/3748302},
doi = {10.1145/3748302},
note = {Just Accepted},
journal = {ACM Trans. Inf. Syst.},
month = jul
}

@misc{Langchain,
author = {LangChain},
title = {LangChain},
 howpublished = {\url{https://www.langchain.com/}},
note = {Last accessed: \today}
}

@article{10.1145/3686803,
author = {Li, Jialong and Zhang, Mingyue and Li, Nianyu and Weyns, Danny and Jin, Zhi and Tei, Kenji},
title = {Generative AI for Self-Adaptive Systems: State of the Art and Research Roadmap},
year = {2024},
issue_date = {September 2024},
publisher = {Association for Computing Machinery},
address = {New York, NY, USA},
volume = {19},
number = {3},
issn = {1556-4665},
url = {https://doi.org/10.1145/3686803},
doi = {10.1145/3686803},
journal = {ACM Trans. Auton. Adapt. Syst.},
month = sep,
articleno = {13},
numpages = {60}
}

@book{biswas2025building,
  title={Building Agentic AI Systems: Create intelligent, autonomous AI agents that can reason, plan, and adapt},
  author={Biswas, Anjanava and Talukdar, Wrick},
  year={2025},
  ISBN-13={9781801079273},
  publisher={Packt Publishing Ltd},
address={Online},
url={https://www.packtpub.com/en-us/product/building-agentic-ai-systems-9781801079273?srsltid=AfmBOooA3w-pOUSKfa6PPXhCMBy1kCRlQRKaK3PwhEAPI78EcQrliF8G}
}

@article{saleh2025memindex,
  title={Memindex: Agentic event-based distributed memory management for multi-agent systems},
  author={Saleh, Alaa and Tarkoma, Sasu and Lindgren, Anders and Donta, Praveen Kumar and Dustdar, Schahram and Pirttikangas, Susanna and Lov{\'e}n, Lauri},
  journal={ACM Transactions on Autonomous and Adaptive Systems},
  year={2025},
  publisher={ACM New York, NY}
}

@inproceedings{saleh2025llm,
  title={LLM-powered Smart Spaces with Multi-agent User-centric Adaptation},
  author={Saleh, Alaa and Hassaan, Mazen and Zhang, Qiyang and Tarkoma, Sasu and Pirttikangas, Susanna and Lov{\'e}n, Lauri},
  booktitle={2025 IEEE 45th International Conference on Distributed Computing Systems Workshops (ICDCSW)},
  pages={576--581},
  year={2025},
  organization={IEEE}
}
